\documentclass{article}

\usepackage{microtype}
\usepackage{graphicx}
\usepackage{amsmath}
\usepackage{amssymb}
\usepackage{amsfonts}       % blackboard math symbols
\usepackage{mathtools}
\usepackage{booktabs} % for professional tables
\usepackage{enumitem}
\usepackage{soul}
\usepackage{xspace}
\usepackage{subcaption}
\usepackage{float}
\usepackage{algorithmic,algorithm}

\usepackage[skip=.1in]{caption}

\usepackage[colorlinks=true,citecolor=blue,urlcolor=black]{hyperref}

\usepackage[preprint]{neurips_2022}

\usepackage[utf8]{inputenc} % allow utf-8 input
\usepackage[T1]{fontenc}    % use 8-bit T1 fonts
\usepackage{hyperref}       % hyperlinks
\usepackage{url}            % simple URL typesetting
\usepackage{booktabs}       % professional-quality tables
\usepackage{amsfonts}       % blackboard math symbols
\usepackage{nicefrac}       % compact symbols for 1/2, etc.
\usepackage{microtype}      % microtypography
\usepackage{xcolor}         % colors

\usepackage{wrapfig}

\newcommand{\localtrain}{\textsc{LocalTrain}\xspace}
\newcommand{\serverupdate}{\textsc{ServerUpdate}\xspace}

\newcommand{\anyround}{\textsc{Round}\xspace}
\newcommand{\deterministicround}{\textsc{DeterministicRound}\xspace}
\newcommand{\ditheredround}{\textsc{DitheredRound}\xspace}
\newcommand{\stochasticround}{\textsc{StochasticRound}\xspace}
\newcommand{\leadingzeros}{\textsc{LeadingZeros}\xspace}
\newcommand{\eliasgamma}{\textsc{Gamma}\xspace}
\newcommand{\sign}{\textsc{Sign}\xspace}

\newcommand{\nocompression}{\textsc{NoCompression}\xspace}
\newcommand{\onebitsgd}{\textsc{1BitSGD}\xspace}
\newcommand{\terngrad}{\textsc{TernGrad}\xspace}
\newcommand{\topk}{\textsc{Top-K}\xspace}
\newcommand{\drive}{\textsc{DRIVE}\xspace}
\newcommand{\threelc}{\textsc{3LC}\xspace}
\newcommand{\qsgd}{\textsc{QSGD}\xspace}

\newcommand{\fedsketch}{\textsc{FedSKETCH}\xspace}

\DeclareMathOperator{\Prob}{Prob}
\DeclarePairedDelimiterX{\divergence}[2]{[}{]}{#1\;\delimsize\|\;#2}

\newcommand{\E}{\operatorname{\mathbb E}}
\newcommand{\R}{\operatorname{\mathbb R}}
\newcommand{\Z}{\operatorname{\mathbb Z}}

\allowdisplaybreaks[2]
\graphicspath{{figures/}}

\begin{document}

\title{Optimizing the Communication--Accuracy Trade-off\\ in Federated Learning with Rate--Distortion Theory}

\author{%
    Nicole Mitchell\\
    Google Research \\
    San Francisco, CA \\
    \texttt{nicolemitchell@google.com} \\
    \And
    Johannes Ball\'{e} \\
    Google Research \\
    Mountain View, CA \\
    \texttt{jballe@google.com} \\
    \AND
    Zachary Charles \\
    Google Research \\
    Seattle, WA \\
    \texttt{zachcharles@google.com} \\
    \And
    Jakub Kone\v{c}n\'{y} \\
    Google Research \\
    London, UK \\
    \texttt{konkey@google.com} \\
}

\maketitle

\begin{abstract}
A significant bottleneck in federated learning (FL) is the network communication cost of sending model updates from client devices to the central server. We present a comprehensive empirical study of the statistics of model updates in FL, as well as the role and benefits of various compression techniques. Motivated by these observations, we propose a novel method to reduce the average communication cost, which is near-optimal in many use cases, and outperforms \topk, \drive, \threelc and \qsgd on Stack Overflow next-word prediction, a realistic and challenging FL benchmark. This is achieved by examining the problem using rate--distortion theory, and proposing distortion as a reliable proxy for model accuracy. Distortion can be more effectively used for optimizing the trade-off between model performance and communication cost across clients. We demonstrate empirically that in spite of the non-i.i.d. nature of federated learning, the rate--distortion frontier is consistent across datasets, optimizers, clients and training rounds.
\end{abstract}

\section{Introduction}
\label{intro}
Federated learning (FL) is a machine learning framework in which clients collaboratively train a model under the coordination of a central server or service provider, without sharing their local data.  Prototypical FL algorithms such as FedAvg~\citep{pmlr-v54-mcmahan17a} involve multiple communication rounds in which clients train on their own local data, sharing only their model updates with the server. While federated learning can incur the benefits of centralized learning without the need to store sensitive user data, it also introduces challenges related to data heterogeneity and network constraints~\citep{aopfl}.

\citet[Table 1]{aopfl} propose a taxonomy that divides FL into \emph{cross-device} and \emph{cross-silo} regimes, characterized by their practical constraints. In cross-silo FL, there are a small number of computationally reliable clients. In cross-device FL, there are many intermittently available and computationally unreliable clients. While both regimes face optimization challenges stemming from data heterogeneity and communication constraints~\citep{mlsysGoogleFL, ibmFL, fateFL, huba2021papaya, fieldguideFL}, communication efficiency is particularly critical in cross-device settings, due to network and bandwidth limitations~\citep{mlsysGoogleFL}.

Cross-device FL also faces challenges due to partial participation. Typically, a small subset of clients participate in each communication round, and each client may only participate once (if at all) throughout the entire training procedure~\citep{aopfl}. Communication rounds of cross-device FL often involve clients who have not previously participated, and therefore have no useful \emph{state} learned across previous rounds. Even if we can identify clients who have participated before, such state can suffer from staleness and actually reduce accuracy~\citep{reddi2021adaptive}. Thus, cross-device FL typically employs \emph{stateless} algorithms, in which all training quantities used by a client (save for their data) are provided by the server at each round (e.g., model weights, hyperparameters).

In this work we focus on compression methods for client updates in cross-device settings. While client updates in algorithms such as FedAvg are not actually gradients~\citep{reddi2021adaptive}, client update compression is closely related to gradient compression methods for distributed training (DT). Such methods often employ techniques like quantization \citep{alistarh2017qsgd, suresh2017distributed}, sparsification \citep{aji2017sparse, lin2018deep}, and low rank decomposition \citep{wang2018atomo, vogels2019powersgd} in order to obtain efficiently communicable estimates of gradients.

Although such methods have already been applied to and refined in the context of FL~\citep{sattler2019robust, rothchild2020fetchsgd, reisizadeh2020fedpaq, haddadpour2021federated}, we find that such methods can suffer from a lack of practicality, effectiveness, or flexibility. To that end, we identify three desirable properties of a compression method for FL that (to the best of our knowledge) have not been addressed simultaneously.

\textbf{Desiderata. }First, we would like our method to be stateless. Many prior compression methods for FL (e.g., \citet{sattler2019robust, rothchild2020fetchsgd}) cannot be directly applied to the cross-device setting, as they involve stateful compression algorithms. One commonly used stateful component of such algorithms is an error-feedback mechanism, in which the compression error for a given round is maintained and incorporated into the compression of the subsequent round (see \citep{xu2020compressed} for a more complete discussion on this). As discussed above, the low participation rate and large number of clients restrict us (practically speaking) to stateless compression methods.

Second, while a number of gradient compression techniques can be applied in a stateless manner, their analysis and effectiveness is often based on their worst-case guarantees~\citep{alistarh2017qsgd, suresh2017distributed, beznosikov2020biased, albasyoni2020optimal, vargaftik2021drive}. While useful in the abstract, this analysis ignores statistical properties of the information to be compressed. By contrast, we observe (and show below) that client updates in FL often have a consistent structure that can be exploited to improve efficiency. Across clients, training rounds, models and optimizers, client updates are often highly sparse and follow a pattern similar to a power law distribution tightly centered around zero. By selecting our compression operator carefully, we can leverage this structure to generate a more efficient representation of a client's update for the \emph{average} case.

Last, we would like to leverage rate--distortion optimization to guide the design and usage of our method. Notably, lossy compression techniques such as quantization introduce error and yield a trade-off between the size of the compressed quantity (\emph{bitrate} $R$) and its fidelity (\emph{distortion} $D$). Different applications of FL may target a specific rate or a specific distortion, while minimizing the other, or may aim to minimize both jointly. Basing our choice of updates on rate--distortion optimization allows us to develop a method that trades off between model performance and communication cost globally, i.e. across clients, by specifying a single hyperparameter.

\section{Method}
\label{sec:compression-scheme}

In FL, we often wish to find a model $\theta \in \mathbb{R}^d$ that minimizes a weighted average of client losses
\begin{equation}\label{eq:fl_objective}
\min_\theta f(\theta), \text{ with } f(\theta) = \sum_{k=1}^K w_kf_k(\theta)
\end{equation}
where $K$ is the total number of clients and $f_k$, $w_k$ are the loss function and weight of client $k$. For practical reasons, $w_k$ is often the number of examples held by client $k$ (\emph{example weighting}), which can incur optimization benefits~\citep{li2020federated}. We wish to solve \eqref{eq:fl_objective} without sharing data and with minimal client-to-server communication. To do so we combine FedOpt~\citep{reddi2021adaptive} (generalizing FedAvg~\citep{pmlr-v54-mcmahan17a}) with compression.

In the FedOpt framework, at each round $t$, the server broadcasts its model $\theta_t$ to a set of clients $\mathcal{S}_t$. Each client $k \in \mathcal{S}_t$ uses a procedure $\localtrain$ to train its model locally. $\localtrain(\theta, f)$ is often multiple steps of SGD on $f$ starting at $\theta$. After computing $\theta_t^k = \localtrain(\theta_t, f_k)$, the client sends its weighted update $u_t^k := w_k(\theta_t^k - \theta_t)$ to the server.

To reduce communication, clients can instead send a compressed update $c_t^k := \mathcal{E}(u_t^k)$ to the server, where $\mathcal{E}$ is some encoder. The server decodes the client updates using a decoder $\mathcal{D}$, and computes a weighted average $g_t$ of the $\mathcal{D}(c_t^k)$ (using the weight $w_k$). Finally, the server updates its model using a procedure \serverupdate. As proposed by \citet{reddi2021adaptive}, \serverupdate is typically a first-order optimization step, treating $g_t$ as a gradient estimate, with server learning rate $\eta_s$. For example, if \serverupdate is gradient descent, then $\serverupdate(\theta, g) = \theta - \eta_sg$.

\begin{figure}
\begin{minipage}{0.48\textwidth}
\begin{algorithm}[H]
   \caption{FedOpt with compression}
   \label{alg:fed_opt}
\begin{algorithmic}
    \STATE \textbf{Input:} Number of rounds $T$, initial model $\theta_0 \in \mathbb{R}^d$, \localtrain, \serverupdate, encoder $\mathcal{E}$, decoder $\mathcal{D}$
    \FOR{$t = 0, \dots, T$}
        \STATE$\mathcal{S}_t \gets $(random set of $m$ clients)
        \STATE Broadcast $\theta_t$ to all clients $k \in \mathcal{S}_t$
        \FOR{each client $k \in \mathcal{S}_t$ \textbf{in parallel}}
            \STATE $\theta_t^k \gets \localtrain(\theta_t, f_k)$
            \STATE Compute $u_t^k = w_k(\theta_t^k - \theta_t)$
            \STATE Send $c_t^k = \mathcal{E}(u_t^k)$ to the server
        \ENDFOR
        \STATE $g_t \gets \dfrac{\sum_{k \in \mathcal{S}_t} \mathcal{D}(c_t^k)}{\sum_{k \in \mathcal{S}_t} w_k}$
        \STATE $\theta_{t+1} \gets \serverupdate(\theta_t, g_t)$
   \ENDFOR
\end{algorithmic}
\end{algorithm}
\end{minipage}%
\hfill
\begin{minipage}{0.48\textwidth}
\centering
\includegraphics[height=.75\linewidth]{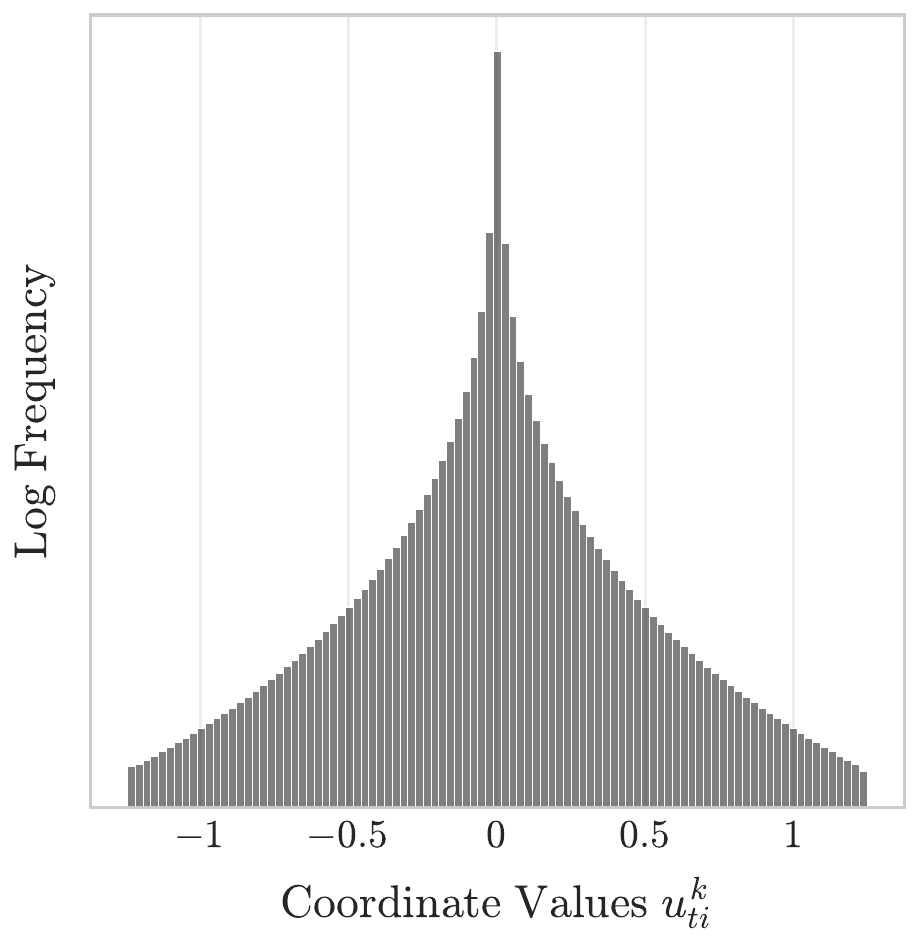}
\caption{Histogram of coordinate values of weighted client updates, averaged over the course of training across all participating clients in Stack Overflow NWP.
}
\label{so-fedadam-hist}
\end{minipage}
\end{figure}

Algorithm \ref{alg:fed_opt} summarizes our framework. Similar algorithms appear elsewhere \citep[e.g.,][]{haddadpour2021federated}. The concern of this paper is to develop appropriate encoding and decoding operators $\mathcal{E}$, $\mathcal{D}$ and control them in a way that is aligned with the global rate--distortion trade-off. In the remainder of this section, we lay out the observations and derivations that underpin our choices.

\subsection{Quantization}

\begin{figure*}[b]
\centering
\includegraphics[width=0.95\linewidth]{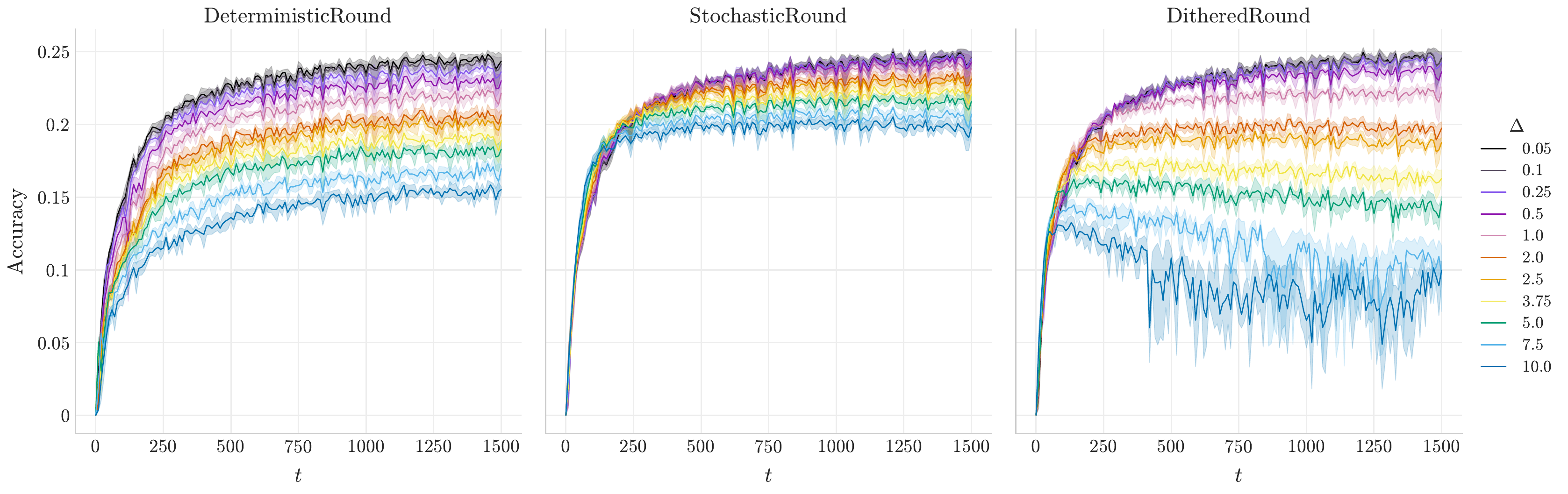}
\vspace{-0.2cm}
\caption{
Accuracy versus round $t$ for varying quantization step sizes and rounding methods on Stack Overflow NWP.
}
\label{so-acc}
\end{figure*}

In order to efficiently communicate client updates, we reduce their fidelity via quantization~\citep{GrNe98}. We use coordinate-wise (scalar) quantization due to its computational tractability. In order to choose an appropriate quantization method, we track the statistical structure of client updates over the course of training using FedOpt on various FL tasks (see Section~\ref{exp-setup}). We find that across tasks, optimizers, types of neural network layers, clients and training rounds, the coordinates of weighted client updates follow a consistent symmetric, unimodal, sparse, and heavy-tailed distribution centered around zero, as illustrated in Figure~\ref{so-fedadam-hist}. Throughout the body of our work, we present results on the Stack Overflow NWP task (details in Appendix~\ref{appendix:experiment_details}). Full results on other tasks are given in Appendix~\ref{appendix:full_results}.

Given this statistical structure, there are two efficient ways to represent this data: using a fixed-length bit representation of each possible quantized value~\citep{konevcny2016federated, vargaftik2021drive}, or using a variable-length bit representation~\citep{suresh2017distributed,alistarh2017qsgd} to represent more likely values with shorter bit sequences, optimizing average representation length (see Section~\ref{entropy-coding}). In addition, we can decide to limit quantization to uniform quantization (i.e., rounding), where the size of each quantization bin is identical, which has computational benefits.

While fixed-bit representations have advantages in differential privacy applications, when combined with uniform quantization such an approach would be inefficient in our setting, effectively representing extremely likely and extremely unlikely values with the same number of bits. Ideally, one would adjust the quantization boundaries such that each quantization bucket has approximately equal likelihood. However, nonuniform quantization tends to be more computationally complex than rounding.

Thus, we elect to use variable-length representations with uniform quantization. We implement uniform quantization by scaling by a positive quantization step size $\Delta \in \R_{>0}$, rounding coordinate-wise to integers, then un-scaling in the decoder. The quantization step size $\Delta$ controls the granularity with which continuous client update values are discretized: a large $\Delta$ yields low-resolution client updates with smaller information content, while a small $\Delta$ yields high-resolution client updates with more information to be transmitted. We consider three methods of rounding: First, $\deterministicround(u,\Delta)$ simply rounds $u/\Delta$ to the nearest integer. Second,
\begin{align*}\stochasticround(u,\Delta) =
\begin{cases}
\lceil u/\Delta \rceil, &\text{with prob. } p = u/\Delta - \lfloor u/\Delta \rfloor \\
\lfloor u/\Delta \rfloor, &\text{with prob. } 1-p. \\
\end{cases}
\end{align*}
Last, $\ditheredround(u,\Delta)$ samples $z \sim \mathcal{U}(-.5, .5)$ and rounds $u/\Delta + z$ to the nearest integer, subtracting $z$ again on the decoder before un-scaling \citep{Schuchman1964-np}. In practice, $z$ is generated pseudo-randomly from a seed so that it is available on both sides. Note that the latter two methods are unbiased (i.e., $\E [\anyround(u/\Delta)] = u/\Delta$).

\textbf{Empirical Evaluation. }We find that for a full range of quantization step sizes $\Delta$, \stochasticround consistently outperforms \deterministicround and \ditheredround across datasets, tasks and optimizers. The comparably high levels of noise introduced by \ditheredround require lower learning rates at coarser quantization levels which slows convergence, and thus renders it unsuitable for our FL simulations. Across quantization step sizes $\Delta$, there is much less variance in the accuracy over training when using \stochasticround, as depicted in Figure~\ref{so-acc}. In the sequel, we only use \stochasticround.

\subsection{Entropy Coding}\label{entropy-coding}
\begin{figure}[b]
\begin{minipage}{.48\textwidth}
\centering
\includegraphics[height=.75\linewidth]{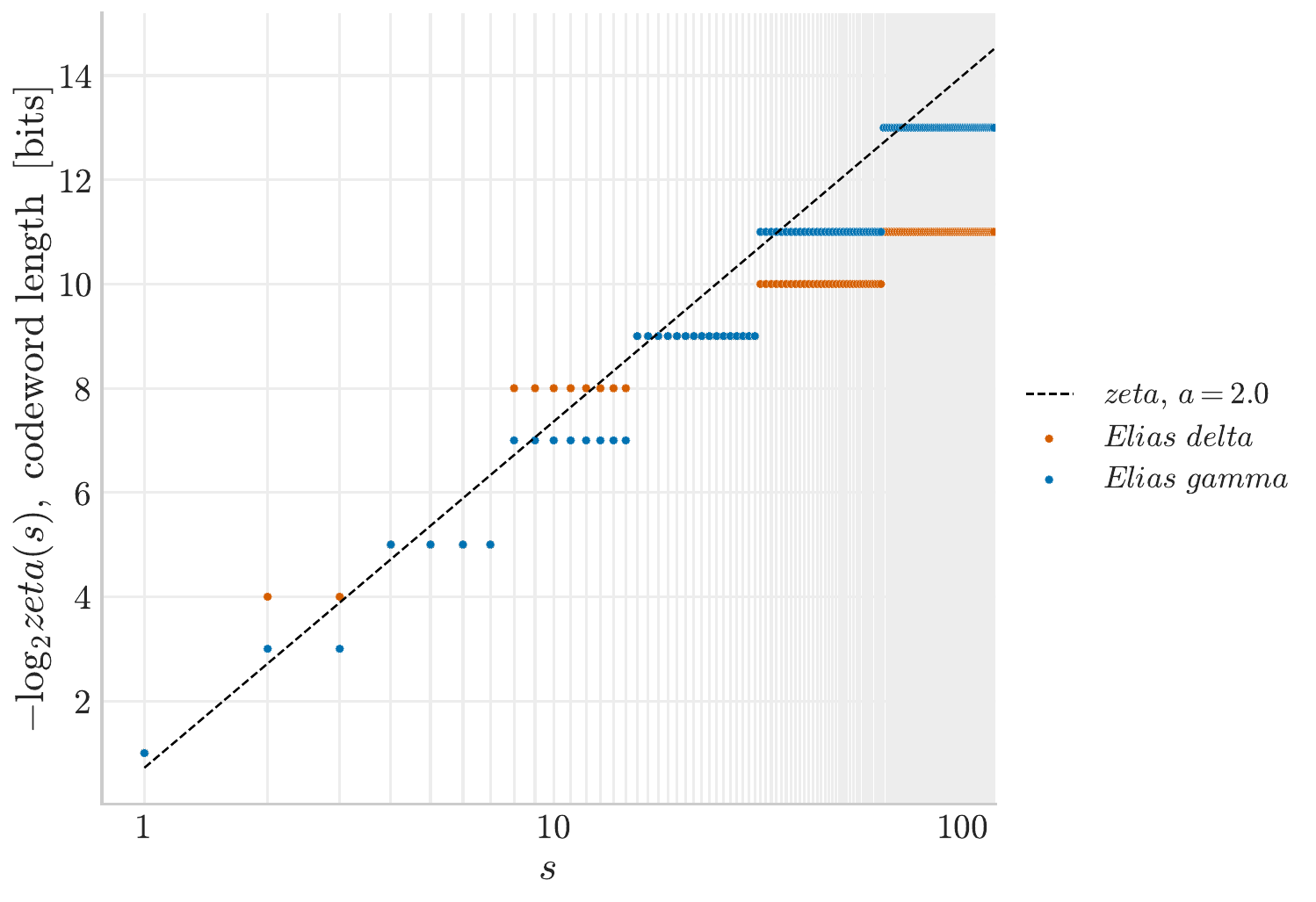}
\vspace{-0.2cm}
\caption{\emph{Elias gamma} and \emph{Elias delta} codeword lengths for the integers roughly follow a \emph{zeta} distribution.}
\label{elias-gamma-elias-delta-zeta}
\end{minipage}%
\hfill
\begin{minipage}{.48\textwidth}
\centering
\includegraphics[height=.75\linewidth]{sparsity/stackoverflow_word-fedadam-sparsity}
\vspace{-0.2cm}
\caption{Average sparsity of quantized client updates across step sizes in Stack Overflow NWP.
}
\label{so-sparsity}
\end{minipage}
\end{figure}

After quantizing client updates $u_t^k$ to integers $q_t^k$, we apply two \emph{entropy coding} techniques that exploit statistical structure to yield smaller representations without distortion: \emph{universal coding} and \emph{run length encoding}. Entropy codes map symbols of the source alphabet, in this case the integers, to binary codewords: $\Z \rightarrow \{0,1\}^*$. According to the source coding theorem, the least possible average number of bits required to communicate a sequence is the entropy of that sequence \citep{shannon1948}. Entropy codes aim to represent a symbol $s$ with a codeword (bit sequence) close in length to $-\log_2 \Prob(s)$, such that in expectation, the bitrate is close to the sequence's entropy.

\textbf{Universal Coding.}
Since the empirical distribution of client update coordinates closely resembles a symmetric, double-sided power law (with a spike at zero, see Figure~\ref{so-fedadam-hist}), we consider using the \emph{gamma} and \emph{delta} universal codes~\citep{El75} to encode the magnitude of each non-zero quantized coordinate of $q_t^k$. As depicted in Figure~\ref{elias-gamma-elias-delta-zeta}, both of these codes reasonably match a \emph{zeta distribution}, the discrete version of a power law, making them good candidates.

Additionally, these codes admit worst-case guarantees, i.e. the expected bitrates of the \emph{gamma} and \emph{delta} codes are upper bounded by $3\times$ and $4\times$ the entropy of the empirical distribution, respectively, as long as probability decreases with magnitude~\citep{El75}. They also do not require bounding the integer magnitudes from above as other codes would, which eliminates a hyperparameter. The signs of the non-zero coordinates can be encoded efficiently using one additional bit per non-zero coordinate, since the empirical distribution is very close to symmetric.

\textbf{Run Length Encoding.}
The likelihood of zeros in the empirical distribution is significantly higher than suggested by a power law, i.e., client updates are sparse, as illustrated in Figure~\ref{so-sparsity}. To efficiently communicate these sparse tensors, we encode the number of repeated zeros between each non-zero coordinate, rather than transmitting each occurrence. For simplicity of implementation, we reuse the same universal code for encoding these run lengths that we use for the non-zero magnitudes.

\begin{figure*}
\centering
\includegraphics[width=\linewidth]{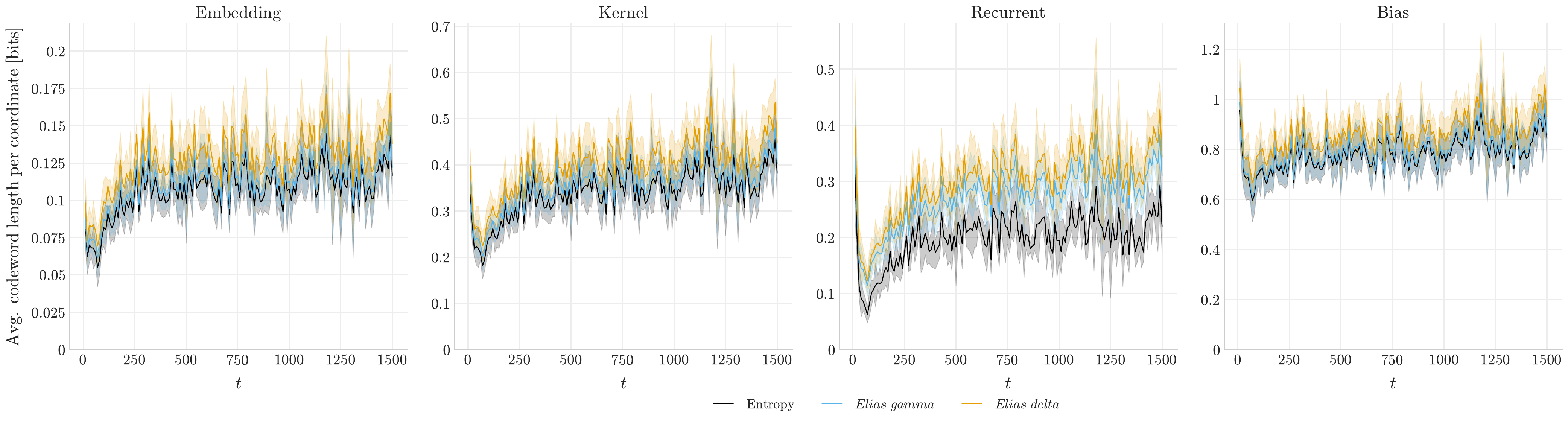}
\vspace{-0.5cm}
\caption{Average codeword length of \emph{Elias gamma} and \emph{delta} codes when applied to quantized client updates (excluding zero coordinates) in Stack Overflow NWP with $\Delta = 0.05$, as well as the entropy of the empirical distribution (excluding zeros). Results are plotted for each layer type of the model.
}
\label{entropy-cross-entropy}
\end{figure*}
\textbf{Empirical Evaluation.}
We measure how the average codeword lengths using the \emph{delta} and \emph{gamma} codes compare to the entropy of the client updates and observe which of these universal codes has the smaller rate (Figure~\ref{entropy-cross-entropy}). We test this across training rounds, quantization step sizes, datasets, tasks, and model layer types. We find that the \emph{gamma} code consistently outperforms the \emph{delta} code with an acceptable overhead over the entropy of the client updates, typically less than 20\%, independently of how finely we quantize.

\begin{wrapfigure}[11]{R}{0.65\textwidth}
\vspace{-1.1cm}
\begin{minipage}{0.65\textwidth}
\begin{algorithm}[H]
\caption{Client-side encoder $\mathcal{E}$}
\label{alg:encoder}
\begin{algorithmic}
\REQUIRE client update $u \in \R^d$, quantization step size $\Delta > 0$
\ENSURE encoded client update $c \in \{0,1\}^*$
\STATE $q \gets \stochasticround(u, \Delta)$
\STATE $c \gets \square$; $i \gets 0$
\WHILE{$i<d$}
    \STATE $r \gets \leadingzeros(q_{i:})$
    \STATE $c \gets c \oplus \eliasgamma (r+1)$; $i \gets i + r$
    \STATE $c \gets c \oplus \sign(q_i) \oplus \eliasgamma(|q_i|)$; $i \gets i + 1$
\ENDWHILE
\STATE \textbf{return} $c$
\end{algorithmic}
\end{algorithm}
\end{minipage}
\end{wrapfigure}

The observations above lead us to the encoding operator $\mathcal{E}$ given in Algorithm~\ref{alg:encoder}. We let $\square$ denote the empty binary string and $\oplus$ denote concatenation of binary strings, e.g. $110 \oplus 10 = 11010$. The decoding operator $\mathcal{D}$ simply consists of parsing the \emph{gamma} code from the binary string, inserting zeros, recovering signs, and multiplying by $\Delta$.\footnote{Open-source implementations of the operators are available at \url{https://github.com/tensorflow/compression}.}

\subsection{Rate--Distortion Optimization}
\label{rd-theory}
With the encoder algorithm determined, we aim to achieve the desired trade-off between total bitrate and final model performance by controlling the quantization parameters. For example, we may wish to optimize the performance of the model (e.g., final accuracy), such that the total rate $R$ is below an acceptable bitrate budget $B$:
\begin{equation}
\label{eq:rate_accuracy}
\min f(\theta) \text{ s.t. } R(\mathcal Q) = \sum_{t,k} |\mathcal{E}(u_t^k, \Delta_t^k)| \le B,
\end{equation}
where $|\cdot|$ denotes bit sequence length and $\Delta_t^k$ is the quantization step size used by client $k$ in round $t$, chosen by a policy $\mathcal Q$. For such a policy to effectively operate on the client side without additional server--client communication or state, we need a proxy for final model performance that is tractable to predict from data available at the client. For this, we consider total distortion
\begin{equation}
\label{eq:distortion_proxy}
D(\mathcal Q) = \sum_{t,k} \bigl\Vert u_t^k - \mathcal D(\mathcal E(u_t^k, \Delta_t^k), \Delta_t^k)\bigr\Vert_2^2,    
\end{equation}
yielding the \emph{rate--distortion optimization} problem $\min D(\mathcal Q) \text{ s.t. } R(\mathcal Q) \le B$ \citep{CoTh06}. We verify empirically that for varying $\Delta$, total distortion is a good proxy for model performance (Figure~\ref{so-acc-distortion}). This relationship holds across tasks (see Appendix~\ref{appendix:distortion-acc}).

Alternatively to \eqref{eq:rate_accuracy}, we can place a limit on distortion while minimizing the rate. In either case, the global optimum may be found by minimizing the Lagrangian $L(\mathcal Q, \lambda) = R(\mathcal Q) + \lambda D(\mathcal Q)$, where $\lambda$ is the Lagrange multiplier. This is a well-studied approach in lossy data compression, and due to the separability of both terms across updates, is reminiscent of the principle of Pareto efficiency, where $\mathcal Q$ represents a mode of resource allocation. In practice, $\lambda$ is often simply adjusted until the objective's hard constraint is satisfied.
Since the Lagrangian is separable across updates, the optimal policy yielding quantization step sizes $\Delta_t^k = \mathcal Q(\lambda, u_t^k)$ can be found in a distributed way: while $\lambda$ is chosen centrally, each client solves the problem
\begin{equation}
\label{eq:client_rd}
\Delta_t^k = \arg\min_\delta |\mathcal{E}(u_t^k, \delta)| + \lambda \bigl\Vert u_t^k - \mathcal D(\mathcal E(u_t^k, \delta), \delta)\bigr\Vert_2^2.
\end{equation}
However, an exhaustive approach to this minimization can place an unacceptable computational burden on the clients, which leads us to the following experiment.

\textbf{Experiment. }For varying $\lambda$, we let clients solve \eqref{eq:client_rd} by grid search, selecting from a pre-determined set of $\delta$'s. We observe significant agreement on $\Delta_t^k$ across clients and rounds for any given $\lambda$  (Figure~\ref{so-fedadam-vote}). That is, empirically, $\Delta_t^k$ is largely independent of $u_t^k$, and there is a monotonic relationship between $\Delta_t^k$ and $\lambda$. Further, we find that this relationship is even consistent across different architectures, tasks, and optimizers (Appendix~\ref{appendix:votes}, Appendix~\ref{appendix:rd}).

\begin{figure}
\centering
\begin{minipage}{.48\textwidth}
\centering
\includegraphics[height=.75\linewidth]{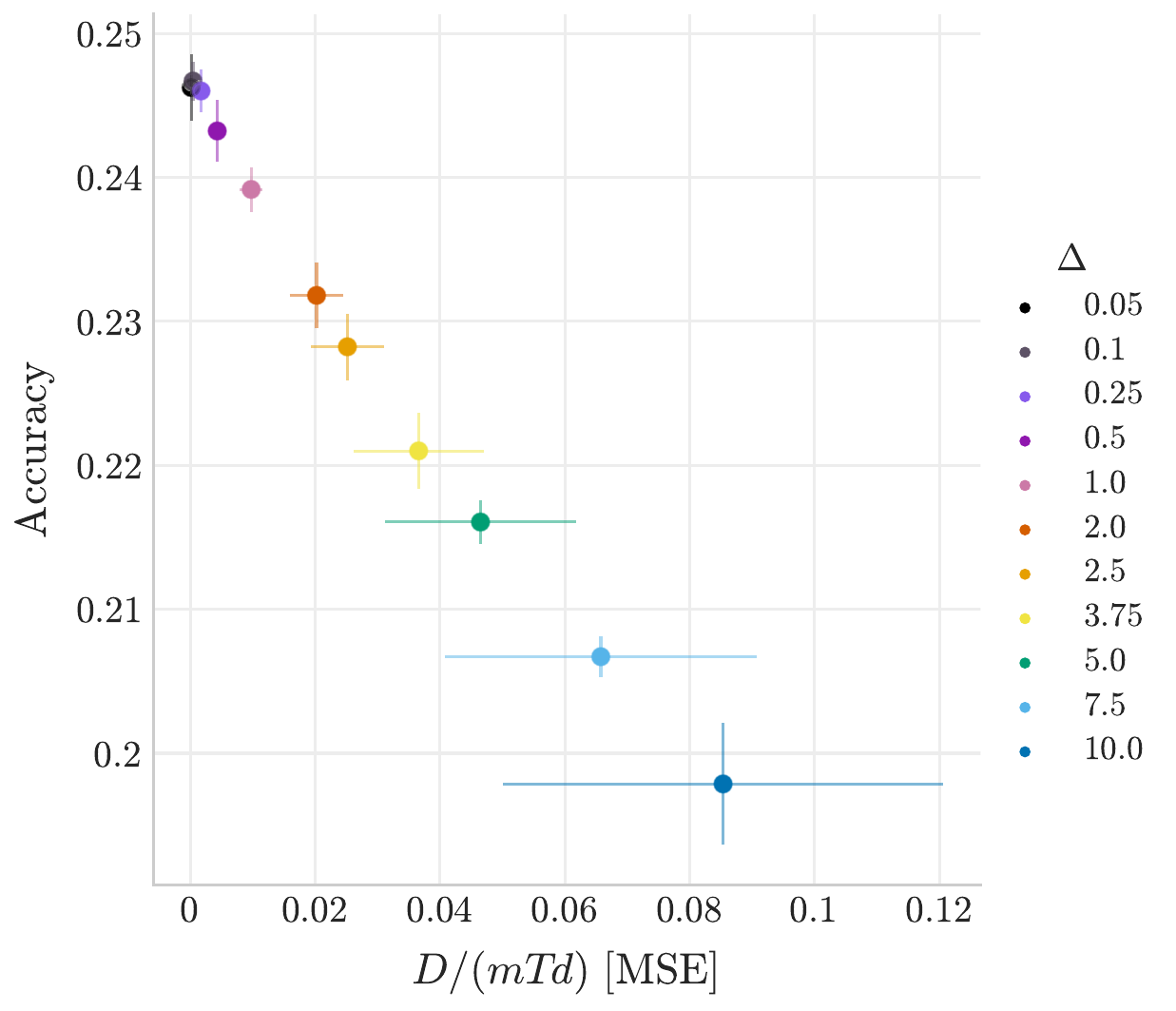}
\vspace{-0.2cm}
\caption{\label{so-acc-distortion}%
Accuracy of a model versus average per-coordinate distortion for various quantization step sizes $\Delta$ on Stack Overflow NWP. Error bars indicate the variance over 5 random trials.
}
\end{minipage}%
\hfill
\begin{minipage}{.48\textwidth}
\centering
\includegraphics[height=.75\linewidth]{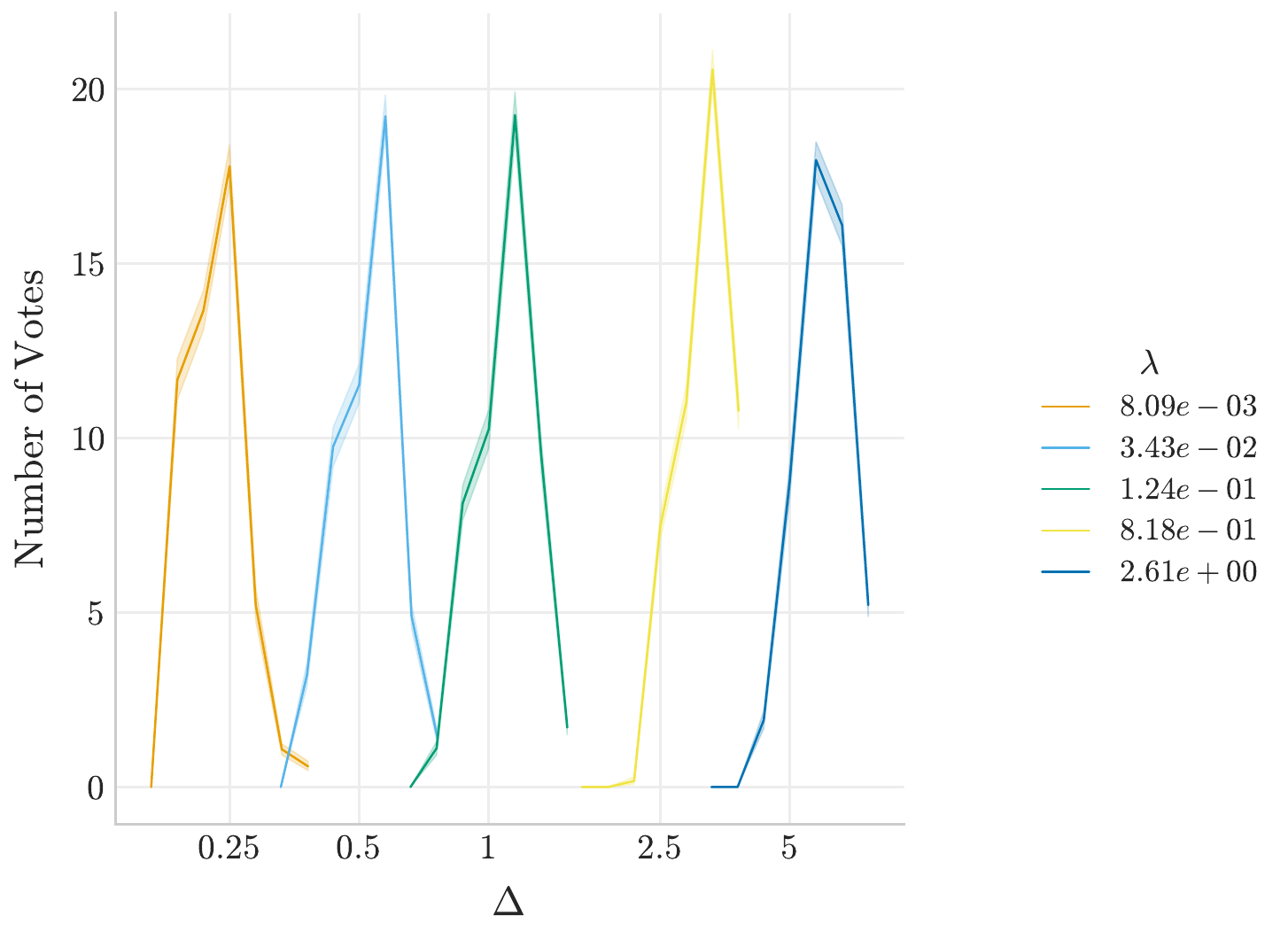}
\vspace{-0.2cm}
\caption{Histogram of client-selected quantization step sizes $\Delta_t^k$ for a given $\lambda$, averaged across training rounds, on Stack Overflow NWP.
}
\label{so-fedadam-vote}
\end{minipage}
\end{figure}

Thus, we can set $\mathcal Q(\lambda, u_t^k) \equiv \Delta$, and let the server control a global $\Delta$ rather than $\lambda$, eliminating a hyperparameter and the need to solve \eqref{eq:client_rd} on the client side. The client side of our method is thus remarkably simple and completely described by Algorithm~\ref{alg:encoder}. It has only one global hyperparameter $\Delta$ which needs to be adjusted until the desired trade-off between total bitrate budget and model performance is met. The choice of this hyperparameter is relatively predictable, due to the consistency across tasks and optimizers. A similar relationship between quantization step size and the Lagrange parameter is exploited in contemporary video compression methods, where quantization must be controlled across video frames~\citep{SuWi98}.

\section{Experimental Setup}
\label{exp-setup}
In order to inform our design choices above, we perform empirical evaluations of Algorithm \ref{alg:fed_opt} on a variety of federated tasks drawn from benchmarks in \citep{reddi2021adaptive}.\footnote{Code available at: \url{https://github.com/google-research/federated/tree/1b31b84/compressed_communication}}

\textbf{Datasets, Models, and Tasks. }We use three datasets: CIFAR-100~\citep{krizhevsky2009learning}, EMNIST~\citep{cohen2017emnist}, and Stack Overflow~\citep{stackoverflow}. For CIFAR-100, we use the client partition proposed by \citet{reddi2021adaptive}. The other two datasets have natural client partitions where each client is an author (of handwritten digits and forum posts). For CIFAR-100, we train a ResNet-18, replacing batch normalization with group normalization (see \citep{hsieh2019non}). For EMNIST, we train a network with two convolutional layers, max-pooling, dropout, and two dense layers. For Stack Overflow, we perform next-word prediction (NWP) using an RNN with a single LSTM layer, and tag prediction (TP) using a multi-class logistic regression model. For full details, see Appendix \ref{appendix:models_and_datasets}.

\textbf{Algorithms. }We focus on two special cases of Algorithm \ref{alg:fed_opt}: FedAvg~\citep{pmlr-v54-mcmahan17a} and FedAdam~\citep{reddi2021adaptive}. In both, \localtrain is $E$ epochs of mini-batch SGD with client learning rate $\eta_c$. For FedAvg and FedAdam, \serverupdate is SGD or Adam (respectively) with server learning rate $\eta_s$.
We use FedAvg and FedAdam on the vision tasks (CIFAR-100 and EMNIST), but only FedAdam on the language tasks (Stack Overflow), as FedAvg performs poorly there~\citep{reddi2021adaptive}. We set $E = 1$ and use a batch size of 32 throughout. We perform $T = 1500$ rounds of training for each task. At each round, we sample $m = 50$ clients uniformly at random. We tune $\eta_c, \eta_s$ over $\{10^{-3}, 10^{-2}, \dots, 10\}$ by selecting the values that minimize the average validation loss over 5 random trials.

\textbf{Other Benchmarks. }We evaluate Algorithm \ref{alg:fed_opt} with our compression method and existing compression methods on the tasks described above (results in Figure~\ref{fig:so-good}). As a baseline, we include runs with \nocompression, where $\left(\mathcal{E}, \mathcal{D}\right)$ are no-ops and clients communicate their weighted updates at 32-bit precision. The accuracy achieved with \nocompression can be understood as the target accuracy we aim to reach with compression. We also compare to \topk~\citep{aji2017sparse}, \drive~\citep{vargaftik2021drive}, \threelc~\citep{Lim2018-rh}, and \qsgd~\citep{alistarh2017qsgd}. For details on these methods, see Appendix \ref{appendix:algorithm_benchmarks}.

Notable methods we do not compare with include \fedsketch \citep{haddadpour2020fedsketch}, \onebitsgd \citep{seide20141}, and \terngrad \citep{wen2017terngrad}. The first two use error-correction mechanisms which do not comply with our stateless requirement, while the latter two are often outperformed by \qsgd, and can suffer in performance when client updates are unevenly distributed~\citep{xu2020compressed} as is often the case in FL.

\section{Discussion}
\begin{figure}
\centering
\includegraphics[height=.36\linewidth]{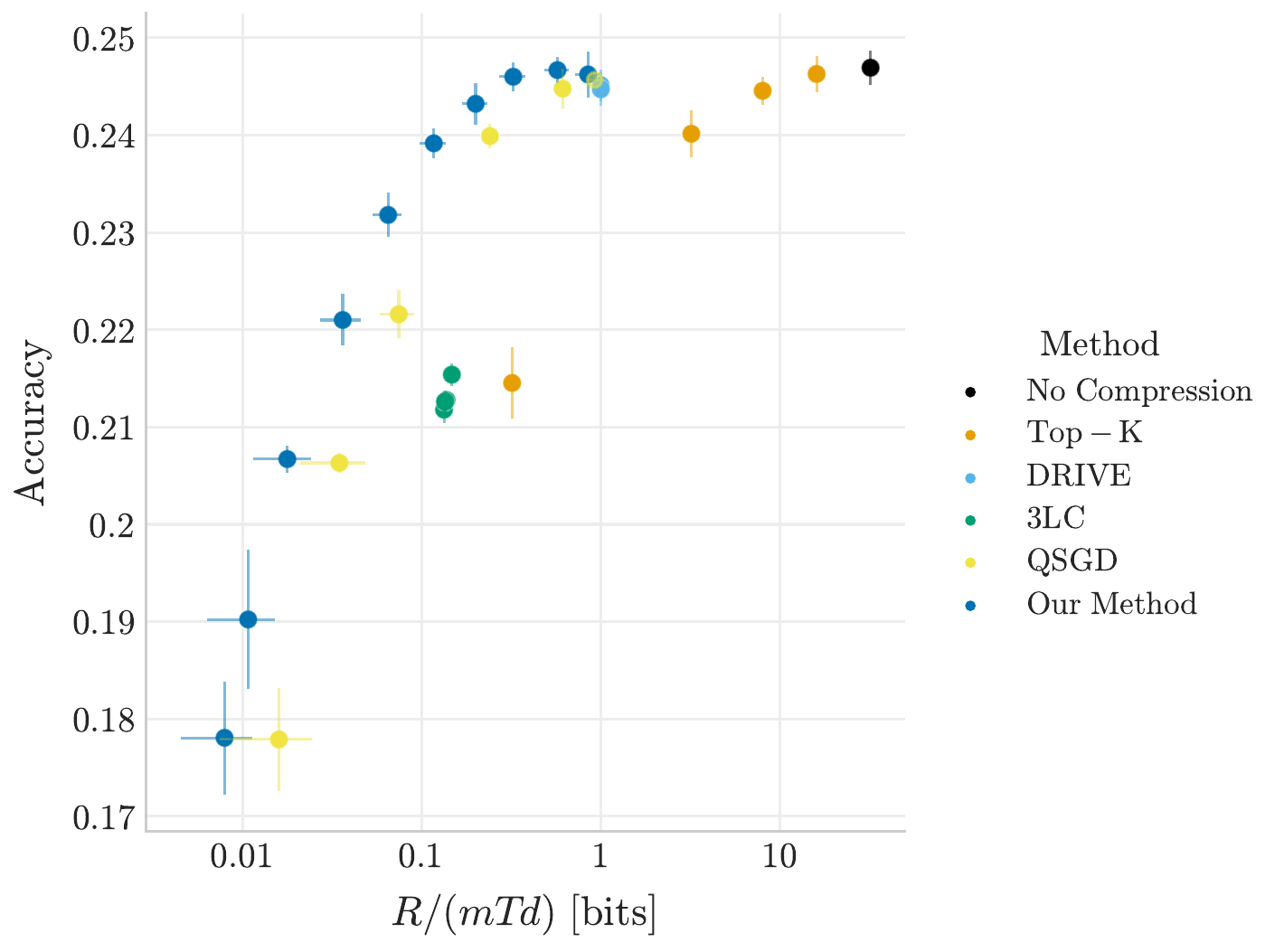}
\vspace{-0.2cm}
\caption{\label{fig:so-good}
Final model accuracy versus per-coordinate bit rate for a variety of methods on Stack Overflow NWP. Error bars indicate variance in average per-coordinate rate and final model accuracy over 5 random trials.
}
\end{figure}
Compared to the benchmarks above, our method is highly competitive. In most cases our method outperforms all others in terms of the accuracy--rate trade-off, as demonstrated in Figure~\ref{fig:so-good}. For those few cases in which our method does not raise the accuracy--communication cost frontier, we match the performance of the best existing methods to which we compare. Plots of overall performance on the other tasks are included in Appendix~\ref{appendix:overall-results}.

The poor performance of \threelc can be attributed to the lack of error correction when applied in the stateless setting. \topk's accuracy is highly dependent on the sparsity of client updates, and with a fixed dimension it does not achieve impressive compression ratios. \drive and \qsgd are most competitive with our method. The key distinguishing feature of \drive is their use of random rotations. For \qsgd the key differentiating factor is normalization, which leads to a different effective $\Delta$ per update. We focus on each of these design choices in the ablation studies that follow.
 
\textbf{Ablation: Random Rotations. }Random rotations have been used in fixed-rate federated compression methods by \citet{konevcny2016federated}, \citet{suresh2017distributed}, and \citet{vargaftik2021drive}. \citet{suresh2017distributed} motivate using fixed-rate compression for compatibility with secure aggregation \citep{bonawitz2017practical} and \citet{vargaftik2021drive} argue that variable-length encodings are expensive to compute. While using a fixed-bit code in combination with a uniform quantizer on a heavy-tailed distribution is highly inefficient from an entropy coding standpoint, applying a random rotation to the input distribution before quantization will ameliorate this effect, as it will ``Gaussianize'' the distribution.

\begin{figure}
\centering
\begin{minipage}{0.48\textwidth}
\centering
\includegraphics[height=.75\linewidth]{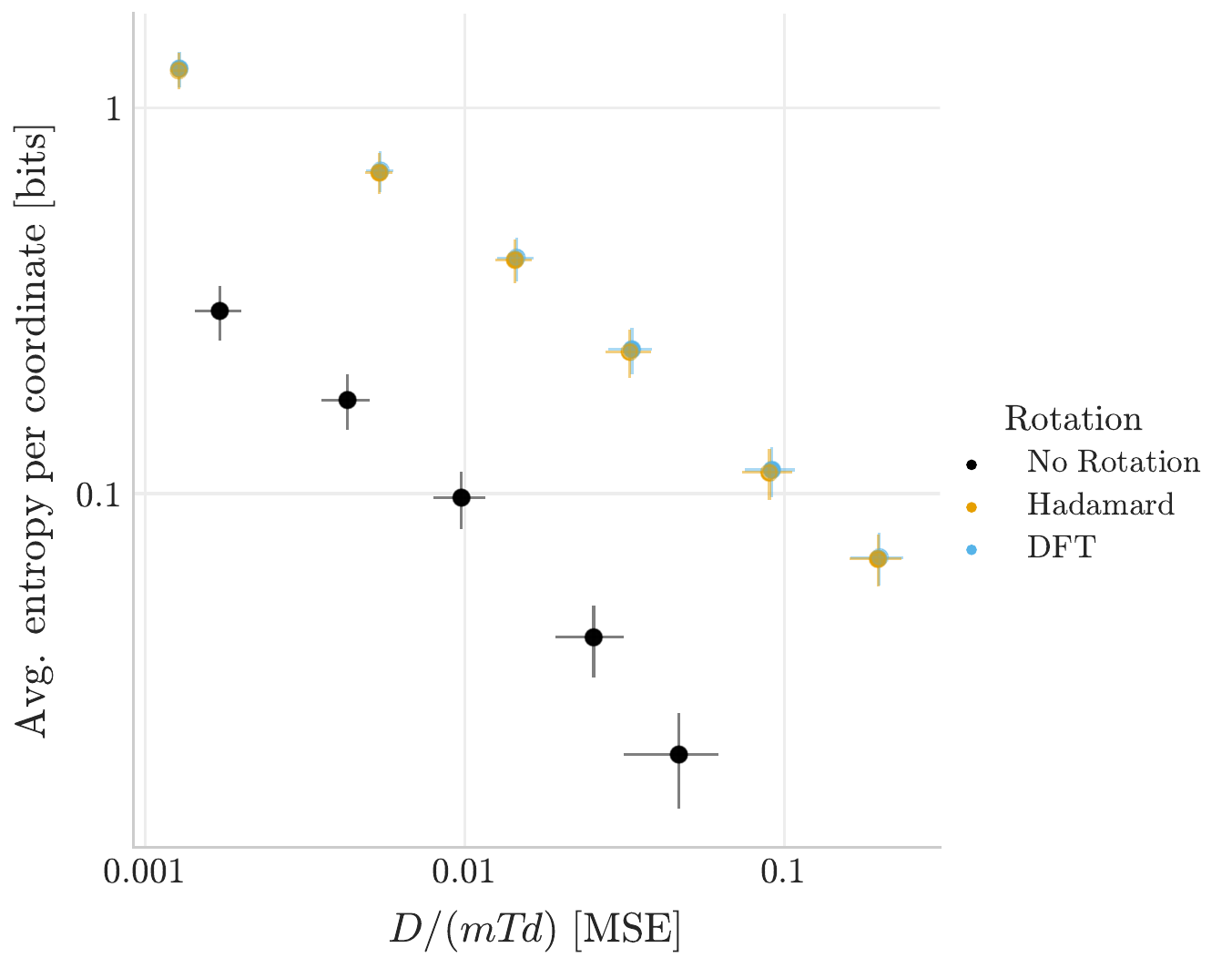}
\vspace{-0.2cm}
\caption{\label{so-fedadam-rotations}
Average per-coordinate entropy versus distortion of client updates in Stack Overflow NWP, after a random rotation (or no rotation) is applied. Error bars indicate the variance across 5 random trials.
}
\end{minipage}%
\hfill
\begin{minipage}{0.48\textwidth}
\centering
\includegraphics[height=.75\linewidth]{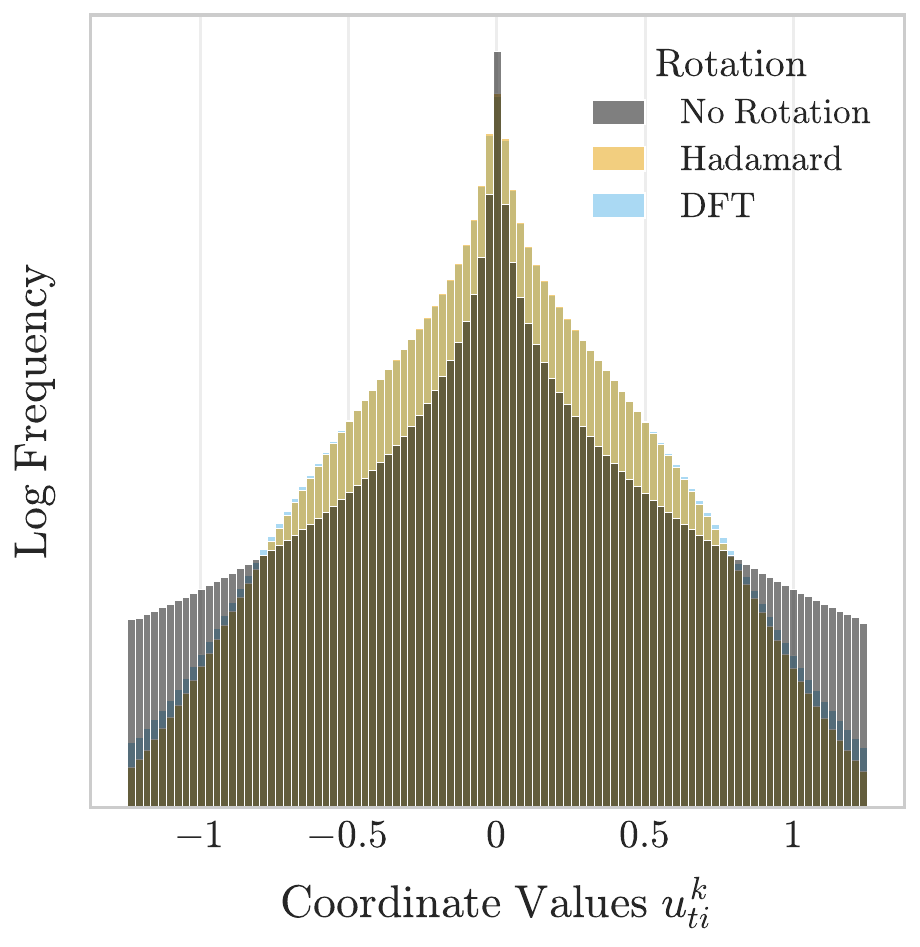}
\vspace{-0.2cm}
\caption{\label{rotation-histogram}
Histogram of client update coordinates in Stack Overflow NWP after applying a fixed, randomly sampled rotation (or no rotation). Frequencies are averaged across rounds.
}
\end{minipage}
\end{figure}

This can hide the existing structure of the model updates; it increases the per-coordinate entropy, and hence the expected bitrate (recall that a Gaussian is the max-entropy distribution for a given variance). Figure~\ref{so-fedadam-rotations} shows that applying a random Hadamard or DFT rotation before quantization results in a worse entropy--distortion frontier than if quantization is applied in the original coordinate space. Even using a random rotation fixed throughout training produces higher entropy marginal distributions, indicating that $u_t^k$ is already in a suitable coordinate system for compression (Figure~\ref{rotation-histogram}).

As we show in Appendix~\ref{appendix:rotations}, for Stack Overflow and EMNIST tasks, the entropy--distortion frontier is significantly better in the original space than after applying random rotations. The frontier is only slightly better for the CIFAR-100 task. This intuitively makes sense, as the CIFAR-100 client updates are not as sparse as in other tasks; the sparsity is observed primarily in the output, as nearly all clients only have a fraction of the possible image labels.

\textbf{Ablation: Normalization. }Compression methods designed for DT such as \qsgd or \drive \citep{alistarh2017qsgd, vargaftik2021drive} often scale each client update to have the same vector magnitude $\|u_t^k\|$ before applying compression. This is equivalent to choosing a magnitude-dependent per-client $\Delta_t^k$. However, as we show in Section~\ref{rd-theory}, each client should quantize their update to the same fidelity, since other choices lead to an inferior global rate--distortion trade-off. This is the key differentiator between \qsgd and our method.

\begin{figure}
\centering
\begin{minipage}{0.48\textwidth}
\centering
\includegraphics[height=.75\linewidth]{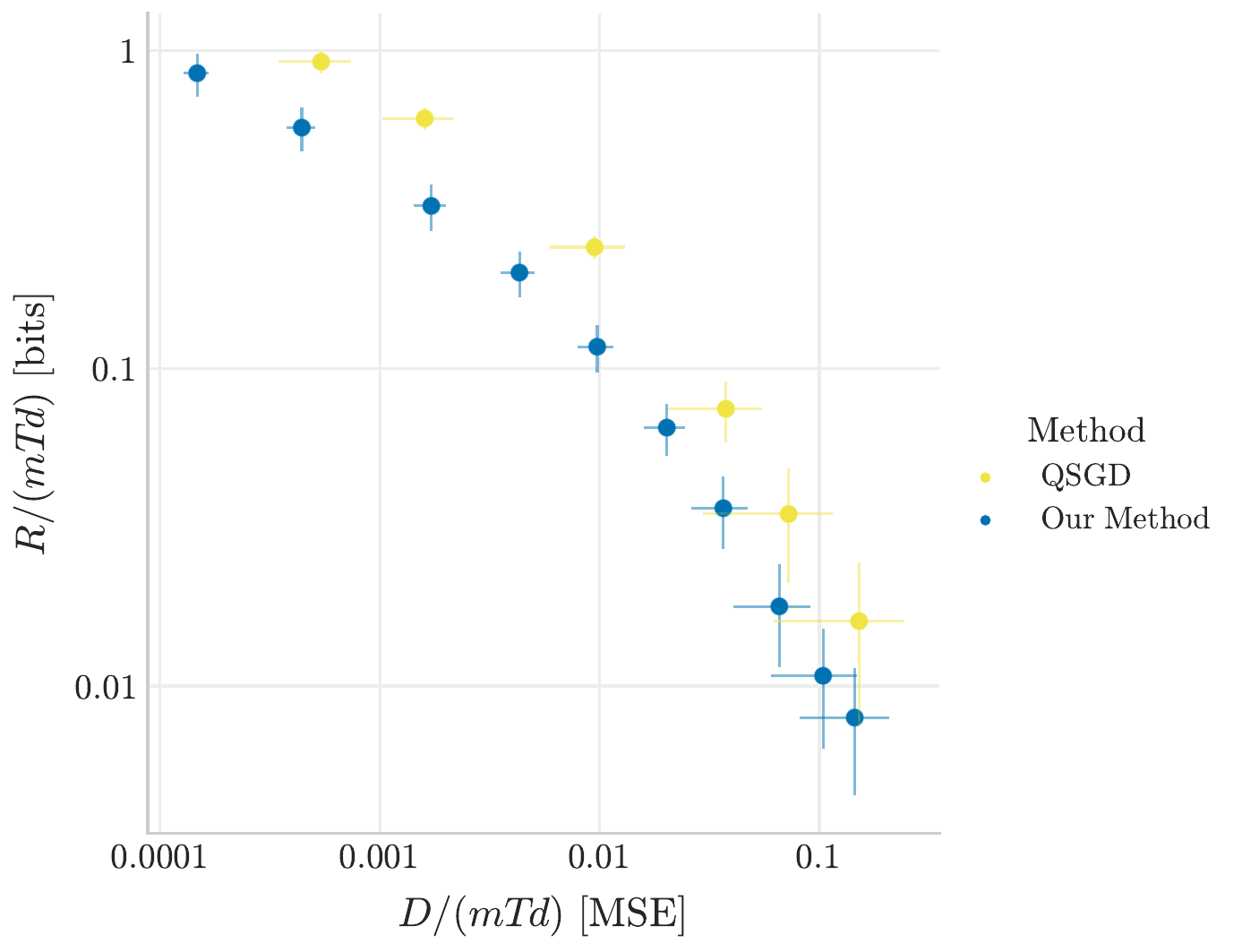}
\vspace{-0.2cm}
\caption{\label{stackoverflow_word-fedadam-normalization}
Average per-coordinate rate versus distortion of client updates in Stack Overflow NWP for QSGD and our method. Error bars indicate variance over 5 random trials.
}
\end{minipage}%
\hfill
\begin{minipage}{0.48\textwidth}
\centering
\includegraphics[height=.75\linewidth]{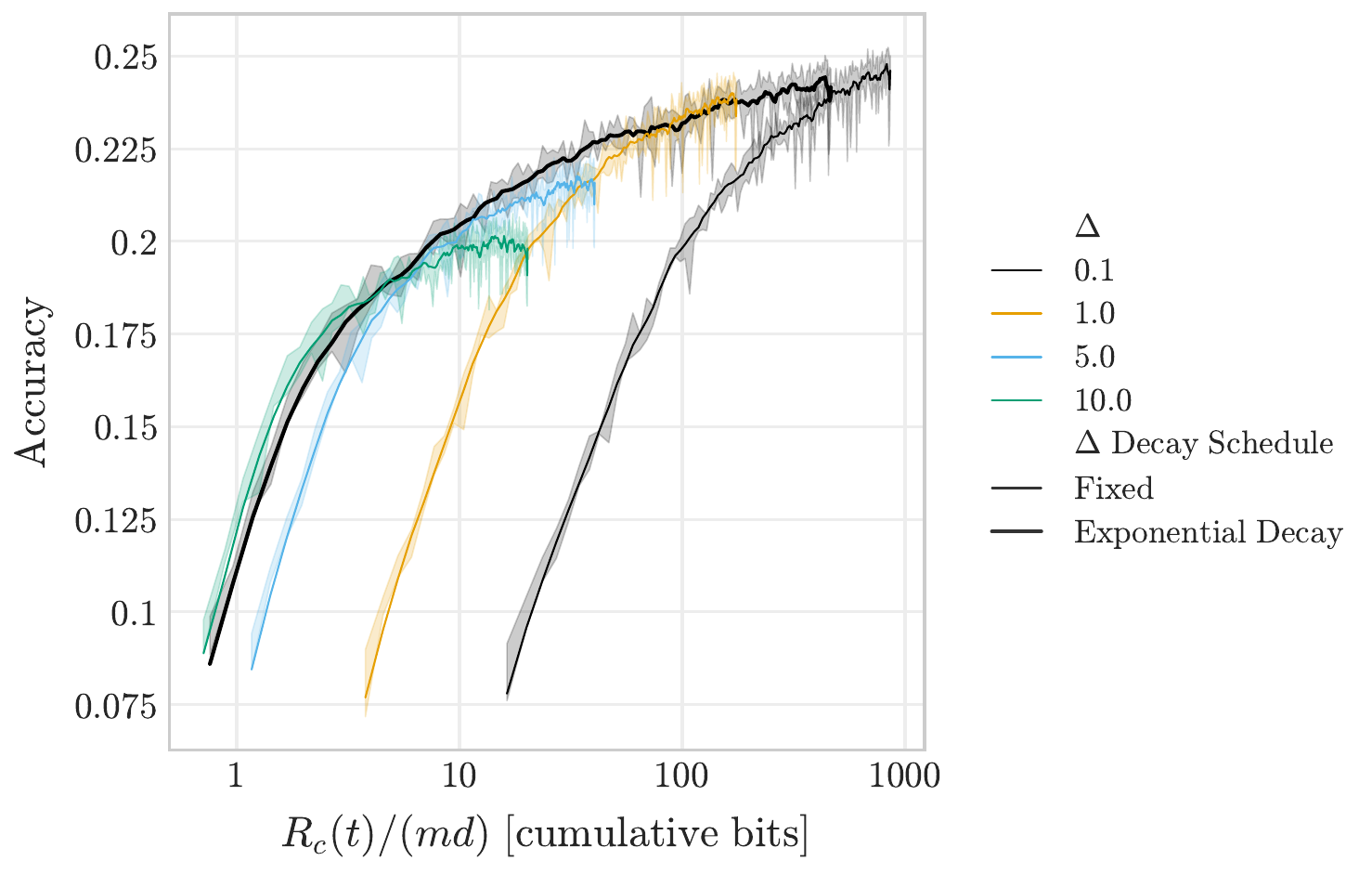}
\vspace{-0.2cm}
\caption{\label{stackoverflow_word-fedadam-delta_decay}
Validation accuracy versus cumulative bitrate for our method with fixed $\Delta$ and exponentially decayed $\Delta$ on Stack Overflow NWP.
}
\end{minipage}
\end{figure}

For optimization in heterogeneous settings \citep{li2020federated}, the update magnitudes can vary dramatically. Comparing the $R$--$D$ curves for our method and \qsgd, we find that our method provides a better rate--distortion performance, particularly on tasks with significant client heterogeneity (Figure~\ref{stackoverflow_word-fedadam-normalization}). When client updates are more homogeneous (as in the CIFAR-100 dataset, where each client has the same number of examples~\citep{reddi2021adaptive}), gains are, as expected, negligible. The normalization strategy of QSGD thus may be the reason why our method performs better.

\section{Conclusions and Outlook}\label{sec:conclusion}
As shown earlier, the present method shows excellent performance over a wide range of tasks and optimizers, outperforming \topk, \drive, \threelc and \qsgd. We accomplish this with a simple stateless algorithm that produces a variable-length encoding, tailored to the empirical distribution of client updates, and which only has one hyperparameter, necessary to control the global trade-off between rate and final accuracy.

The method closely resembles QSGD, which was developed for DT rather than FL. However, our method was developed independently, and we present a data-driven justification of each design choice we made, ensuring its appropriateness in the cross-device FL setting. To the best of our knowledge, this is also the first work to analyze an FL method using $R$--$D$ theory, which turns out to improve the global performance significantly over \qsgd, simply by controlling $\Delta$ appropriately. As far as we can tell, our work also represents the most comprehensive empirical study of the role and benefits of compression techniques in FL, and provides a rigorous set of baselines to support future work.

Our ablations demonstrate that A) popular preprocessing by a random rotation can hurt average-case performance, as it hides the statistical regularities that could be exploited for compression, and \emph{increases} the entropy of the signal to be communicated; and B) normalizing each client's update, a technique common in DT compression methods, is sub-optimal in non-i.i.d. FL settings.

An additional improvement over the results reported above is possible noting that higher levels of distortion are less problematic in earlier rounds, and more so later on, as the training gets closer to convergence. Formally, this is analogous to the fact that for SGD, the variance of the stochastic gradient determines the convergence radius of its iterates~\citep[Theorem 1]{alistarh2017qsgd}. Thus, rather than keeping the $R$--$D$ trade-off fixed over the course of training, we could replace the sum of distortions in \eqref{eq:distortion_proxy} by a weighted sum, varying the trade-off over time. As the relationship between $\Delta$ and $\lambda$ discussed in Section~\ref{rd-theory} holds across rounds, one simple way to explore this idea is to decay $\Delta$.

We applied an exponential decay schedule on $\Delta$ (details in Appendix~\ref{appendix:delta_decay}; one result in Figure~\ref{stackoverflow_word-fedadam-delta_decay}). Training with large $\Delta$ is communication-efficient at first, but accuracy saturates quickly. Training with small $\Delta$ reaches top accuracy, but at the expense of communication. With $\Delta$ decay, top accuracy is reached with overall fewer bits transmitted. We found comparable results across other tasks and optimizers (Appendix~\ref{appendix:delta_decay}), but did not tune the schedules. Exploring time-variability in more depth is a topic for future work, including possibly more sophisticated ways of modulating the $R$--$D$ trade-off over time, such as adapting to running training metrics. Another topic of future work is to explore compatibility with privacy protocols.

% \begin{ack}

% \end{ack}

% \bibsection
% \section*{References}
% \bibliography{main}

\bibliographystyle{plainnat}
\bibliography{references}

\appendix
\onecolumn

\section{Full Experimental Details}\label{appendix:experiment_details}

\subsection{Datasets, Tasks and Models}\label{appendix:models_and_datasets}

We use three datasets throughout our work: CIFAR-100~\citep{krizhevsky2009learning}, the federated extended MNIST dataset (EMNIST)~\citep{cohen2017emnist}, and the Stack Overflow dataset~\citep{stackoverflow}. The first two datasets are image datasets, the last is a language dataset. All datasets are publicly available: CIFAR-100 is published by the authors, EMNIST dataset is covered under the blanket license of Standard Reference Data by NIST, Stack Overflow is licensed under the Creative Commons Attribution-ShareAlike 3.0 Unported License. We specifically use the versions available in TensorFlow Federated~\citep{tff}, which gives a federated structure to all three datasets.

For a summary of the dataset statistics, tasks, and models used, see Table~\ref{table:datasets_tasks_models}. We discuss these in more detail below.

\begin{table}[th]
\caption{Datasets, Tasks \& Models}
\label{table:datasets_tasks_models}
\vskip 0.15in
\begin{center}
\begin{small}
\begin{sc}
\begin{tabular}{lcccccc}
\toprule
Dataset & \multicolumn{2}{c}{Num Clients} & \multicolumn{2}{c}{Num Examples} & Task & Model
\\\cmidrule(lr){2-3}\cmidrule(lr){4-5}
        & Train & Test                     & Train & Test & & \\
\midrule
EMNIST          & 3,400 & 3,400 & 671,585 & 77,483   & \begin{tabular}{@{}c@{}}Character \\ Recognition\end{tabular} & CNN\\
\hline \\
\begin{tabular}{@{}c@{}}Stack \\ Overflow\end{tabular} & 342,477 & 204,088 & 135.8M & 16.6M & \begin{tabular}{@{}c@{}}Next-Word \\ Prediction\end{tabular} & LSTM\\
\hline \\
\begin{tabular}{@{}c@{}}Stack \\ Overflow\end{tabular}   & 342,477 & 204,088 & 135.8M & 16.6M & \begin{tabular}{@{}c@{}}Tag \\ Prediction\end{tabular} & \begin{tabular}{@{}c@{}}Logistic \\ Regression\end{tabular}\\
\hline \\
CIFAR-100       & 500 & 100 & 50,000 & 10,000        & \begin{tabular}{@{}c@{}}Image \\ Recognition\end{tabular} & \begin{tabular}{@{}c@{}}ResNet-18 \\ with GroupNorm\end{tabular}\\
\bottomrule
\end{tabular}
\end{sc}
\end{small}
\end{center}
\end{table}

\paragraph{CIFAR-100} The CIFAR-100 dataset is a vision dataset consisting of $32 \times 32 \times 3$ images with 100 possible labels. This dataset does not have a canonical partitioning among clients. However, an artificial partitioning among clients was created by \citet{reddi2021adaptive} using hierarchical latent Dirichlet allocation to obtain moderate amounts of heterogeneity among clients. This partitioning is based on Pachinko allocation~\citep{li2006pachinko}, and is available in TensorFlow Federated~\citep{cifar100federated}. Under this partitioning, each client typically has only a subset of the 100 possible labels. The dataset has 500 training clients and 100 test clients, each with 100 examples in their local dataset.

We train a ResNet-18~\citep{he2016deep} on this dataset, where we replace all batch normalization layers with group normalization layers~\citep{wu2018group}, as group norm can perform better than batch norm in federated settings~\citep{hsieh2019non}. We use group normalization layers with two groups. We perform small amounts of data augmentation and preprocessing for each train and test sample. We first centrally crop each image $(24, 24, 3)$. We then normalize the pixel values according to their mean and standard deviation.

\paragraph{EMNIST} The EMNIST dataset contains $28\times 28$ gray-scale pixel images of hand-written alphanumeric characters. There are 62 alphanumeric characters in the dataset, and the characters are partitioned among clients according to their author. The dataset has 3,400 clients, who have both train and test datasets. The dataset has natural heterogeneity stemming from the writing style of each person. We train a convolutional network on the dataset (the same as in \citep{reddi2021adaptive}). The network uses two convolutional layers (each with $3\times 3$ kernels and strides of length 1), followed by a max pooling layer using dropout with $p = 0.25$, a dense layer with 128 units and dropout with $p = 0.5$, and a final dense output layer.

\paragraph{Stack Overflow} Stack Overflow is a language dataset consisting of question and answers from the Stack Overflow site. The questions and answers also have associated metadata, including tags. Each client corresponds to a user. The specific train/validation/test split from~\citep{stackoverflow} has 342,477 train clients, 38,758 validation clients, and 204,088 test clients. Notably, the train clients only have examples from before 2018-01-01 UTC, while the test clients only have examples from after 2018-01-01 UTC. The validation clients have examples with no date restrictions, and all validation examples are held-out from both the test and train sets. We perform two tasks on this dataset: next-word prediction and tag prediction. Both tasks are drawn from \citep{reddi2021adaptive} and use the same models there.

\emph{Next-Word Prediction:} We use padding and truncation to ensure that each sentence has 20 words. We then represent the sentence as a sequence of indices corresponding to the 10,000 most frequently used words, as well as indices representing padding, out-of-vocabulary words, the beginning of a sentence, and the end of a sentence. We perform next-word prediction on these sequences using an a recurrent neural network (RNN)~\citep{mikolov2010recurrent} that embeds each word in a sentence into a learned 96-dimensional space. It then feeds the embedded words into a single LSTM layer~\citep{hochreiter1997long} of hidden dimension 670, followed by a densely connected softmax output layer. The metric used in the main body is the accuracy over the 10,000-word vocabulary; it does not include padding, out-of-vocab, or beginning or end of sentence tokens when computing the accuracy.

\emph{Tag Prediction:} We represent each sentence in each client's local dataset as a bag-of-words vector, normalized to have sum 1, restricting to only the 10,000 most frequently used words in the vocabulary. The tags are represented as binary vectors indicating the presence of a tag in the associated meta-data. We restrict to the 500 most frequently occurring tags. We then use a multi-class logistic regression model for tag prediction, using a one-versus-rest classification strategy.

\subsection{Algorithmic Benchmarks}\label{appendix:algorithm_benchmarks}

We evaluate our compression method against existing approaches on the tasks described above. As a baseline, we include runs with \nocompression, where $\left(\mathcal{E}, \mathcal{D}\right)$ are no-ops and clients communicate their weighted updates at full-precision. This method has fixed rate at 32-bits per coordinate and no distortion. The accuracy achieved with \nocompression can be understood as the target accuracy we aim to reach with compression. We also compare with the following methods:

\begin{itemize}
    \item \topk \citep{aji2017sparse} is a fixed-dimension sparsification method, where only the $k\%$ largest magnitude coordinates within each client update are communicated along with a bitmask indicating their position within the tensor. We compare with $k\in\{1\%,10\%,25\%,50\%\}$.
    \item \drive \citep{vargaftik2021drive} is a fixed-rate method which is designed to provide worst-case guarantees that bound distortion for any input. \drive applies a random rotation, then quantizes each tensor coordinate to a single bit, its sign, and computes a scale factor $S$. We compare to \drive with a structured random rotation, either the Hadamard Transform or the Discrete Fourier Transform, and we use the unbiased the scale factor $S=||x||_2^2/||\mathcal{R}(x)||_1$.
    \item \threelc \citep{Lim2018-rh} is a variable-length compression method that combines 3-value quantization with sparsification and utilizes lossless coding to represent the quantized tensor components compactly. Though \threelc uses memory to compensate for errors, this approach is not realistic in FL and so we instead use stochastic quantization, as mentioned as an appropriate alternative by \citet{Lim2018-rh}. We compare with stochastic quantization \threelc using sparsity factor $s=\{1.00, 1.50, 1.755, 1.90\}$.
    \item \qsgd \citep{alistarh2017qsgd} is a variable-length compression method similar to our own, which combines stochastic quantization with Elias source coding. A key differentiator is that the authors parametrize their quantizer by the number of quantization levels $s$ rather than the size of each level, and they first \emph{normalize} each client's update and communicate its norm along with the quantized and encoded tensor, so that the tensor can be descaled after it is dequantized. We compare to \qsgd over a range of $s=\{16, 32, 64, 256, 1024, 2048\}$.
\end{itemize}

All code for implementing our method, implementing these benchmarks and launching experiments is available at: \url{https://github.com/google-research/federated/tree/1b31b84/compressed_communication}.

Other notable existing gradient compression methods include \onebitsgd \citep{seide20141}, \terngrad \citep{wen2017terngrad} and \fedsketch \citep{haddadpour2020fedsketch}. We omit a direct comparison with \onebitsgd as this method includes error correction and other techniques have been demonstrated to outperform it \citep{alistarh2017qsgd}. We also omit reference to \terngrad as it often behaves comparably to or worse than \qsgd in practice~\citep{xu2020compressed}, and is actually equivalent to \qsgd if the number of quantization levels $s$, is set to $\|g\|_2 / \|g\|_\infty$~\citep{wang2018atomo}. \terngrad's 3-value quantization also resembles \threelc, which yields a more compact representation \citep{Lim2018-rh}. \citet{xu2020compressed} note that \terngrad performs best when the gradient components are evenly distributed, which is not the case in our FL settings. We omit a direct comparison with \fedsketch as the error correction used in this method does not comply with our stateless requirement.

%$\Delta = \{0.05, 0.25, 0.1, 0.5, 1.0, 2.0, 3.755, 7.5, 12.5, 17.5\}$

\newpage
\section{Full Results}\label{appendix:full_results}

\subsection{Histograms}\label{appendix:histograms}

The same experiment as in Figure~\ref{so-fedadam-hist}, with other tasks and optimizers. We observe a relatively consistent statistical structure across all tasks: client updates tend to resemble a symmetric power law distribution with a spike at zero. We observe that the CIFAR-100 client updates are significantly less sparse than for the other tasks. Additionally, the Stack Overflow tag prediction client updates are more heavily negative; this asymmetry can be attributed to the infrequency of tags on each client when using many-output logistic regression.

\begin{figure}[H]
\begin{subfigure}{0.48\linewidth}
    \centering
    \includegraphics[height=.75\linewidth]{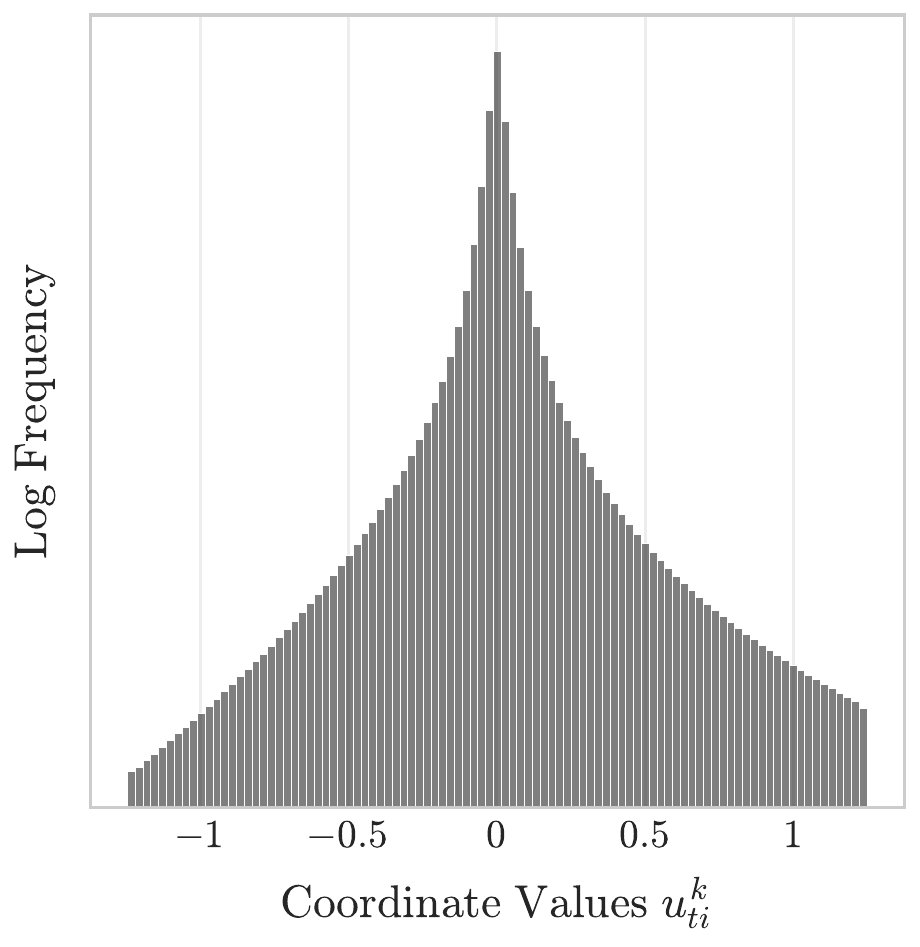}
    \caption{CIFAR-100, FedAdam}
\end{subfigure}
\begin{subfigure}{0.48\linewidth}
    \centering
    \includegraphics[height=.75\linewidth]{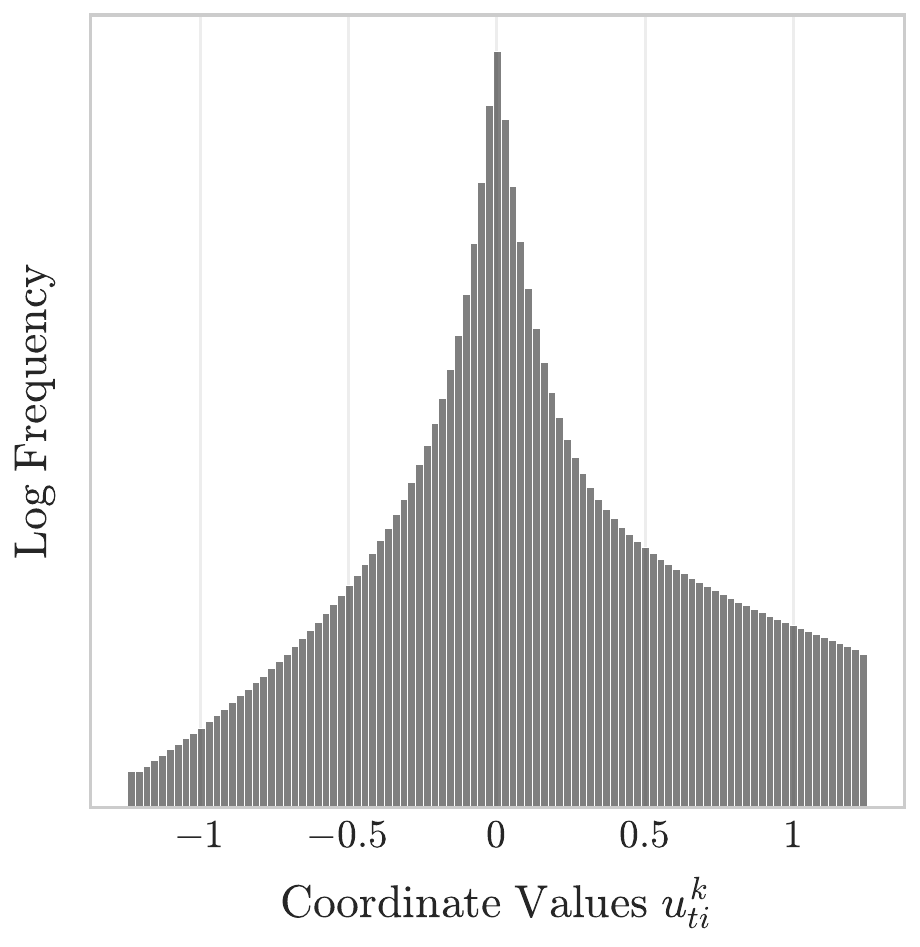}
    \caption{CIFAR-100, FedAvg}
\end{subfigure}
\begin{subfigure}{0.48\linewidth}
    \centering
    \includegraphics[height=.75\linewidth]{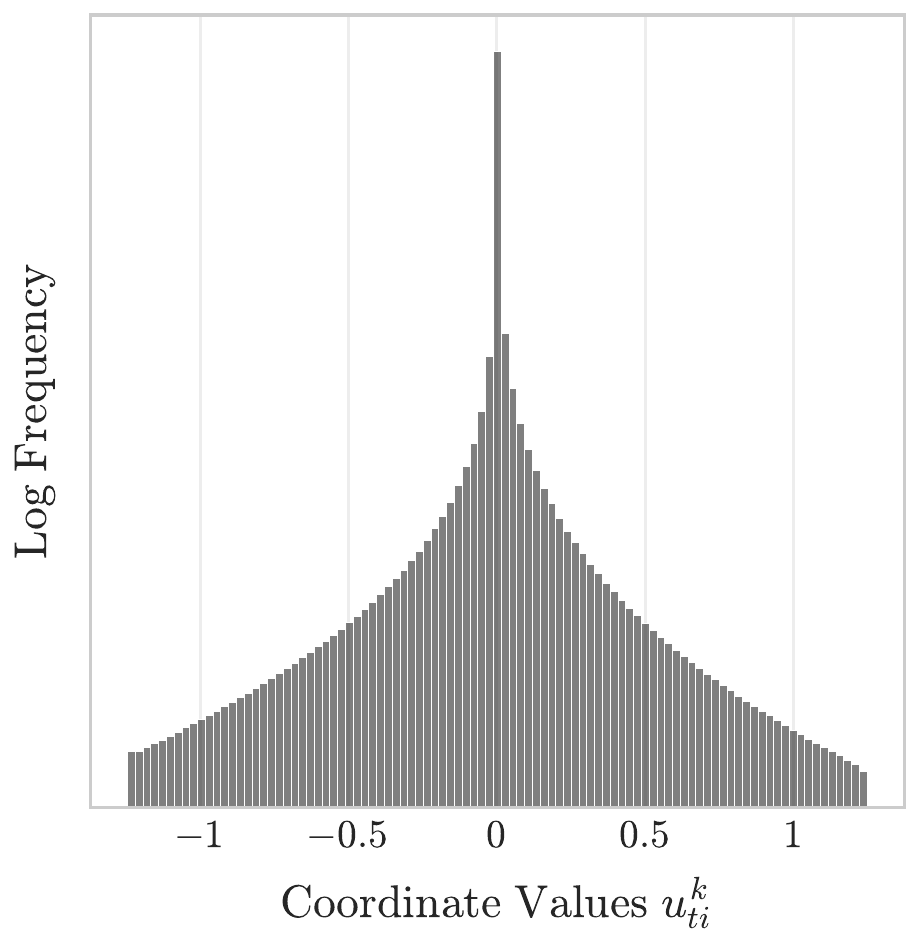}
    \caption{EMNIST, FedAdam}
\end{subfigure}
\centering
\begin{subfigure}{0.48\linewidth}
    \centering
    \includegraphics[height=.75\linewidth]{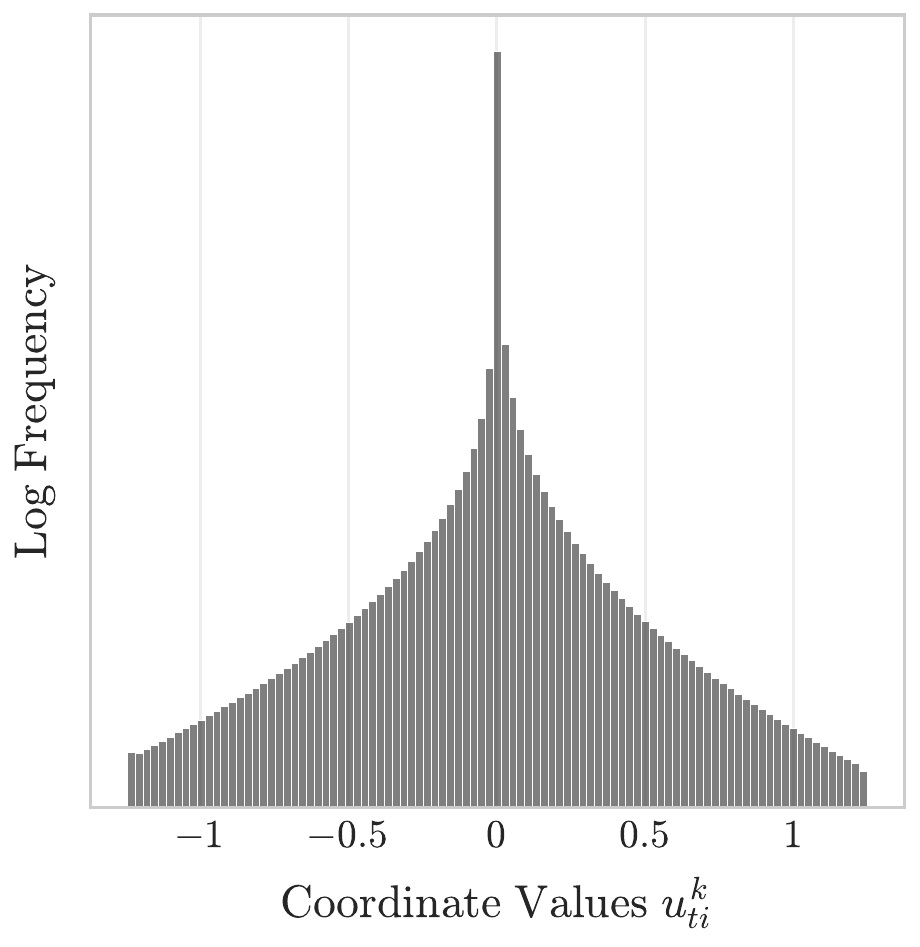}
    \caption{EMNIST, FedAvg}
\end{subfigure}
\begin{subfigure}{0.48\linewidth}
    \centering
    \includegraphics[height=.75\linewidth]{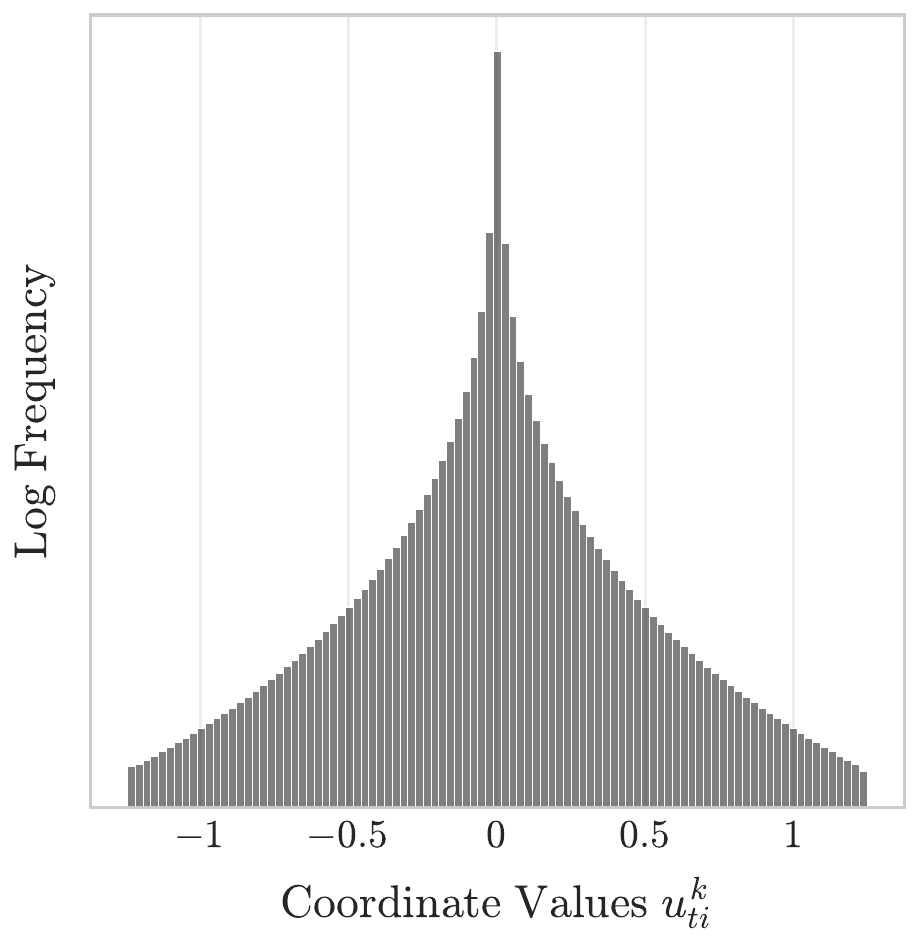}
    \caption{Stack Overflow NWP, FedAdam}
\end{subfigure}
\centering
\begin{subfigure}{0.48\linewidth}
    \centering
    \includegraphics[height=.75\linewidth]{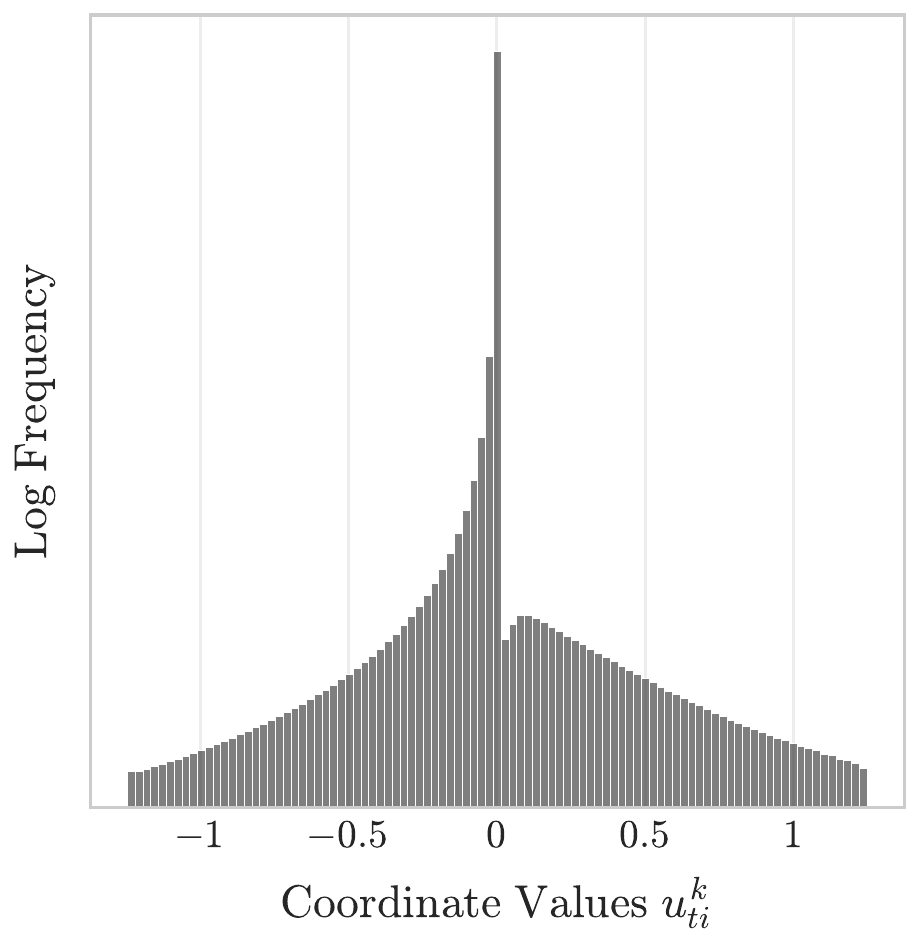}
    \caption{Stack Overflow TP, FedAdam}
\end{subfigure}
\caption{The histograms of weighted client updates averaged over the course of training across tasks and optimizers tend to resemble a symmetric power law distribution with a spike at zero. Notably, the CIFAR-100 client updates are significantly less sparse. The Stack Overflow TP client updates are more heavily negative due to the infrequency of tags on each client.}
\end{figure}

\newpage
\subsection{Rounding Methods}\label{appendix:rounding}

The same experiment as in Figure~\ref{so-acc}, with other tasks and optimizers. We find that regardless of task and optimizer, across quantization step sizes $\Delta$, \stochasticround performs best.

\begin{figure}[H]
\centering
\begin{subfigure}{\linewidth}
    \centering
    \includegraphics[width=\linewidth]{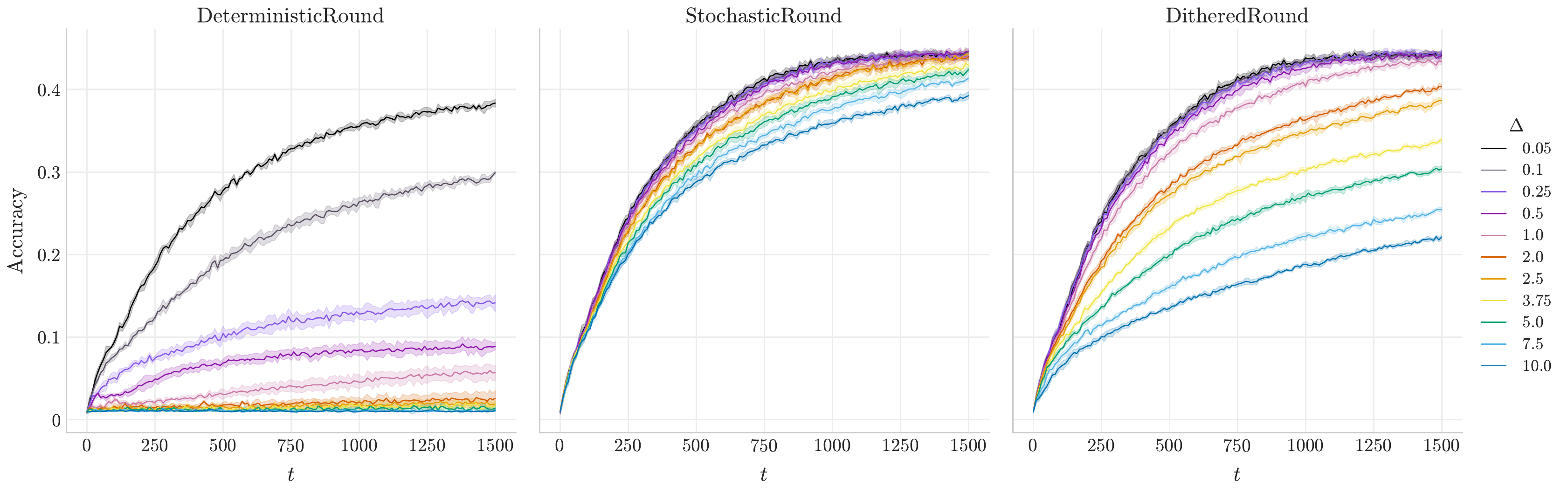}
    \caption{CIFAR-100, FedAdam}
\end{subfigure}
\centering
\begin{subfigure}{\linewidth}
    \centering
    \includegraphics[width=\linewidth]{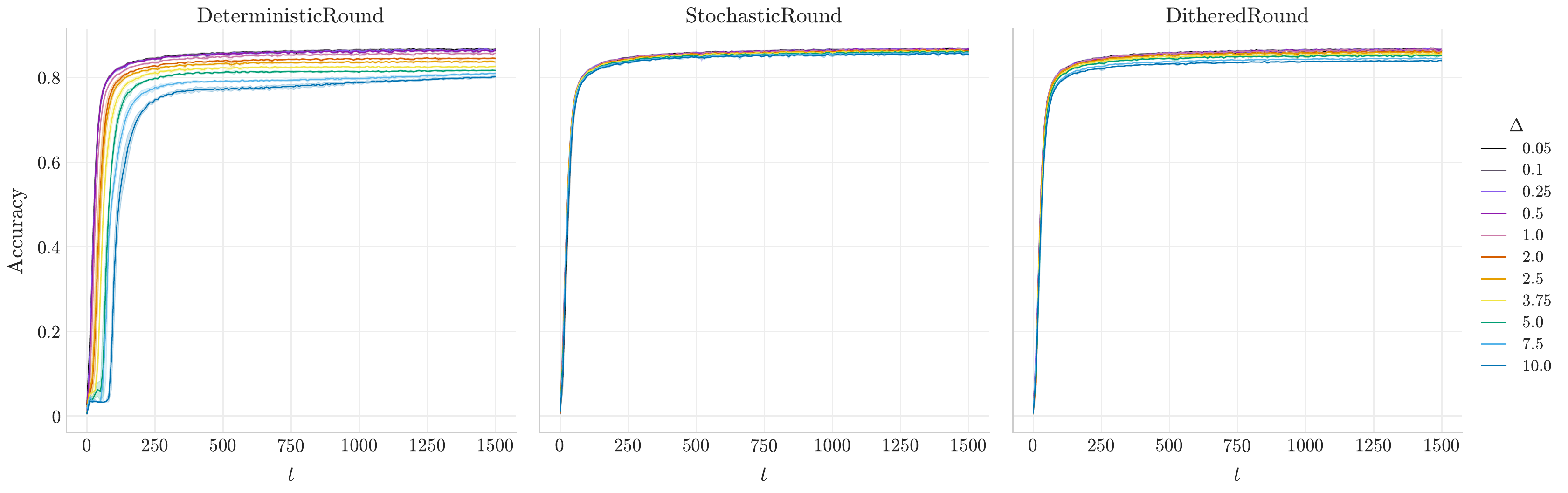}
    \caption{EMNIST, FedAdam}
\end{subfigure}
\centering
\begin{subfigure}{\linewidth}
    \centering
    \includegraphics[width=\linewidth]{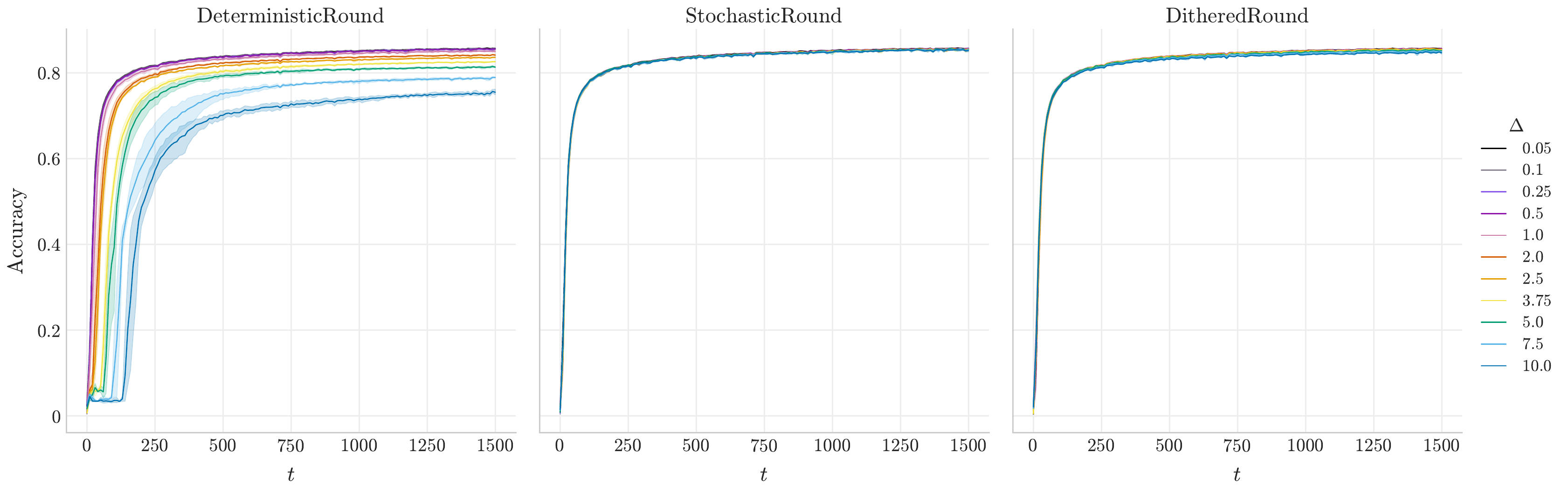}
    \caption{EMNIST, FedAvg}
\end{subfigure}
\caption{Accuracy over the course of training across quantization step sizes and rounding methods. We find \stochasticround to perform best across tasks.}
\end{figure}
\begin{figure}[H]\ContinuedFloat
\centering
\begin{subfigure}{\linewidth}
    \centering
    \includegraphics[width=\linewidth]{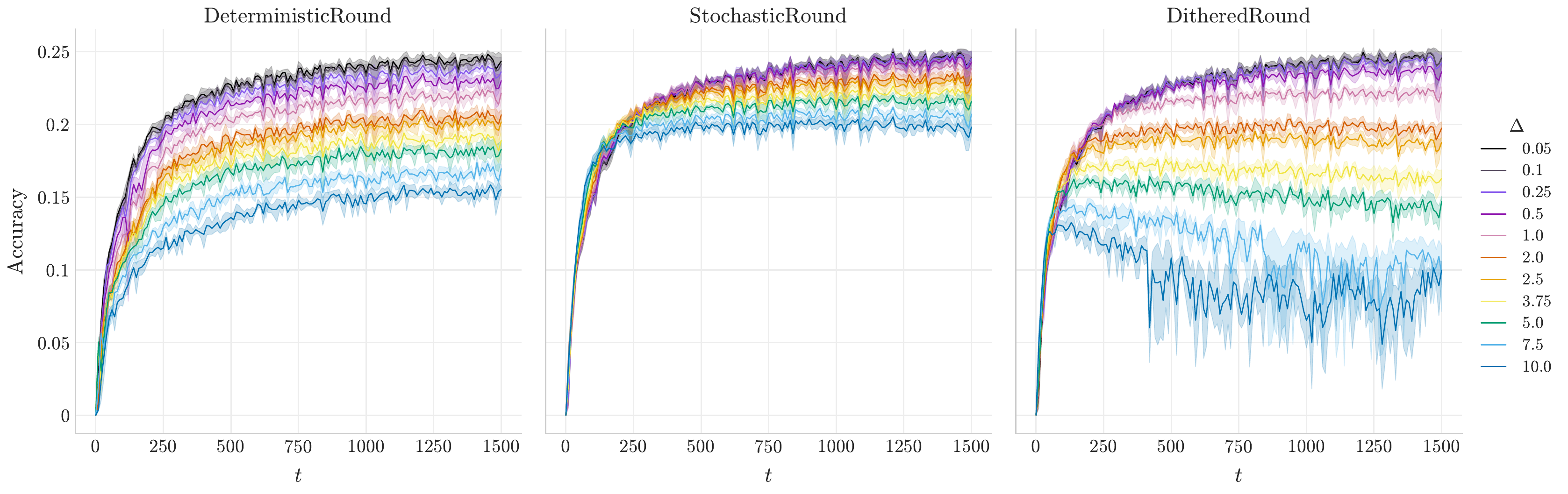}
    \caption{Stack Overflow NWP, FedAdam}
\end{subfigure}
\centering
\begin{subfigure}{\linewidth}
    \centering
    \includegraphics[width=\linewidth]{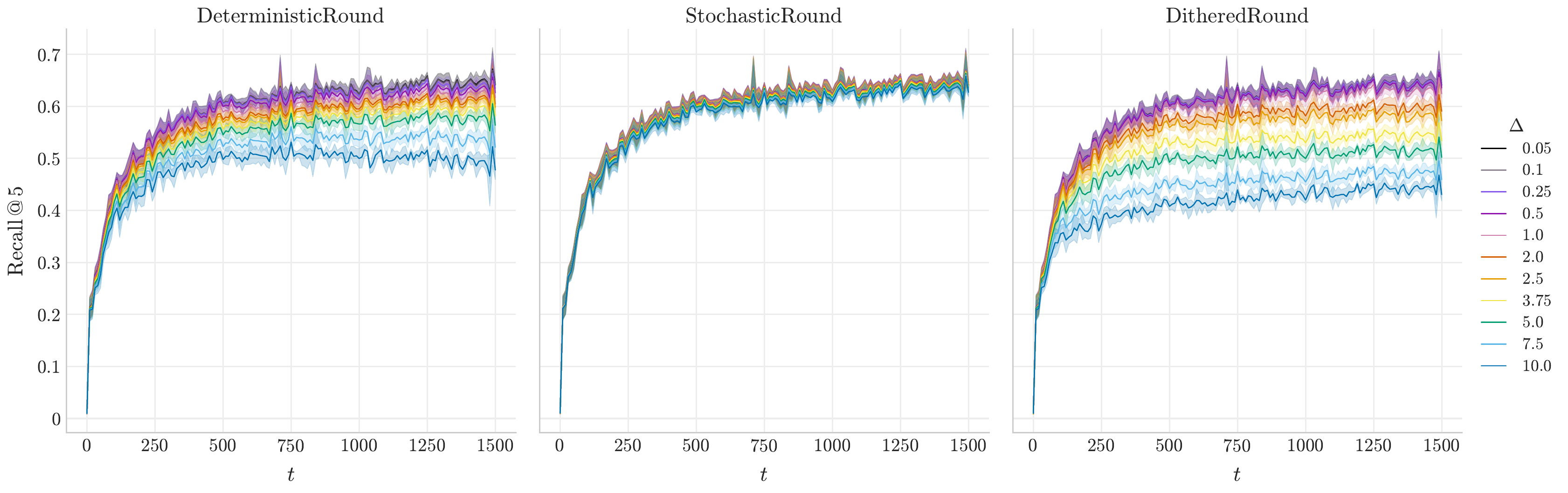}
    \caption{Stack Overflow TP, FedAdam}
\end{subfigure}
\caption{Accuracy over the course of training across quantization step sizes and rounding methods. We find \stochasticround to perform best across tasks. (cont.)}
\end{figure}

\newpage
\subsection{Sparsity}\label{appendix:sparsity}

The same experiment as in Figure~\ref{so-sparsity}, with other tasks and optimizers. We observe relatively high levels of sparsity across all tasks. However, the CIFAR-100 client updates are significantly less sparse than for the other tasks.

\begin{figure}[H]
\begin{subfigure}{0.48\linewidth}
    \centering
    \includegraphics[height=.75\linewidth]{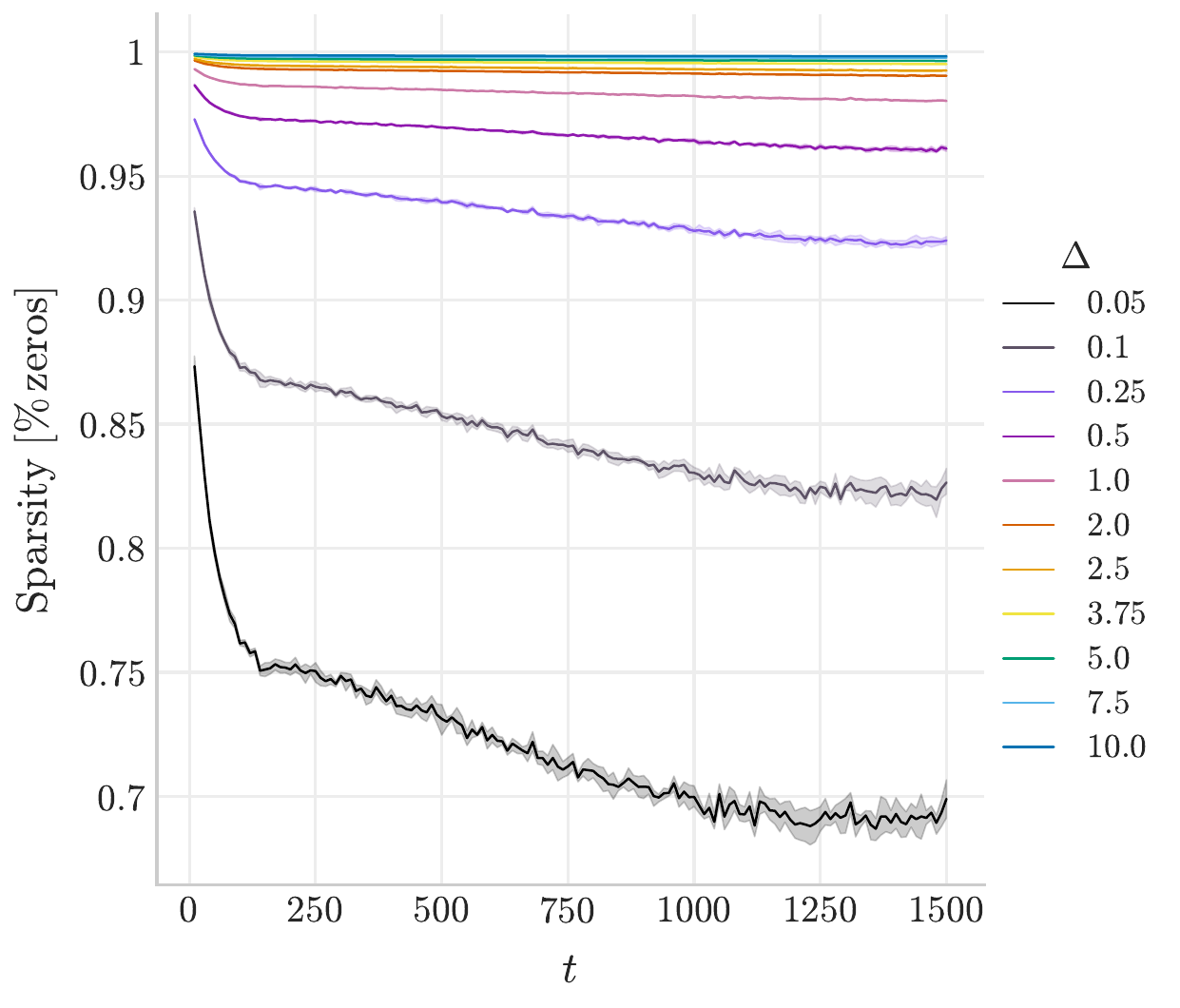}
    \caption{CIFAR-100, FedAdam}
\end{subfigure}
\begin{subfigure}{0.48\linewidth}
    \centering
    \includegraphics[height=.75\linewidth]{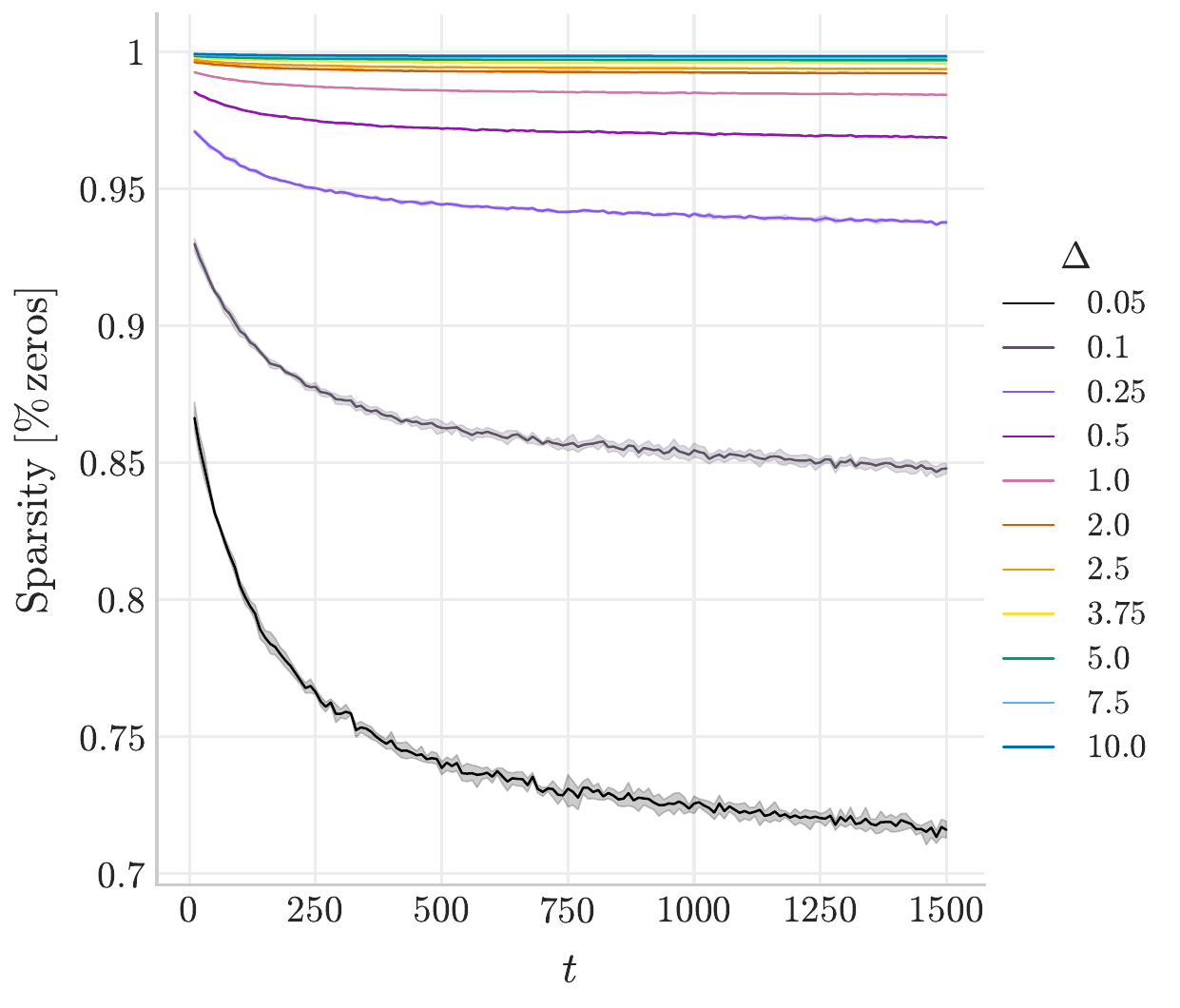}
    \caption{CIFAR-100, FedAvg}
\end{subfigure}
\begin{subfigure}{0.48\linewidth}
    \centering
    \includegraphics[height=.75\linewidth]{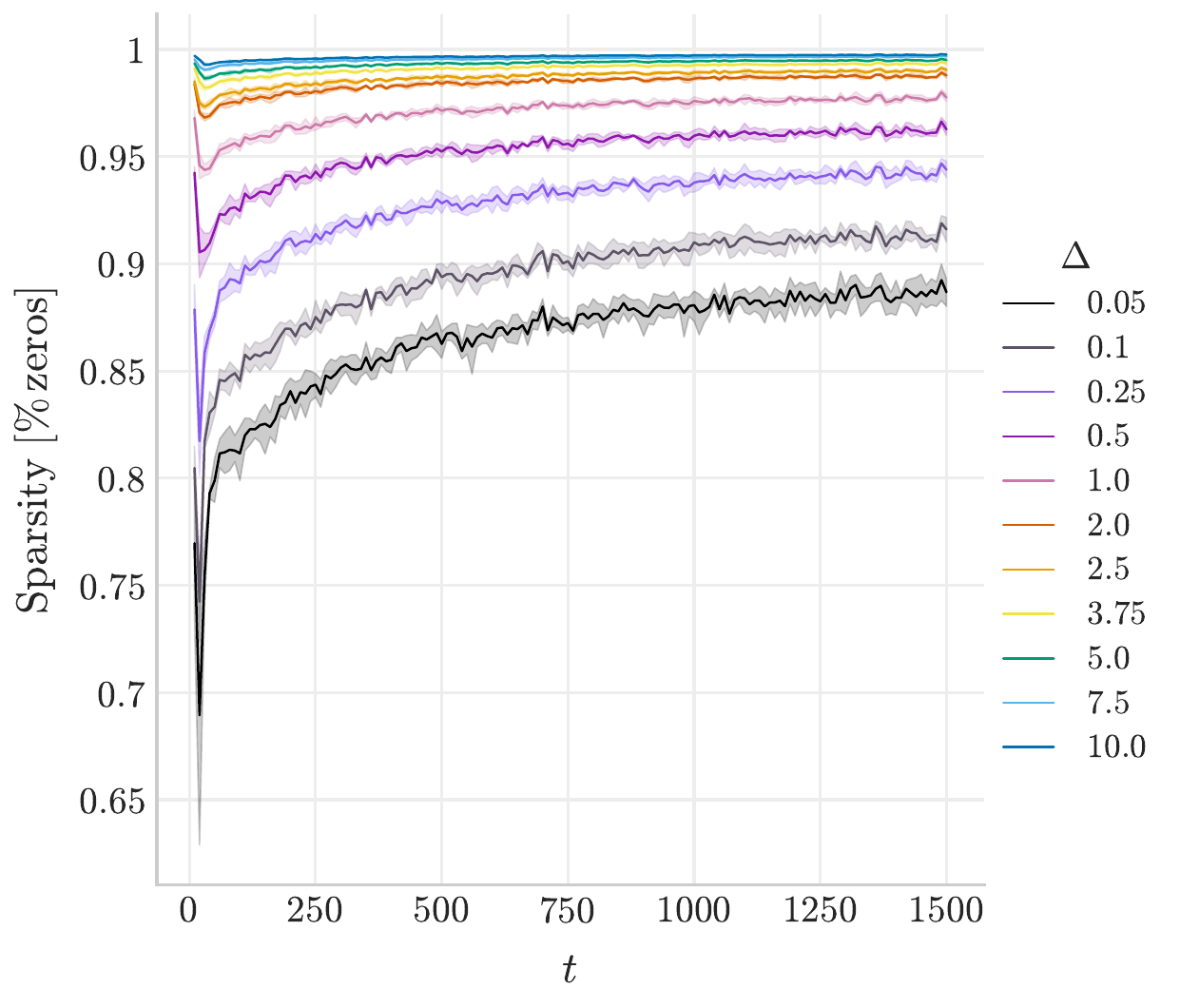}
    \caption{EMNIST, FedAdam}
\end{subfigure}
\begin{subfigure}{0.48\linewidth}
    \centering
    \includegraphics[height=.75\linewidth]{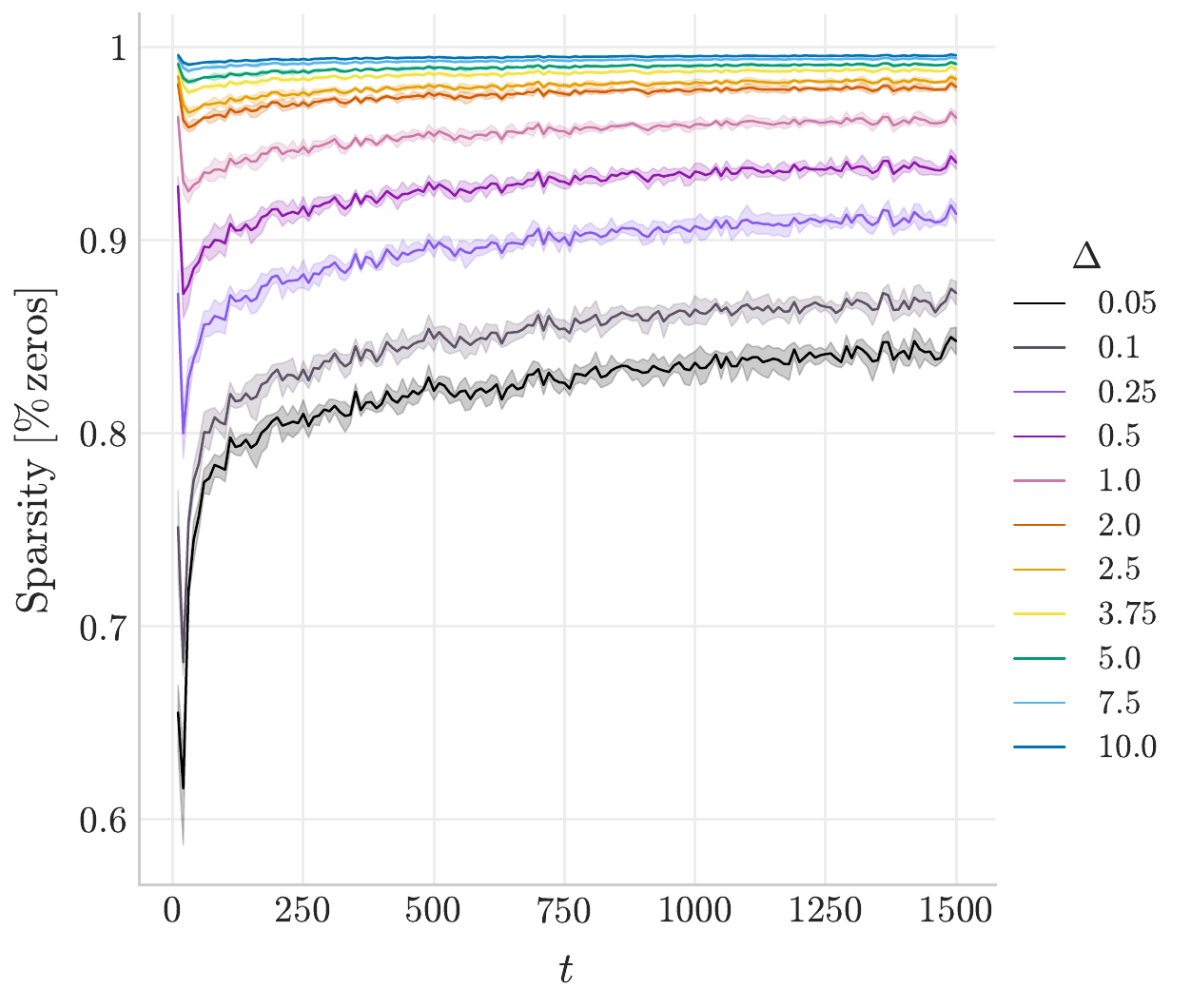}
    \caption{EMNIST, FedAvg}
\end{subfigure}
\begin{subfigure}{0.48\linewidth}
    \centering
    \includegraphics[height=.75\linewidth]{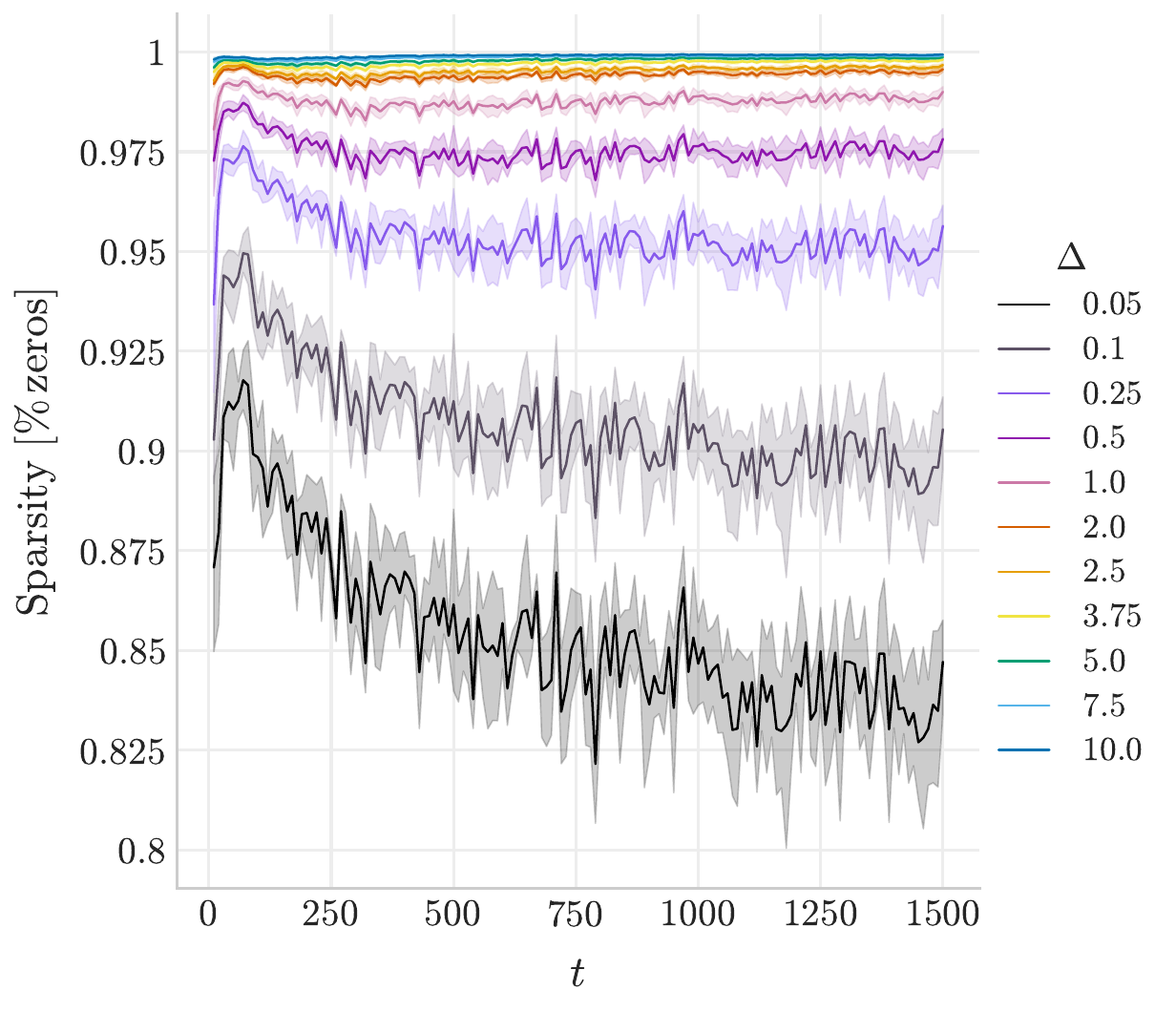}
    \caption{Stack Overflow NWP, FedAdam}
\end{subfigure}
\centering
\begin{subfigure}{0.48\linewidth}
    \centering
    \includegraphics[height=.75\linewidth]{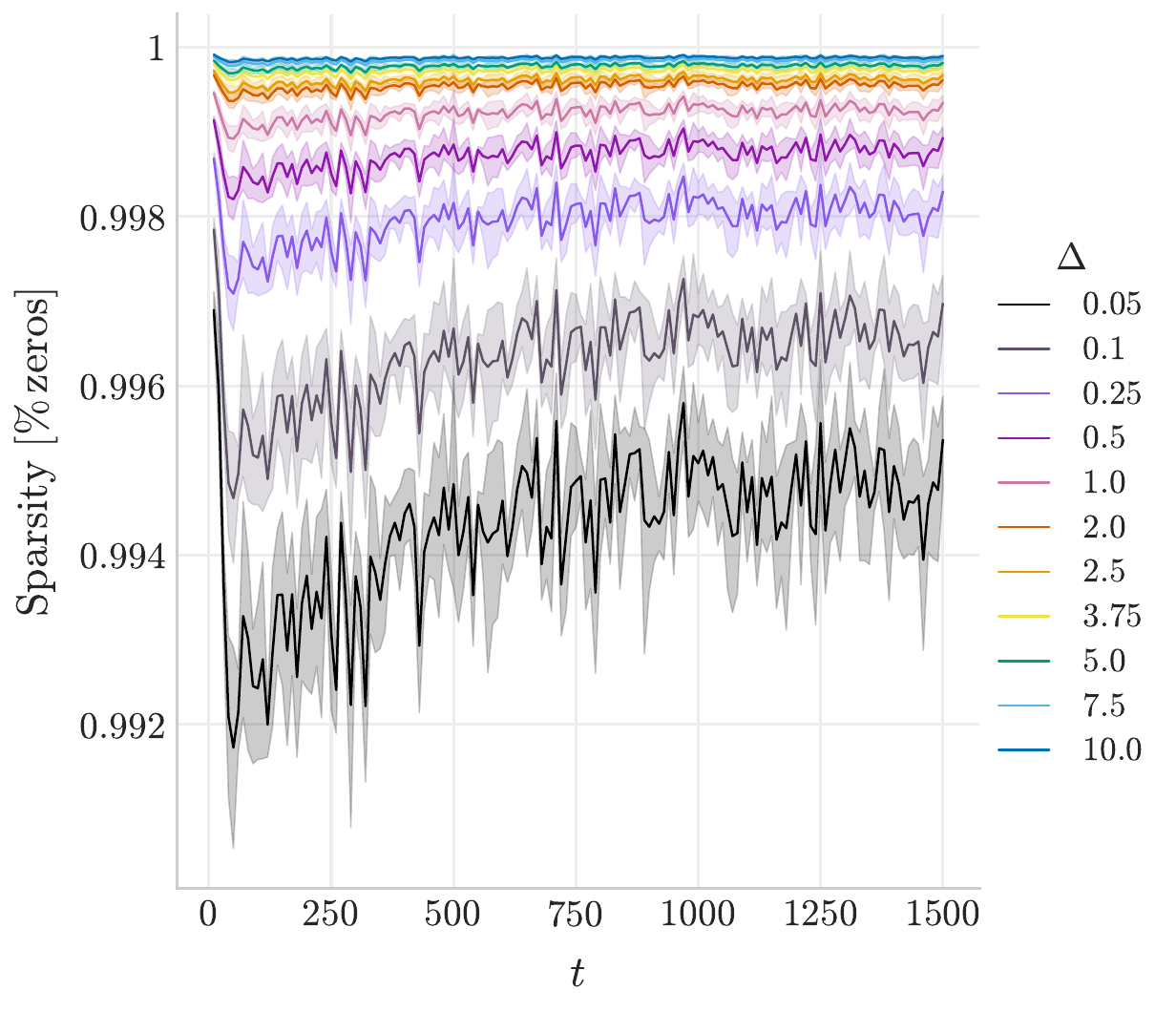}
    \caption{Stack Overflow TP, FedAdam}
\end{subfigure}
\caption{Average sparsity of quantized client updates across step sizes.}
\end{figure}

\newpage
\subsection{Universal Code Overhead}\label{appendix:entropy-cross-entropy}

The same experiment as in Figure~\ref{entropy-cross-entropy}, with additional tasks and values of $\Delta$. We find that the \emph{Elias gamma} code consistently outperforms the \emph{Elias delta} code with an acceptable overhead over the entropy of the quantized non-zero magnitudes across values of $\Delta$ for both the EMNIST character recognition and Stack Overflow next-word prediction tasks.

\begin{figure}[H]
\centering
\begin{subfigure}{\linewidth}
    \centering
    \includegraphics[width=0.75\linewidth]{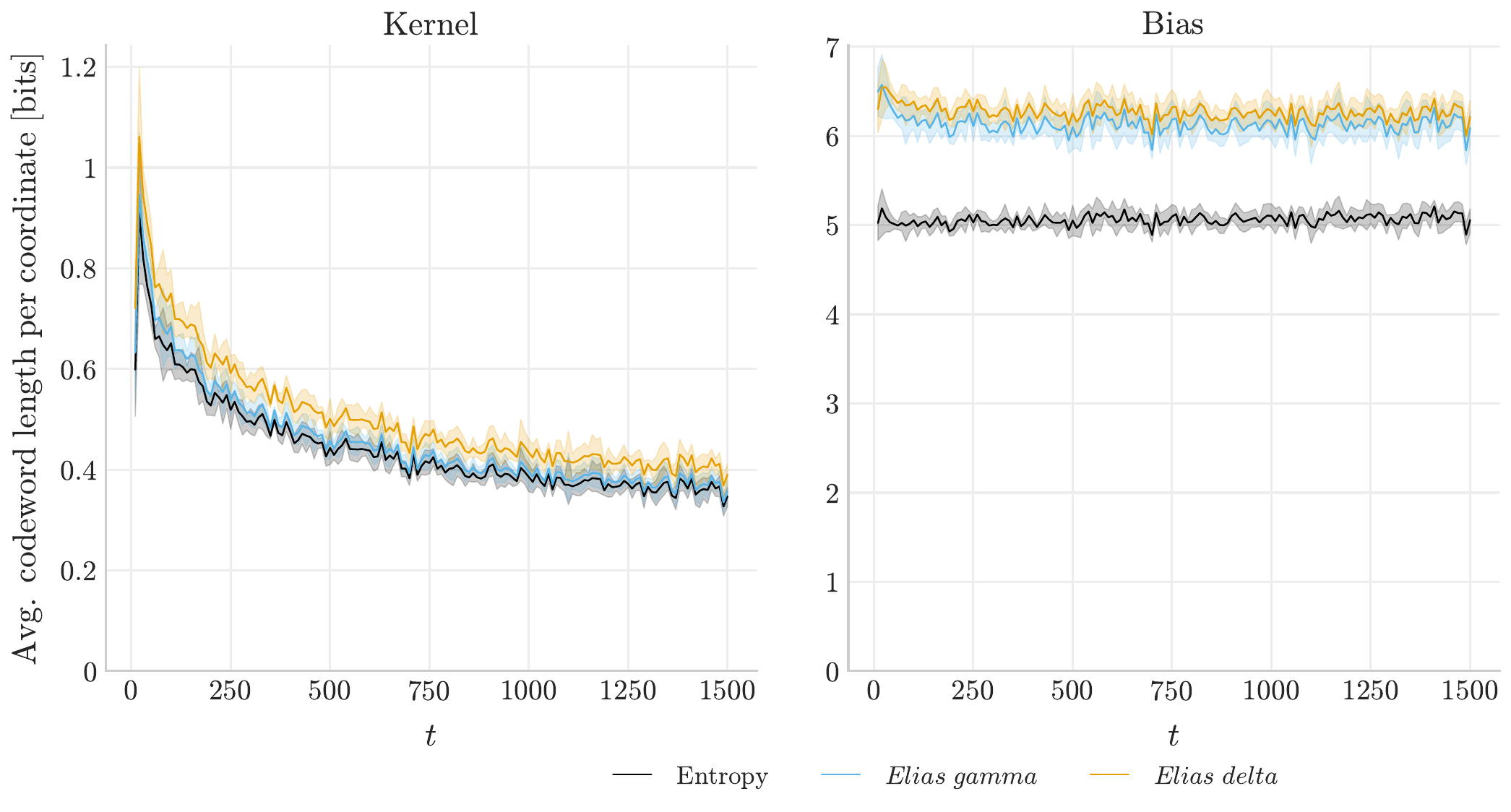}
    \caption{EMNIST, FedAdam, $\Delta = 0.05$}
\end{subfigure}
\centering
\begin{subfigure}{\linewidth}
    \centering
    \includegraphics[width=0.75\linewidth]{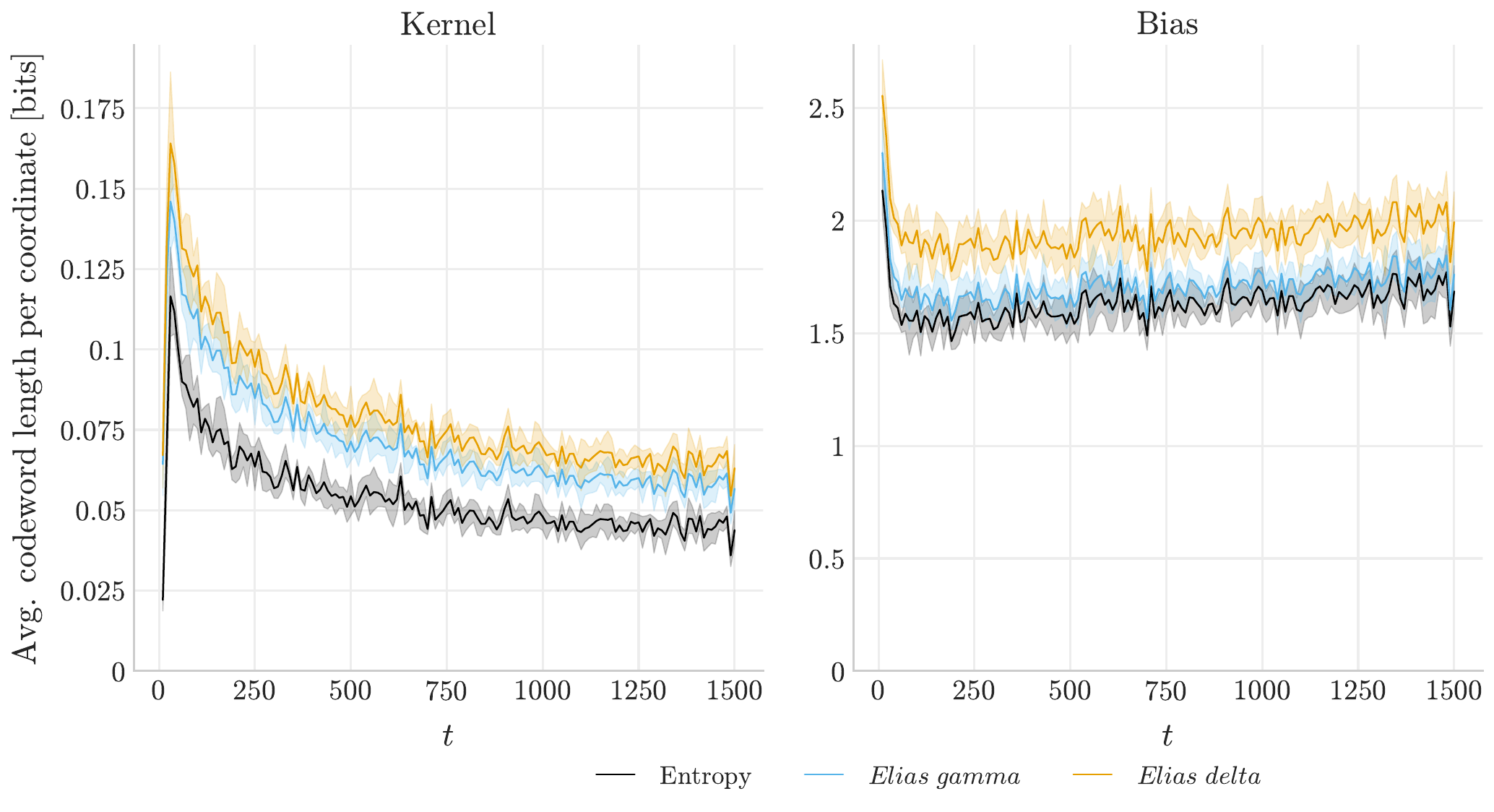}
    \caption{EMNIST, FedAdam, $\Delta = 0.5$}
\end{subfigure}
\centering
\begin{subfigure}{\linewidth}
    \centering
    \includegraphics[width=0.75\linewidth]{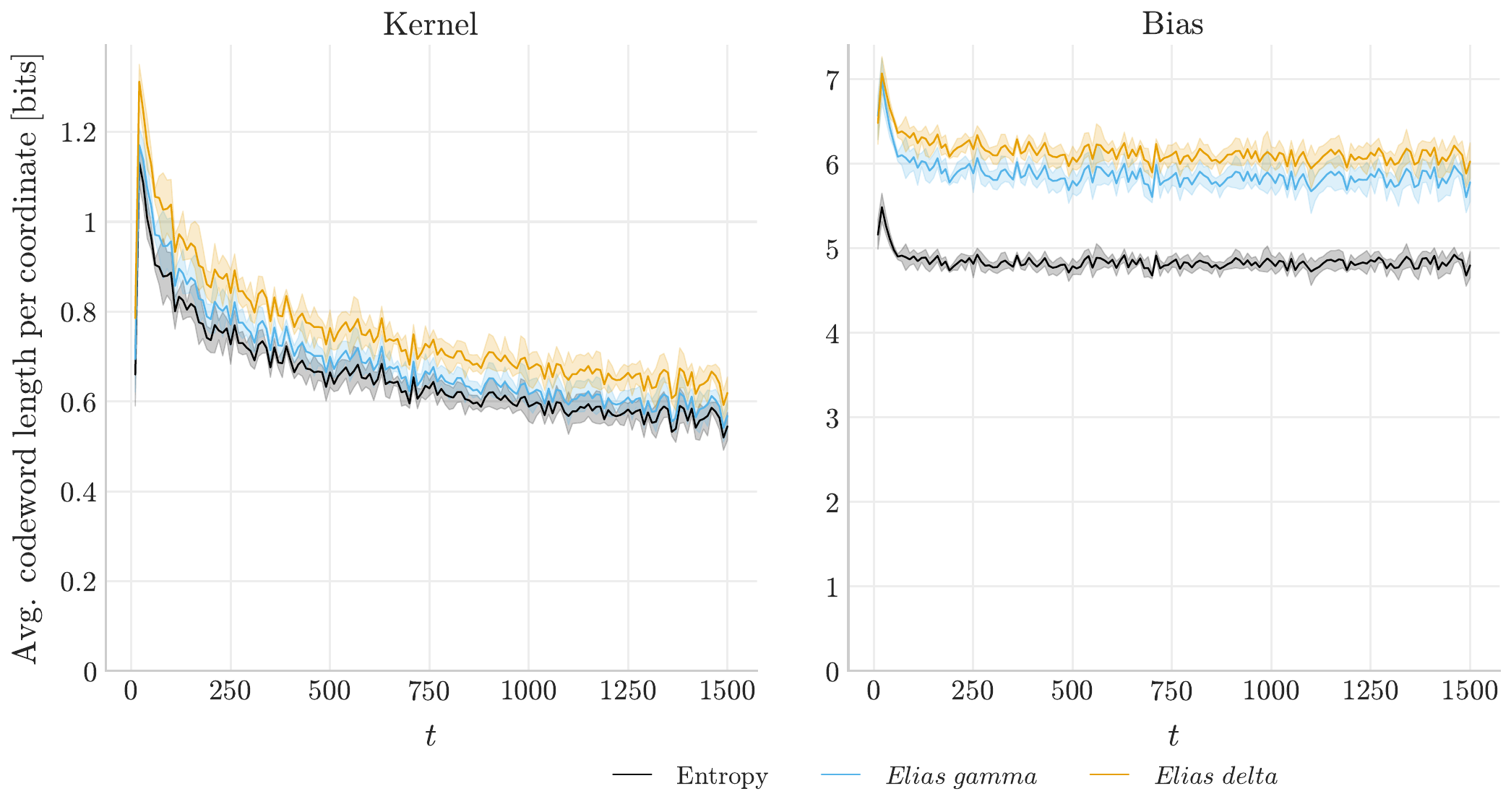}
    \caption{EMNIST, FedAvg, $\Delta = 0.05$}
\end{subfigure}
\caption{Encoding the magnitude of quantized client updates using the \emph{Elias gamma} and \emph{Elias delta} codes results in an acceptable overhead over the entropy of the data throughout training across different model layer types.}
\end{figure}
\begin{figure}[H]\ContinuedFloat
\centering
\begin{subfigure}{\linewidth}
    \centering
    \includegraphics[width=0.75\linewidth]{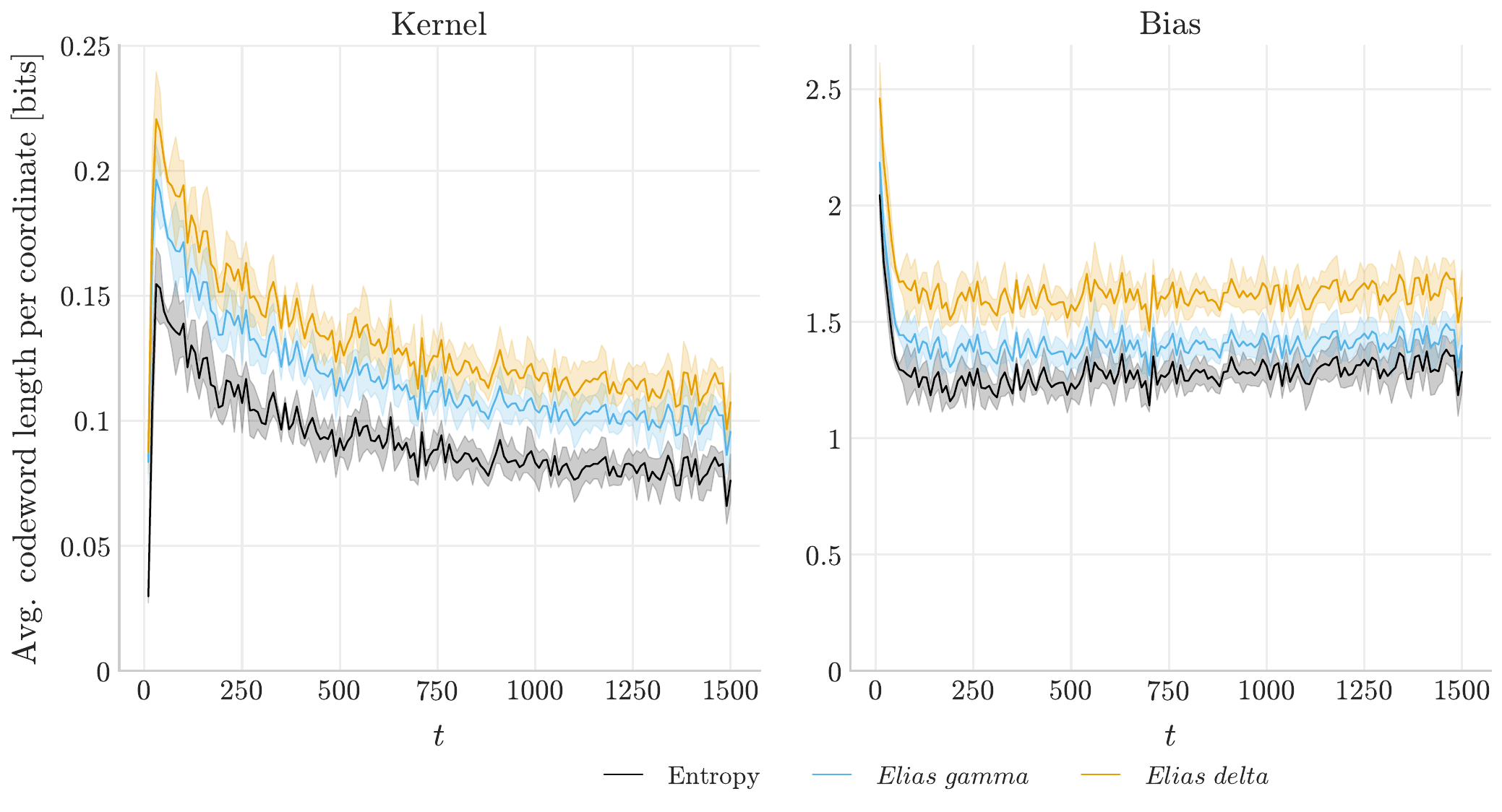}
    \caption{EMNIST, FedAvg, $\Delta = 0.5$}
\end{subfigure}
\centering
\begin{subfigure}{\linewidth}
    \centering
    \hspace*{-.25\linewidth}
    \includegraphics[width=1.5\linewidth]{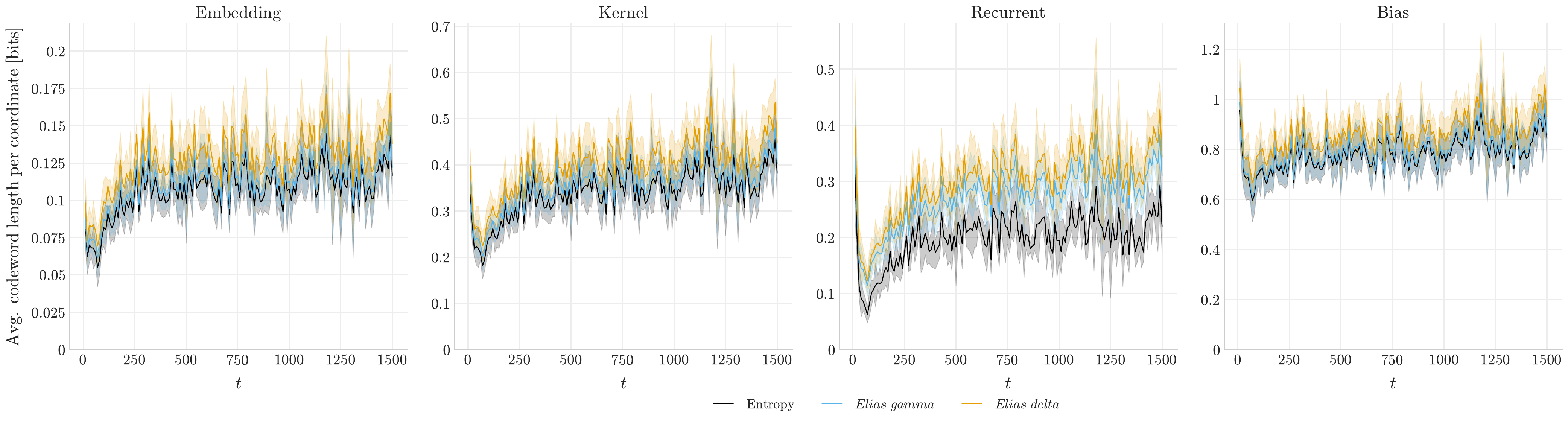}
    \caption{Stack Overflow NWP, FedAdam, $\Delta = 0.05$}
\end{subfigure}
\centering
\begin{subfigure}{\linewidth}
    \centering
    \hspace*{-.25\linewidth}
    \includegraphics[width=1.5\linewidth]{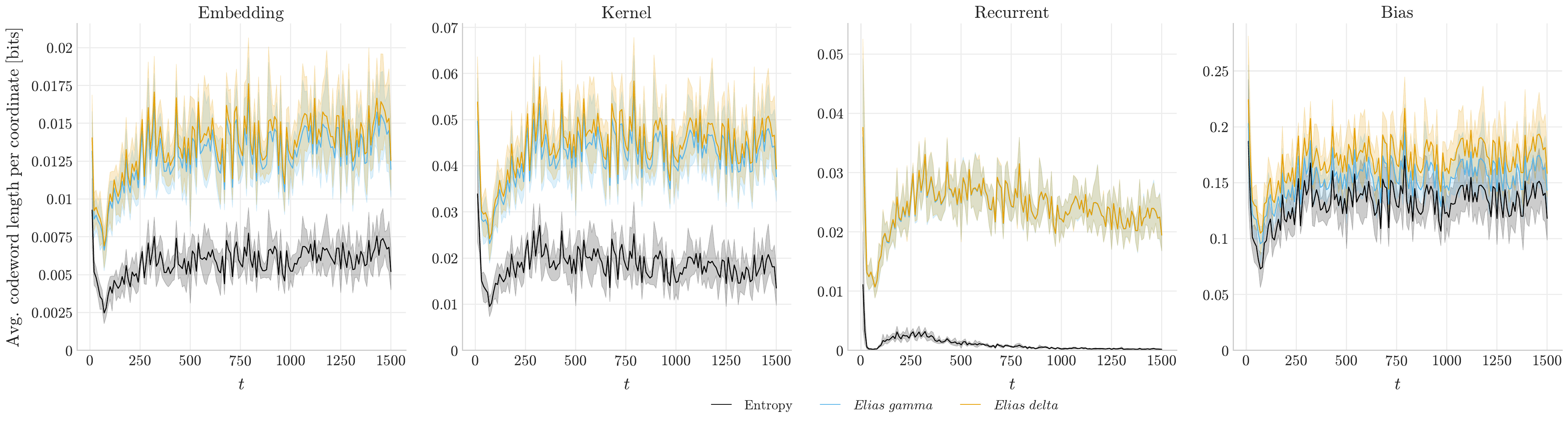}
    \caption{Stack Overflow NWP, FedAdam, $\Delta = 0.5$}
\end{subfigure}
\caption{Encoding the magnitude of quantized client updates using the \emph{Elias gamma} and \emph{Elias delta} codes results in an acceptable overhead over the entropy of the data throughout training across different model layer types. (cont.)}
\end{figure}

\newpage
\subsection{Distortion--Accuracy}\label{appendix:distortion-acc}

The same experiment as in Figure~\ref{so-acc-distortion}, with other tasks and optimizers, showing how distortion is a good proxy for accuracy across the board.

\begin{figure}[H]
\begin{subfigure}{0.48\linewidth}
    \centering
    \includegraphics[height=.75\linewidth]{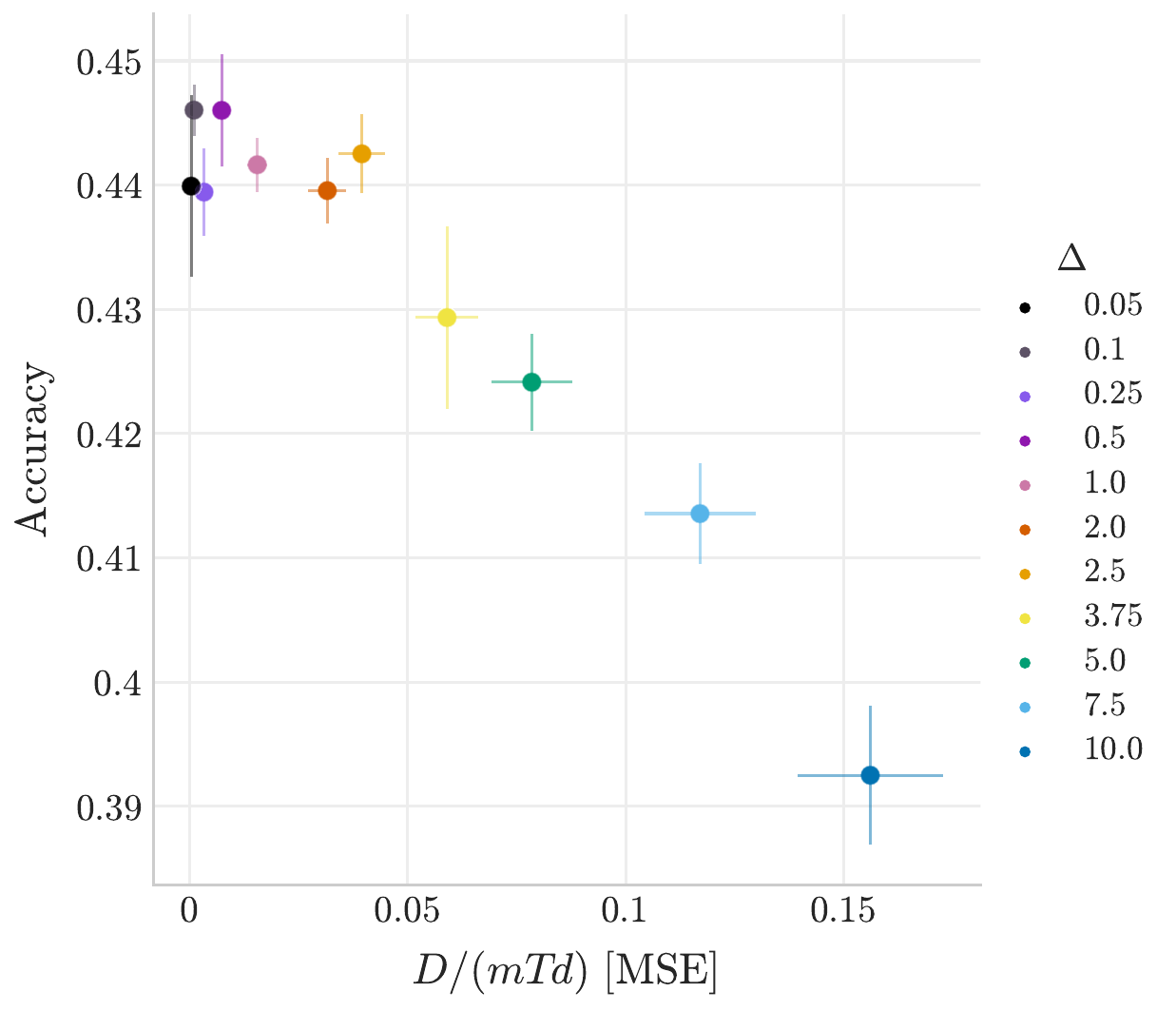}
    \caption{CIFAR-100, FedAdam}
\end{subfigure}
\begin{subfigure}{0.48\linewidth}
    \centering
    \includegraphics[height=.75\linewidth]{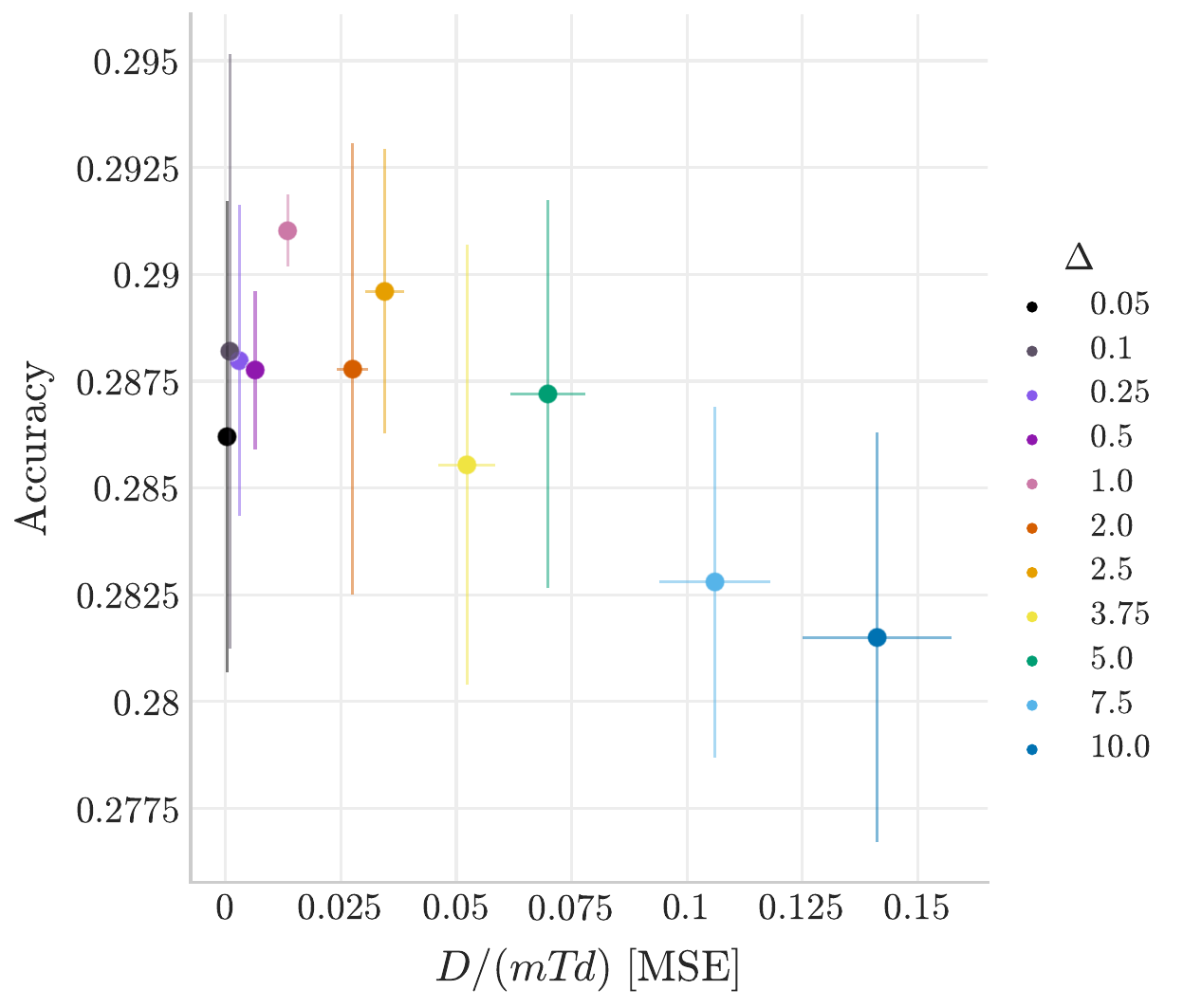}
    \caption{CIFAR-100, FedAvg}
\end{subfigure}
\begin{subfigure}{0.48\linewidth}
    \centering
    \includegraphics[height=.75\linewidth]{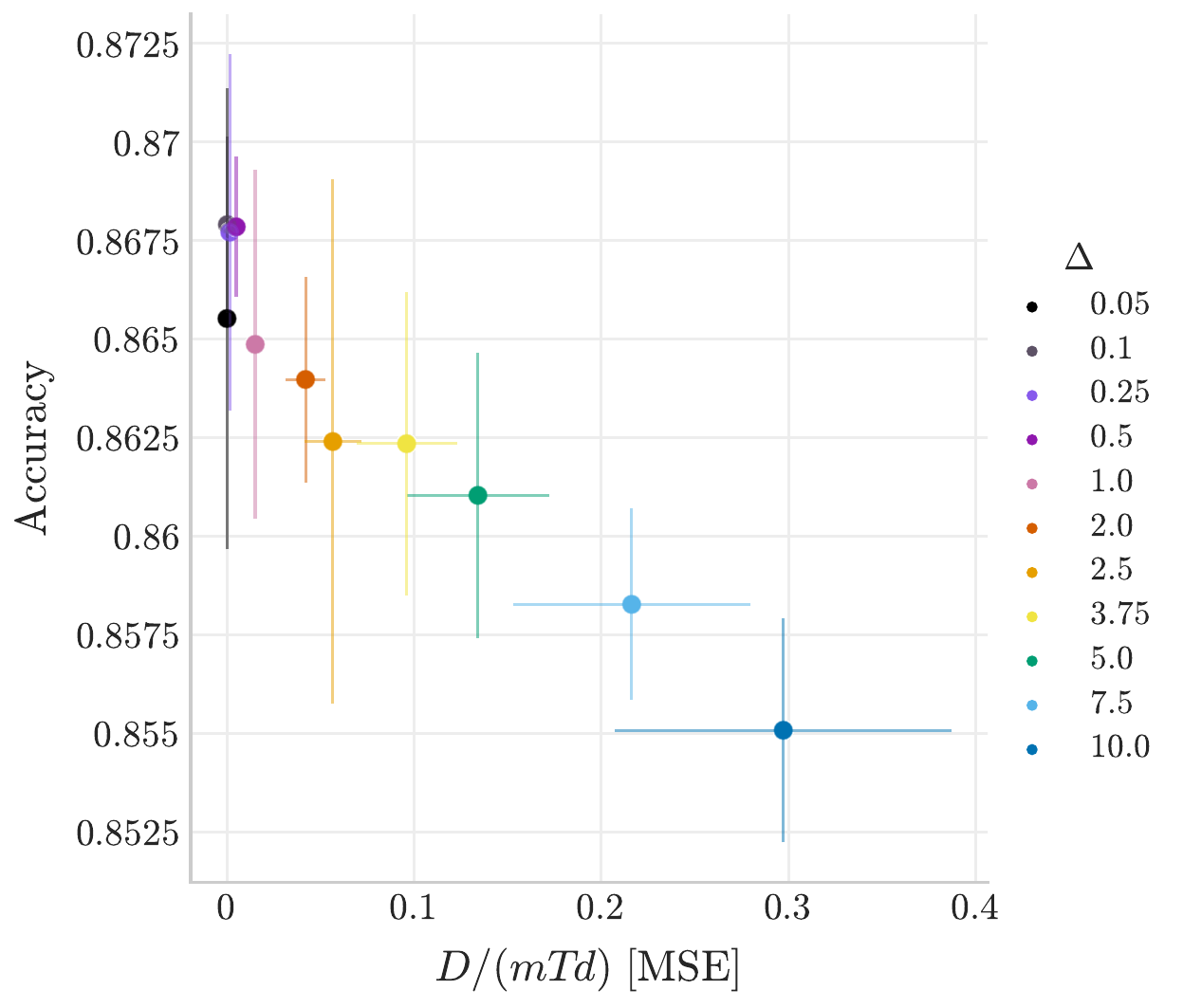}
    \caption{EMNIST, FedAdam}
\end{subfigure}
\begin{subfigure}{0.48\linewidth}
    \centering
    \includegraphics[height=.75\linewidth]{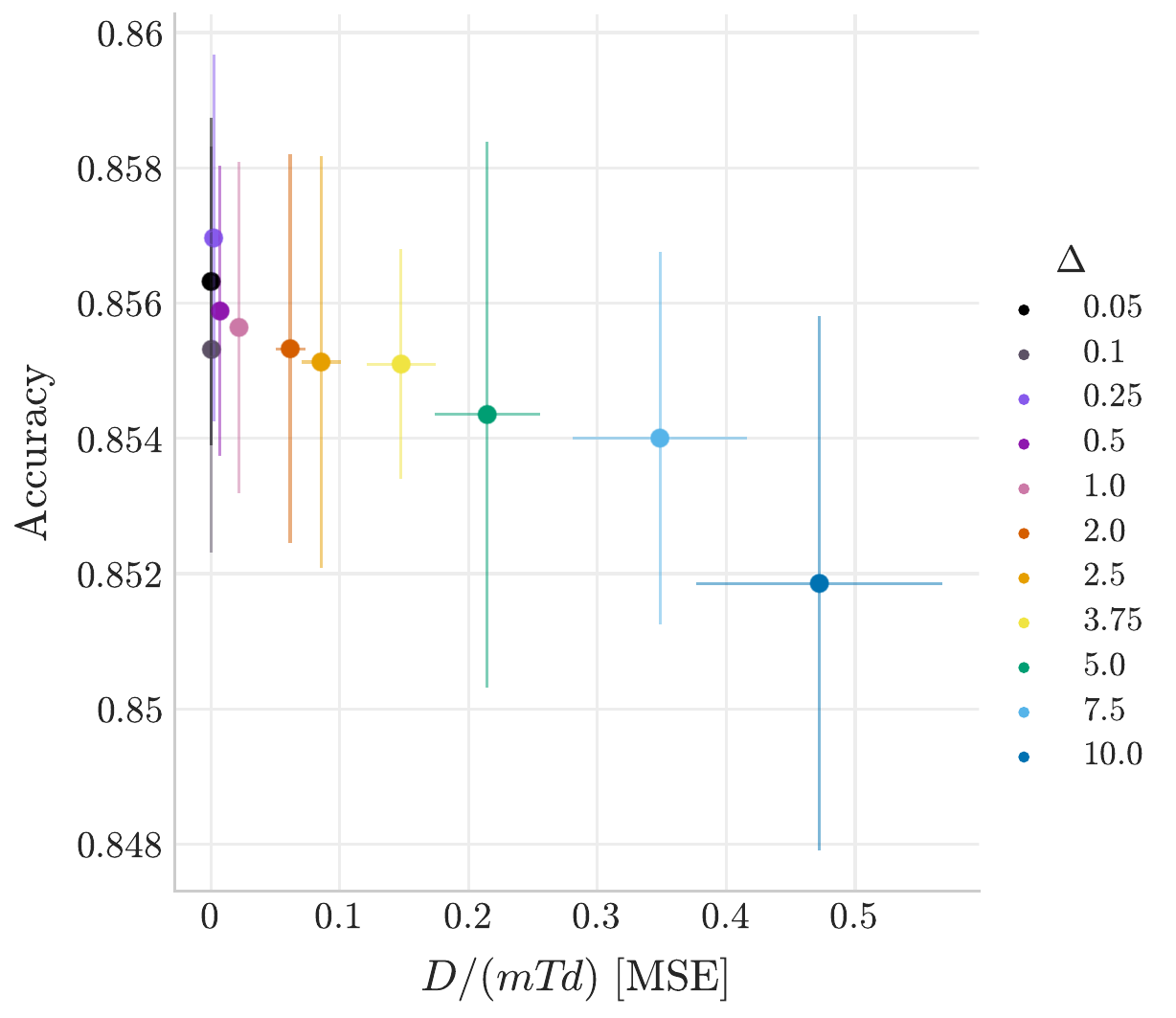}
    \caption{EMNIST, FedAvg}
\end{subfigure}
\begin{subfigure}{0.48\linewidth}
    \centering
    \includegraphics[height=.75\linewidth]{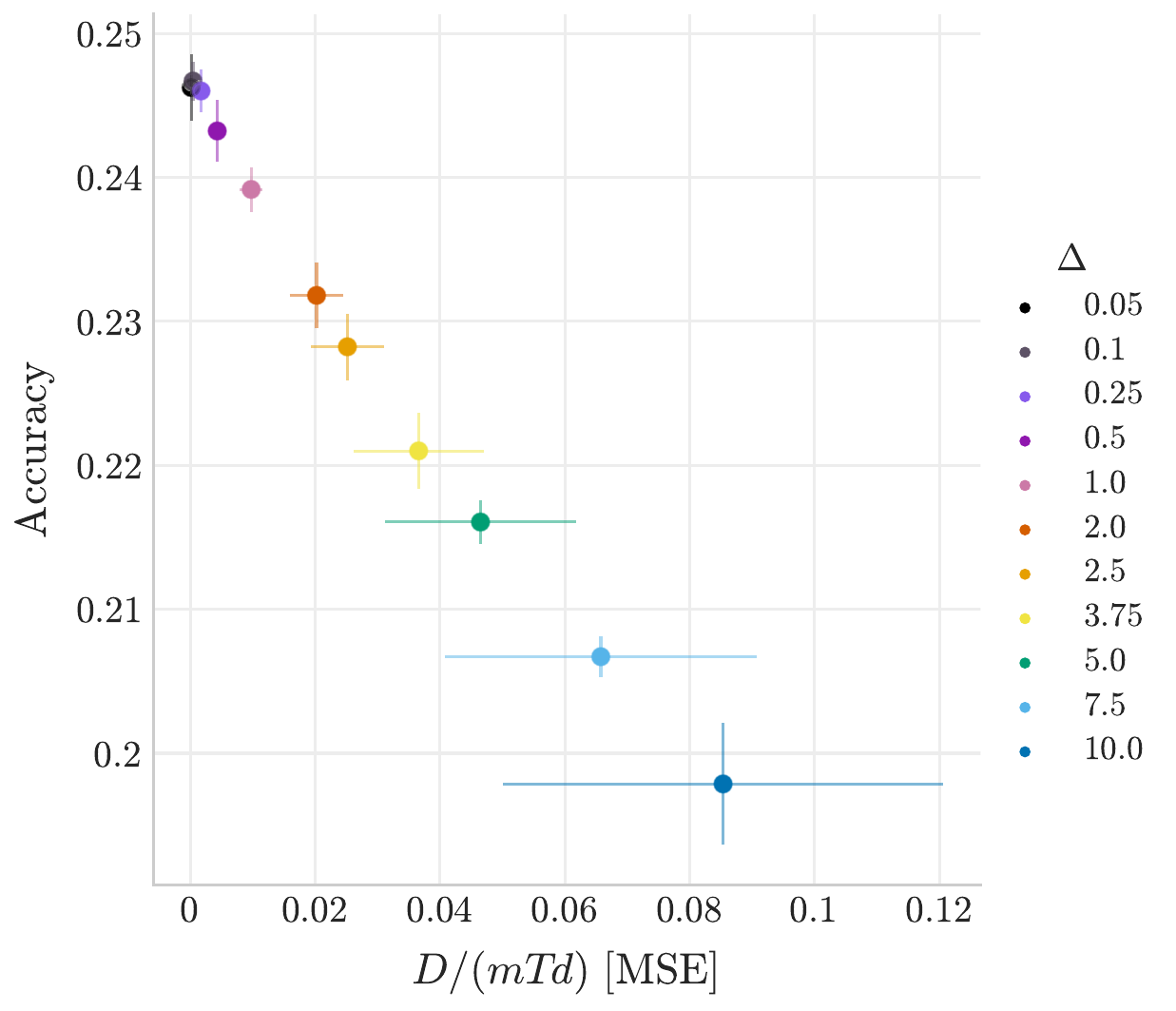}
    \caption{Stack Overflow NWP, FedAdam}
\end{subfigure}
\centering
\begin{subfigure}{0.48\linewidth}
    \centering
    \includegraphics[height=.75\linewidth]{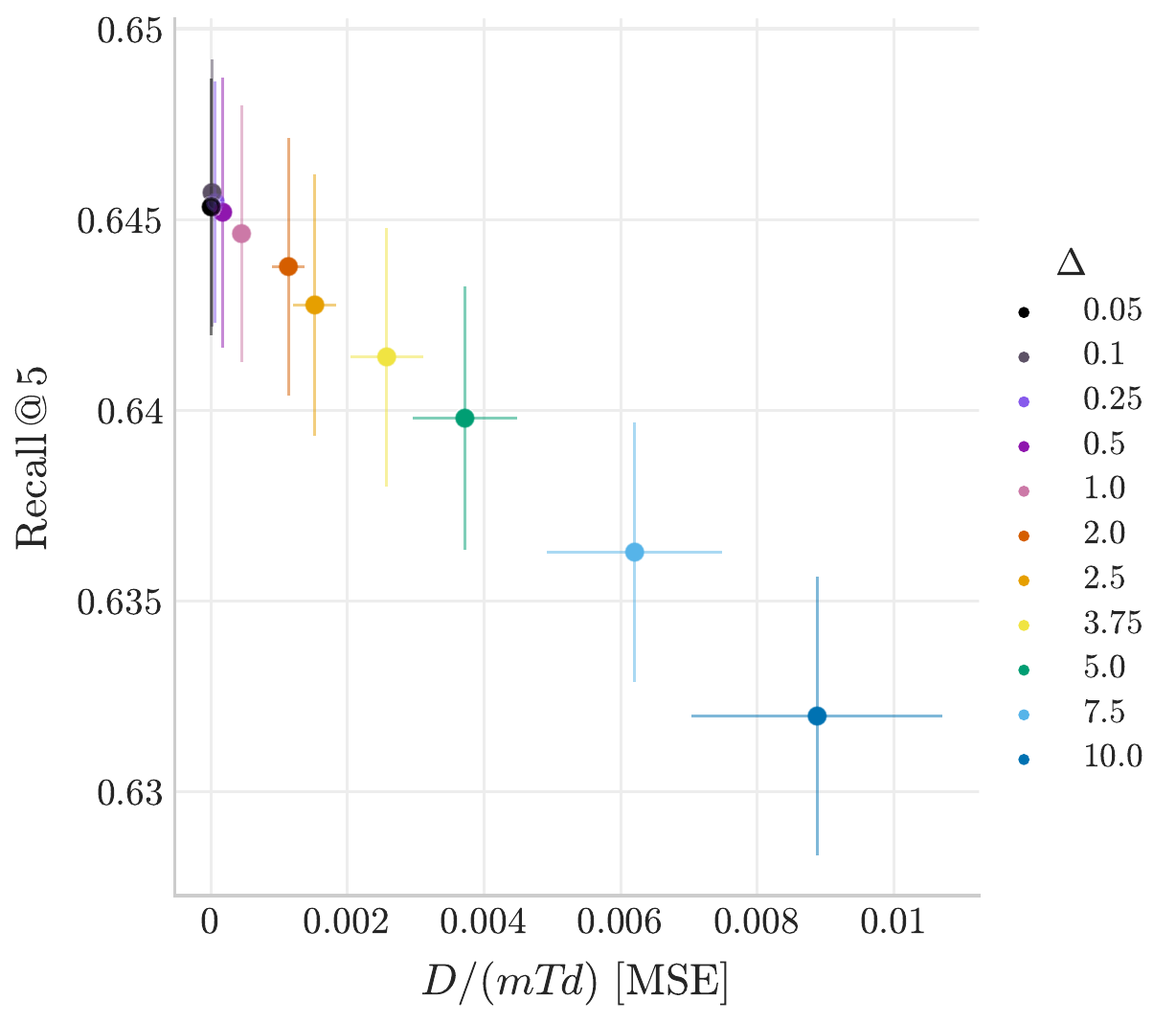}
    \caption{Stack Overflow TP, FedAdam}
\end{subfigure}
\caption{Distortion is a good proxy for the final accuracy of a trained model. Error bars indicate variance in average per-coordinate distortion and final accuracy over five random trials.}
\end{figure}

\newpage
\subsection{Client R-D Optimization Experiments}\label{appendix:votes}
The same experiment as in Figure~\ref{so-fedadam-vote}, with other tasks and optimizers. We find that to minimize their local rate--distortion objective specified by the given $\lambda$, clients select a consistent quantization step size $\Delta$ across all our experiments.

\begin{figure}[H]
\begin{subfigure}{0.48\linewidth}
    \centering
    \includegraphics[height=.75\linewidth]{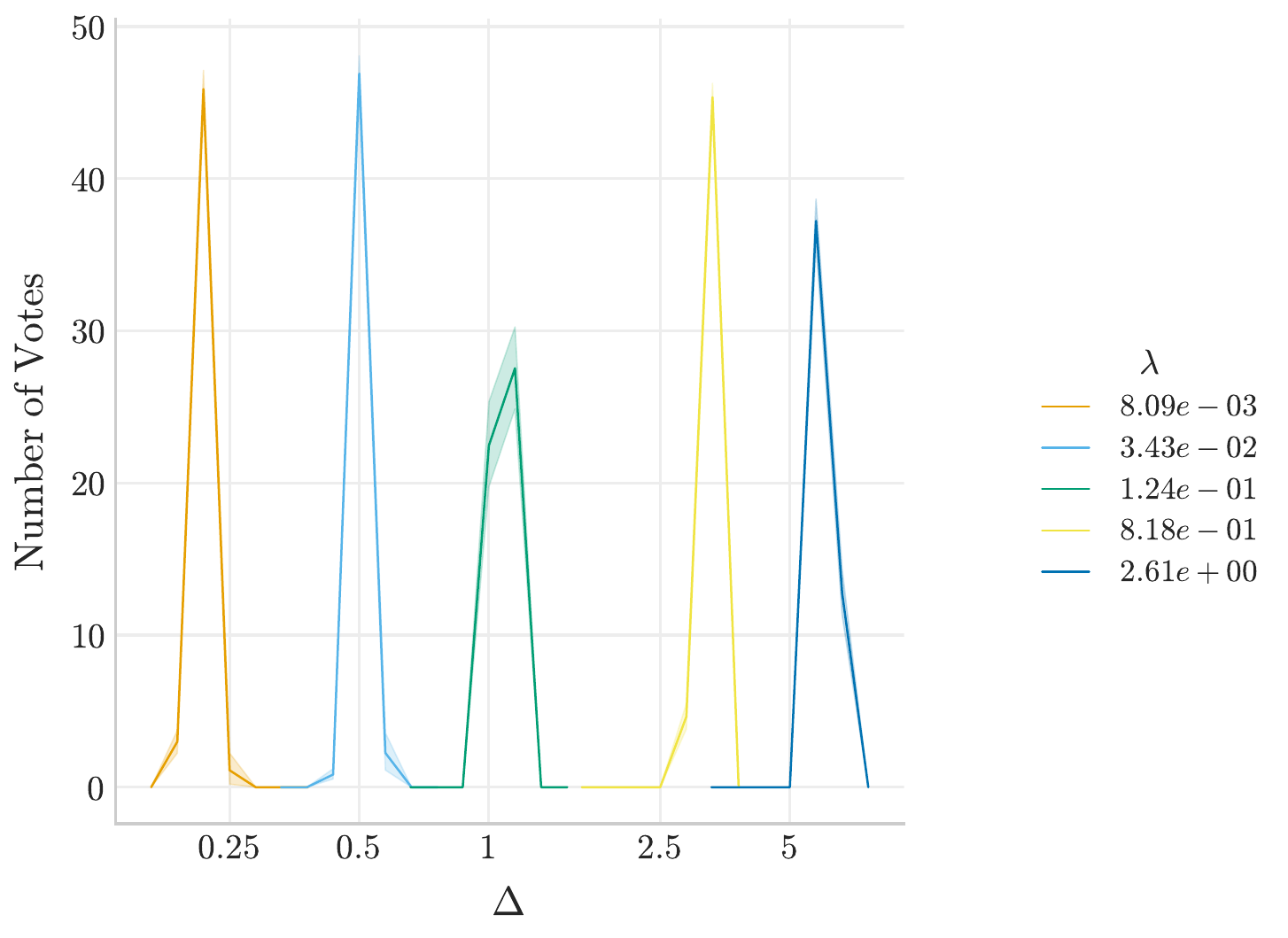}
    \caption{CIFAR-100, FedAdam}
\end{subfigure}
\begin{subfigure}{0.48\linewidth}
    \centering
    \includegraphics[height=.75\linewidth]{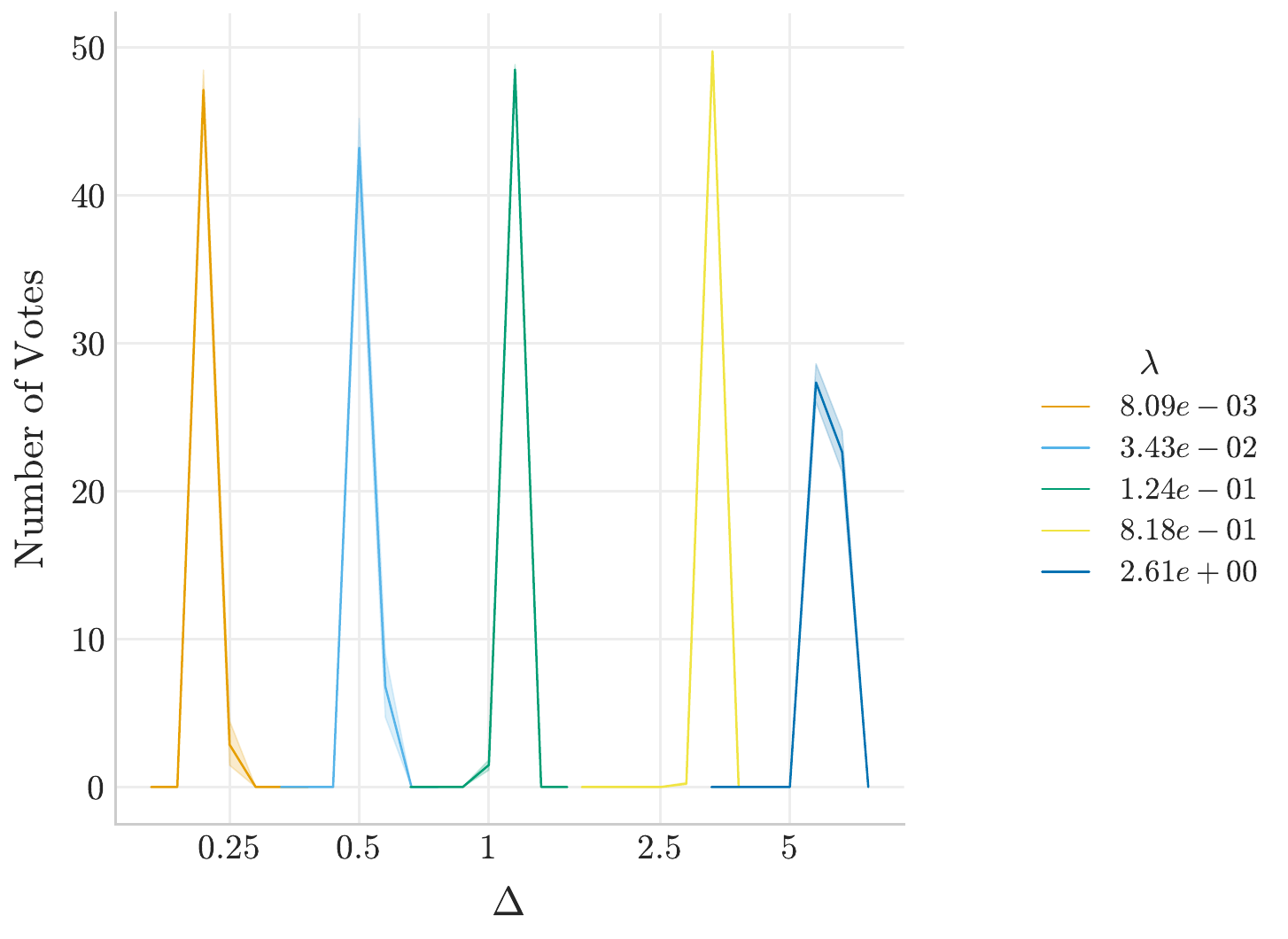}
    \caption{CIFAR-100, FedAvg}
\end{subfigure}
\begin{subfigure}{0.48\linewidth}
    \centering
    \includegraphics[height=.75\linewidth]{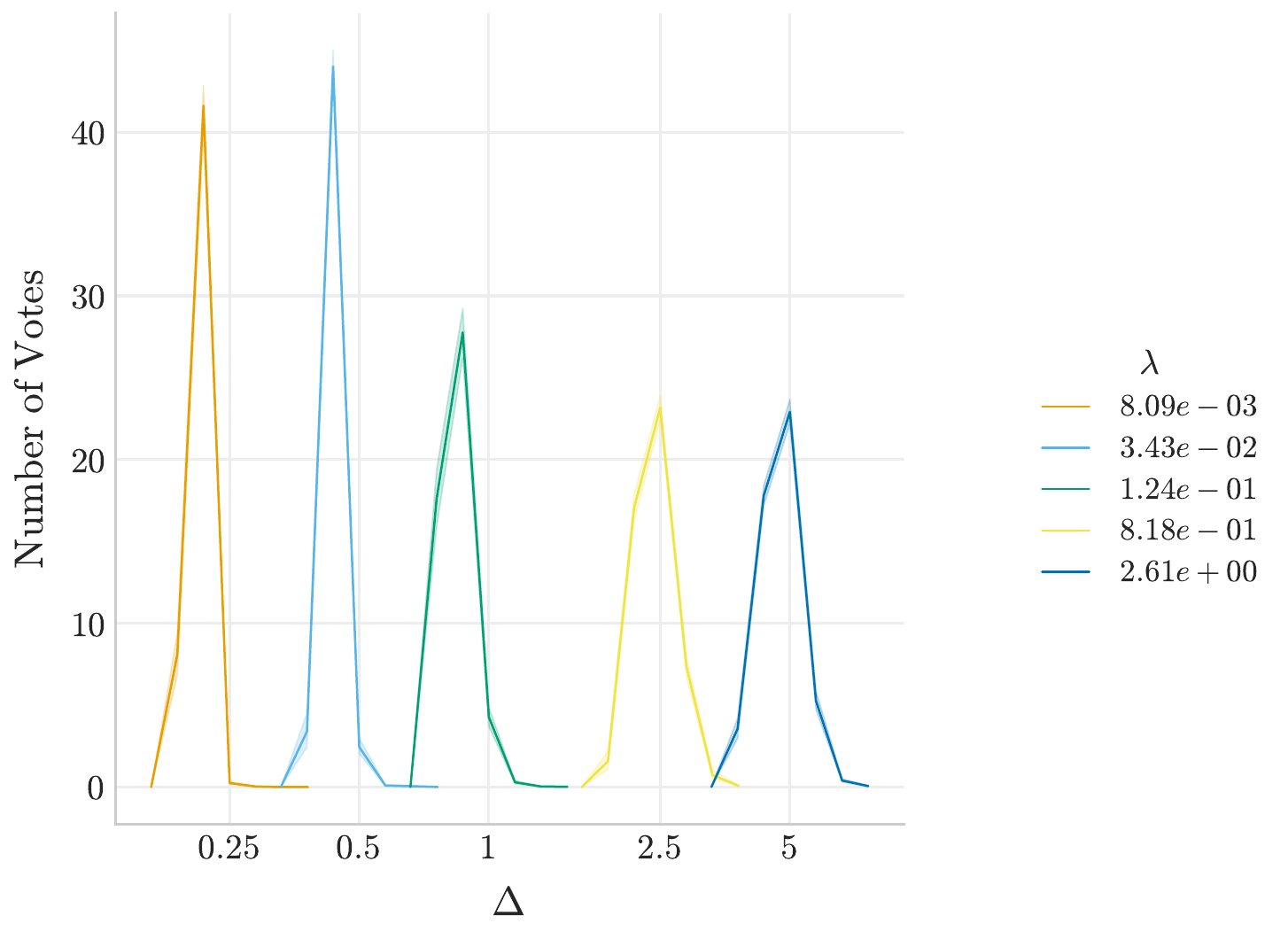}
    \caption{EMNIST, FedAdam}
\end{subfigure}
\begin{subfigure}{0.48\linewidth}
    \centering
    \includegraphics[height=.75\linewidth]{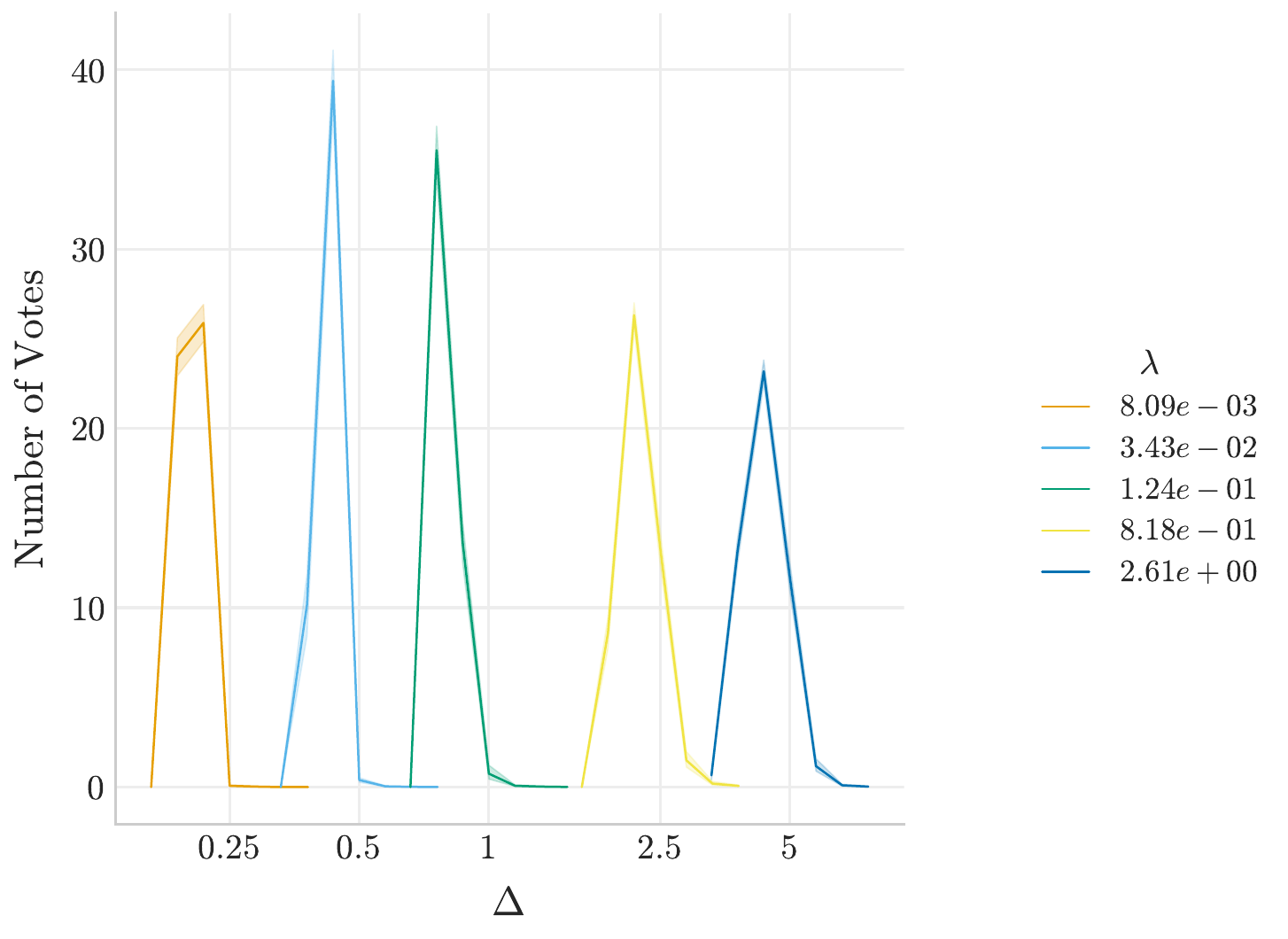}
    \caption{EMNIST, FedAvg}
\end{subfigure}
\begin{subfigure}{0.48\linewidth}
    \centering
    \includegraphics[height=.75\linewidth]{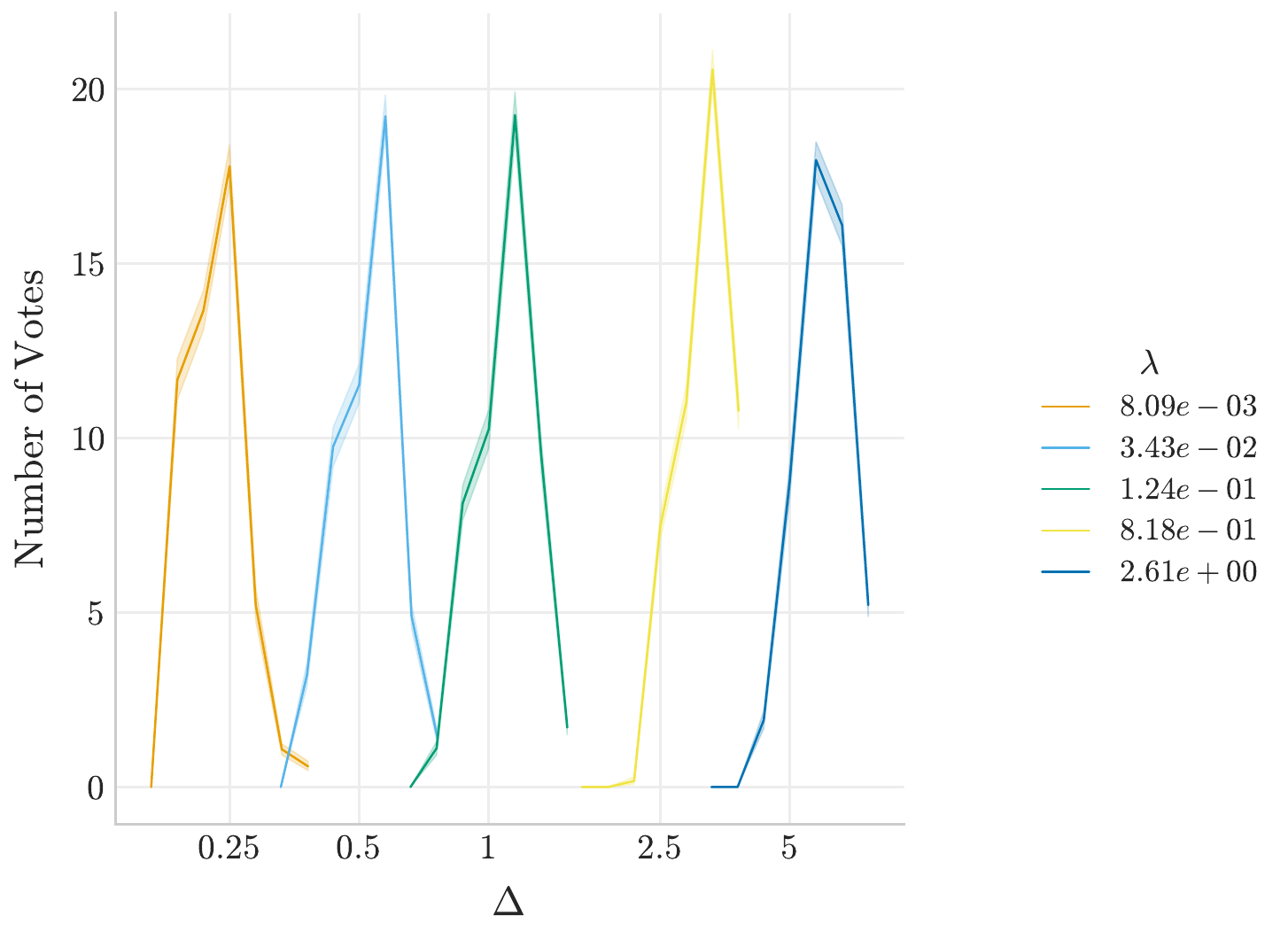}
    \caption{Stack Overflow NWP, FedAdam}
\end{subfigure}
\centering
\begin{subfigure}{0.48\linewidth}
    \centering
    \includegraphics[height=.75\linewidth]{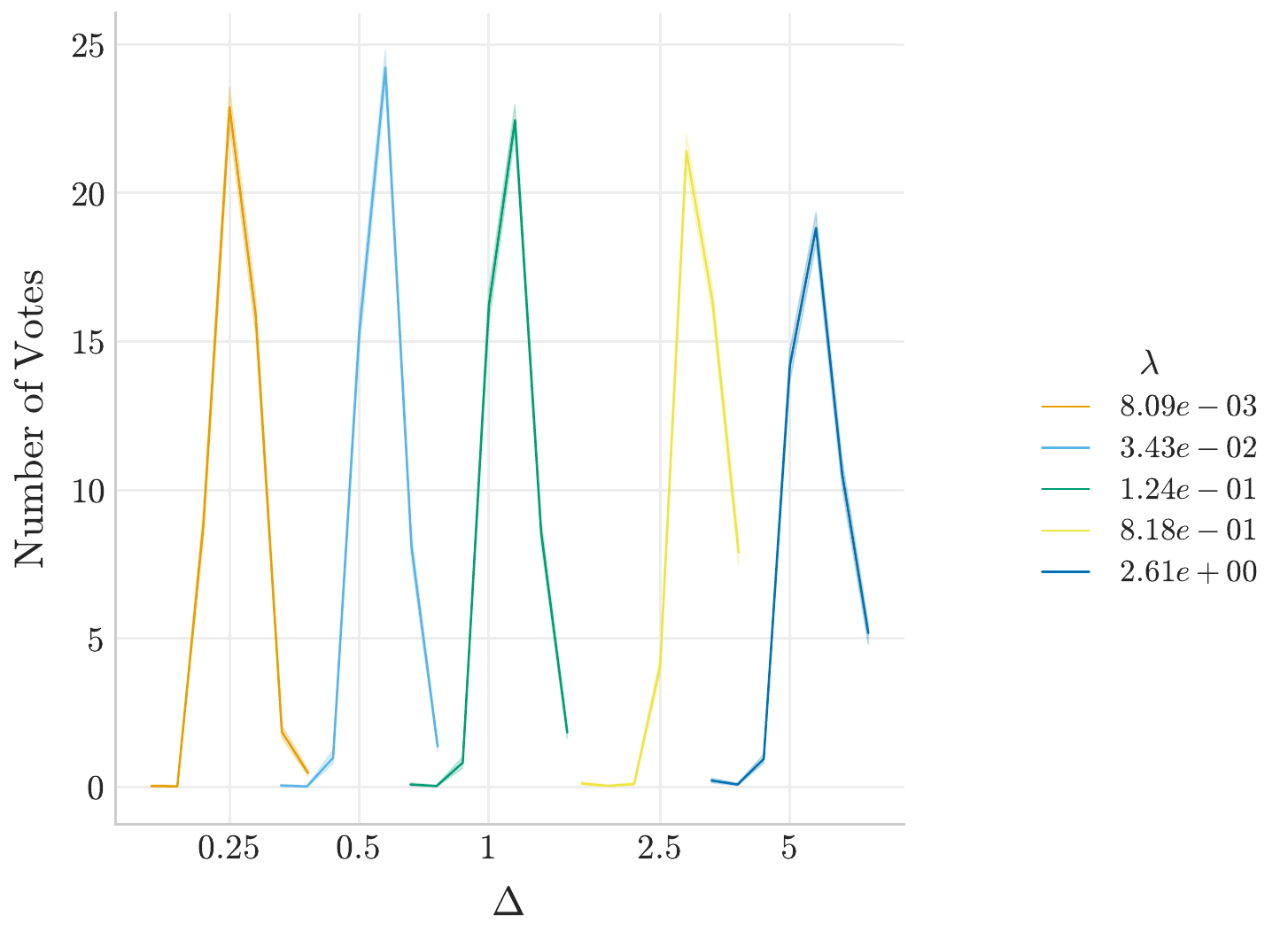}
    \caption{Stack Overflow TP, FedAdam}
\end{subfigure}
\caption{Histograms on selected $\Delta$ for a given $\lambda$, averaged across training rounds.}
\end{figure}

\newpage
\subsection{Rate--Distortion Across Model Architectures}\label{appendix:rd}

Comparing the $R$-$D$ frontiers across tasks and optimizers, we observe that the same quantization step size $\Delta$ tends to yield a similar rate--distortion trade-off. The per-coordinate rate and distortion values shown are averaged across five random trials over the course of training.

\begin{figure}[H]
\centering
\includegraphics[height=.45\linewidth]{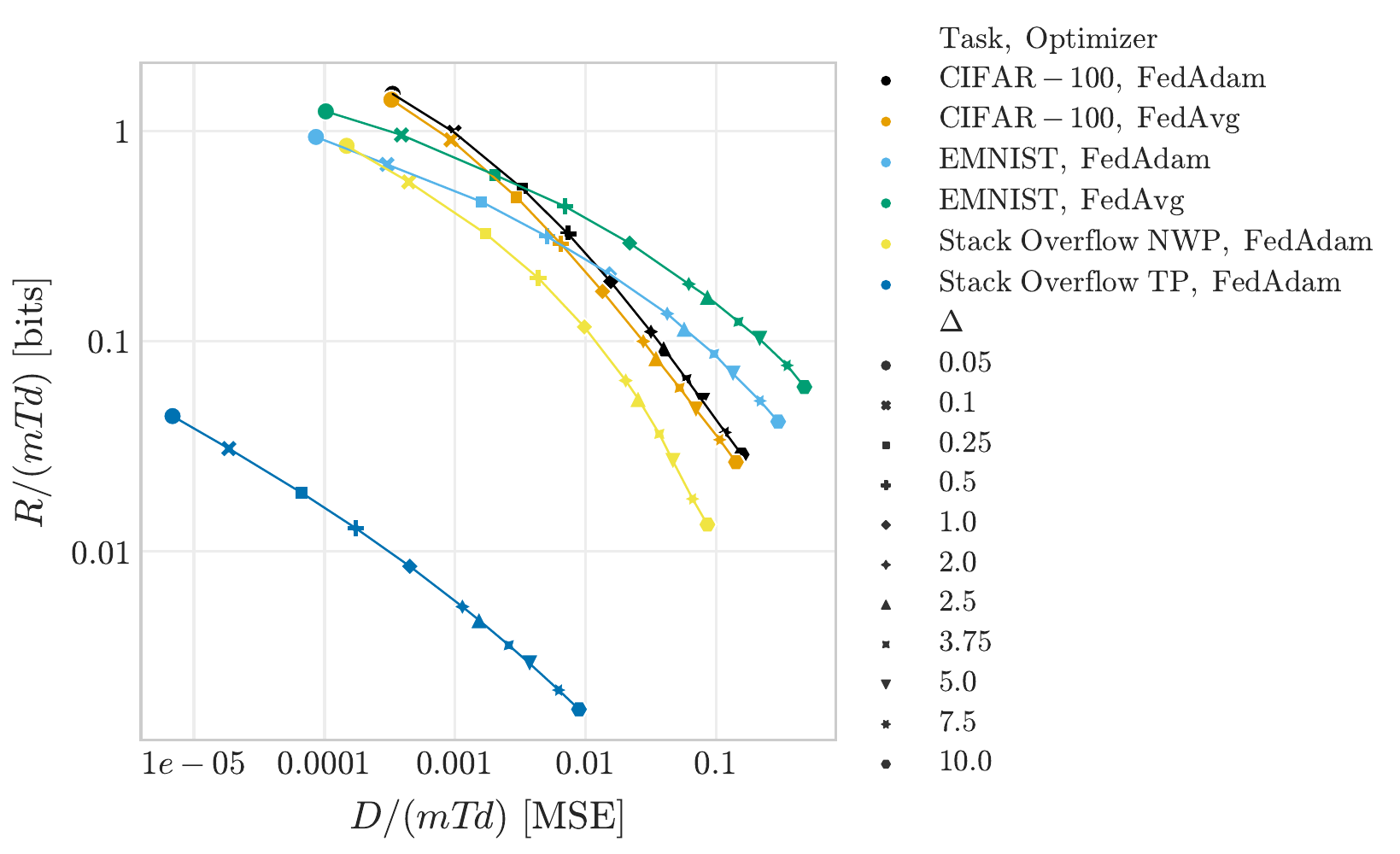}
\vspace{-0.1cm}
\caption{Even though different model architectures tend to have differing $R$-$D$ frontiers, a given choice of $\Delta$ tends to correspond to approximately the same value of $\lambda$ (which is related to the slope of the curve at that point).}
\label{rd}
\end{figure}

\newpage
\subsection{Overall Results}\label{appendix:overall-results}

The same experiment as in Figure~\ref{fig:so-good}, with other tasks and optimizers. Our method significantly outperforms Top-K, DRIVE, and 3LC. Performance is similar to QSGD on tasks with relatively similar data on clients (CIFAR-100 and EMNIST), with a visible gap in Stack Overflow tasks.

\begin{figure}[H]
\begin{subfigure}{0.48\linewidth}
    \centering
    \includegraphics[height=.75\linewidth]{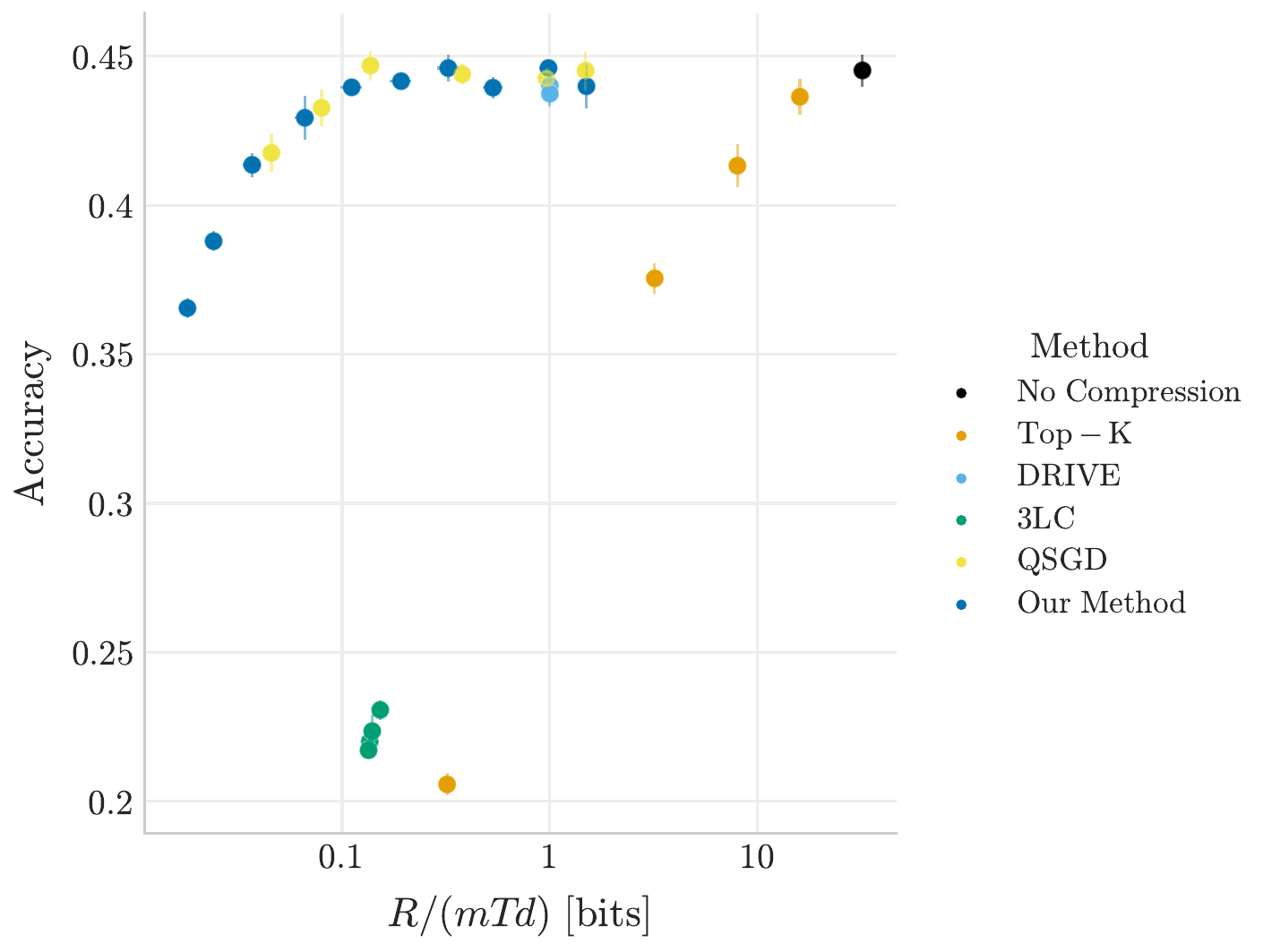}
    \caption{CIFAR-100, FedAdam}
\end{subfigure}
\begin{subfigure}{0.48\linewidth}
    \centering
    \includegraphics[height=.75\linewidth]{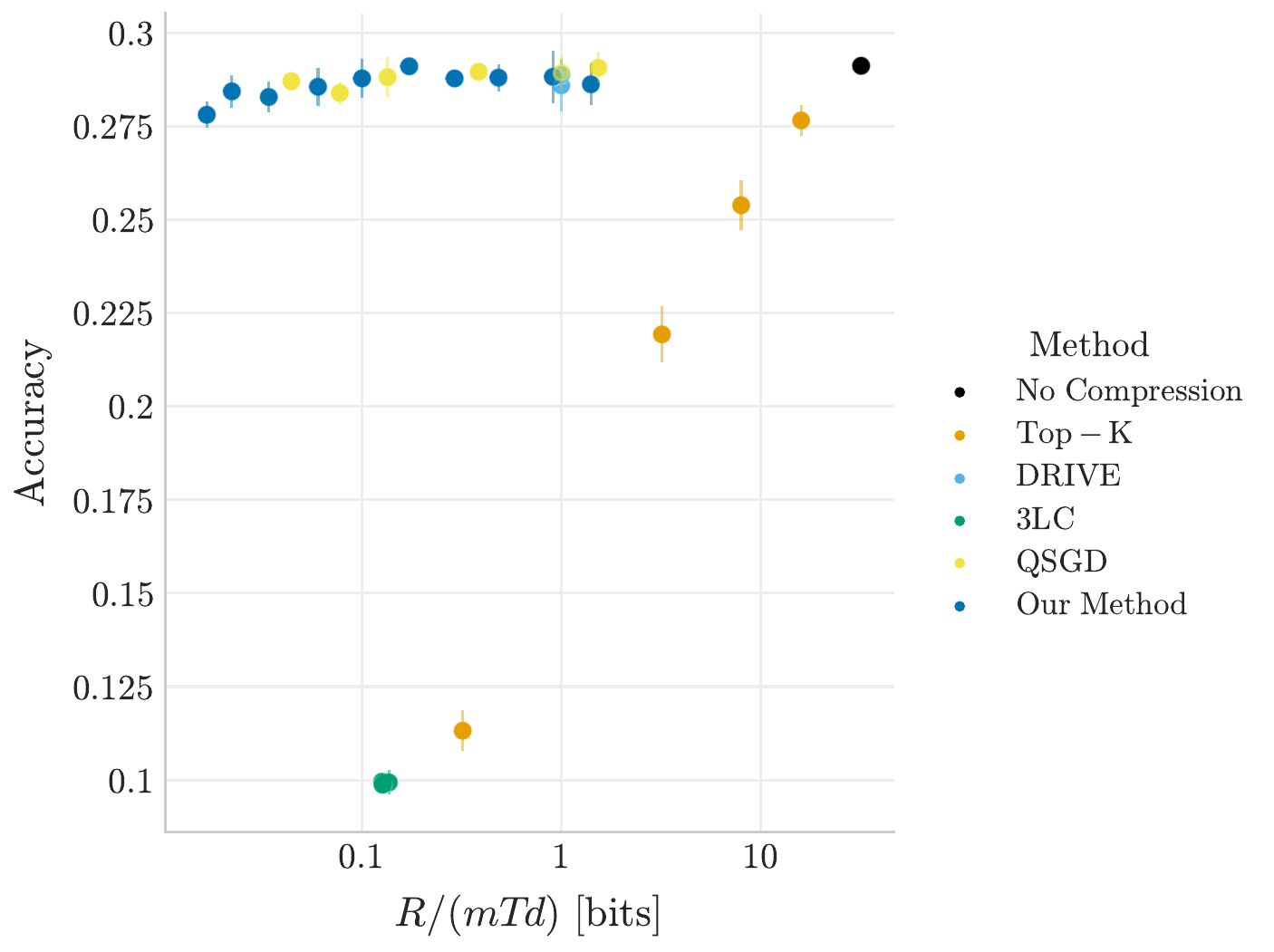}
    \caption{CIFAR-100, FedAvg}
\end{subfigure}
\begin{subfigure}{0.48\linewidth}
    \centering
    \includegraphics[height=.75\linewidth]{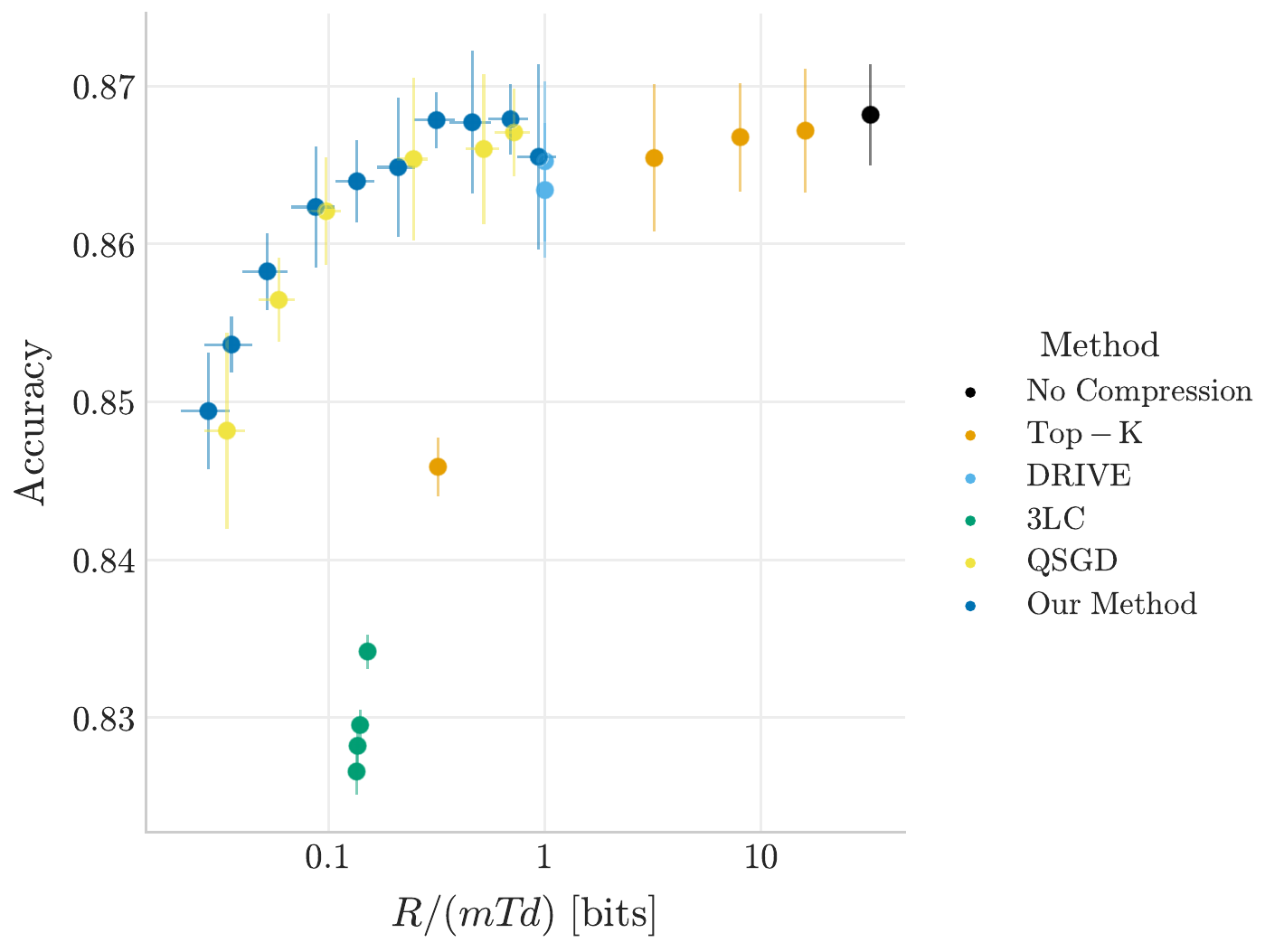}
    \caption{EMNIST, FedAdam}
\end{subfigure}
\begin{subfigure}{0.48\linewidth}
    \centering
    \includegraphics[height=.75\linewidth]{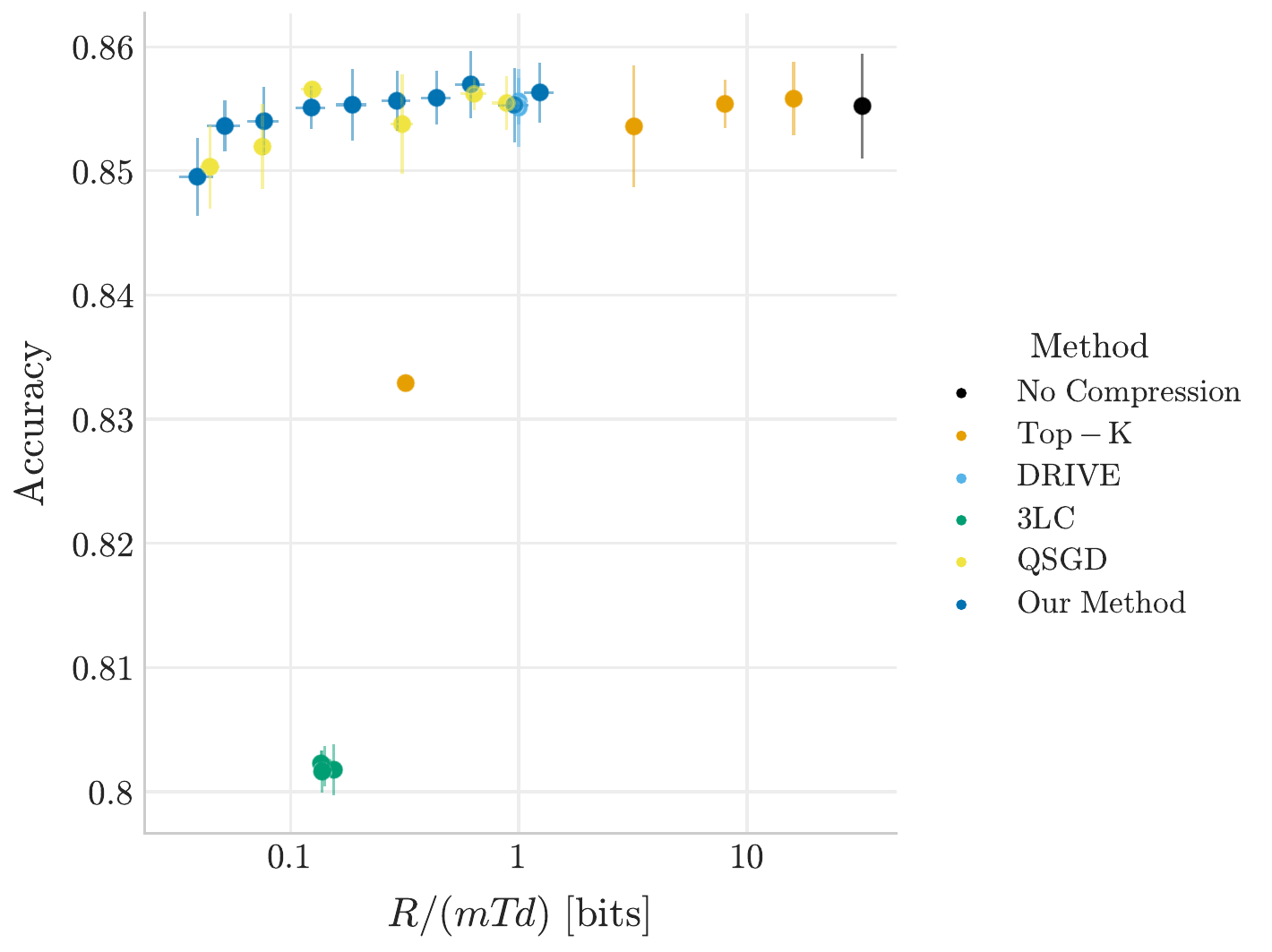}
    \caption{EMNIST, FedAvg}
\end{subfigure}
\begin{subfigure}{0.48\linewidth}
    \centering
    \includegraphics[height=.75\linewidth]{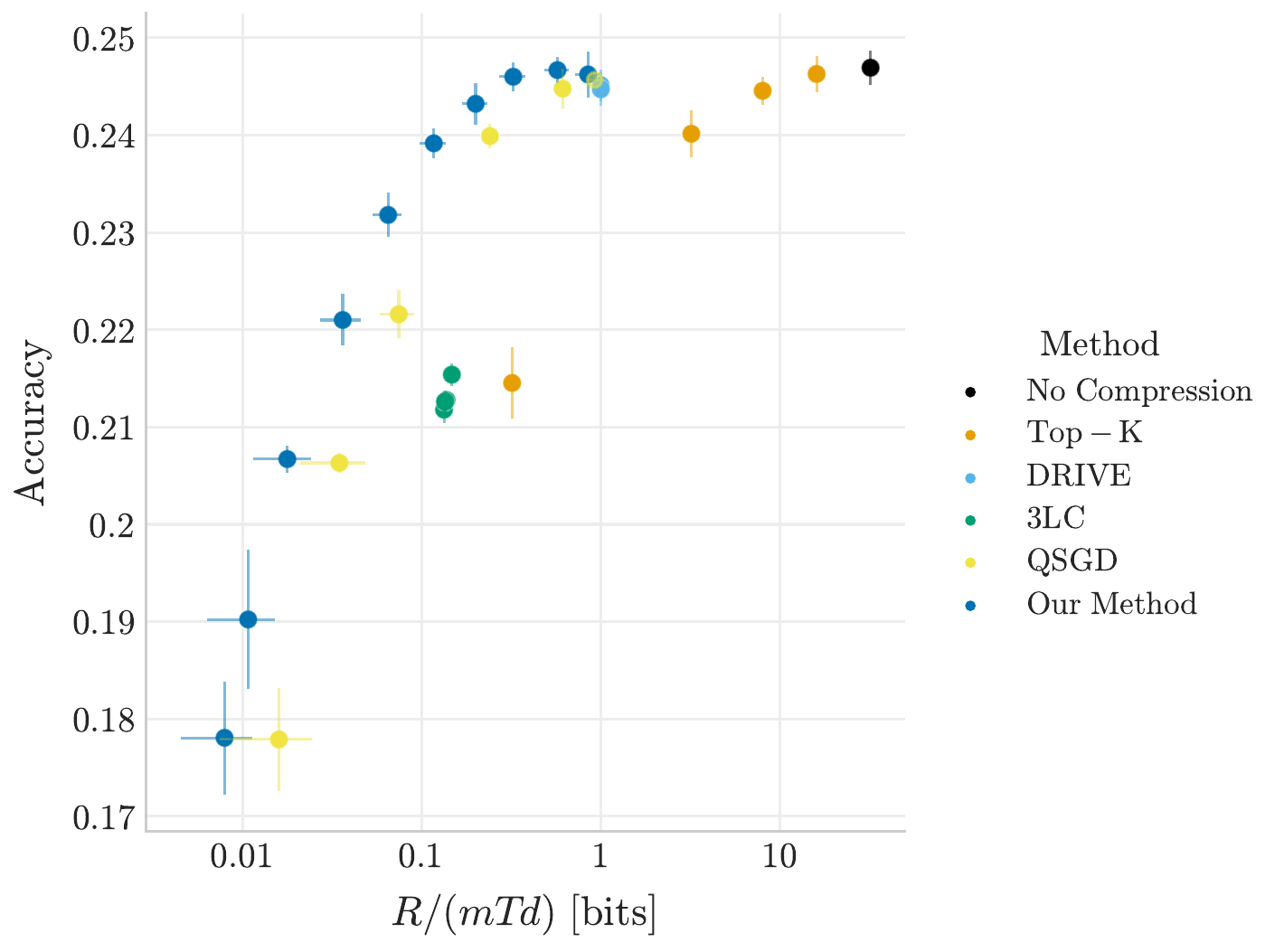}
    \caption{Stack Overflow NWP, FedAdam}
\end{subfigure}
\centering
\begin{subfigure}{0.48\linewidth}
    \centering
    \includegraphics[height=.75\linewidth]{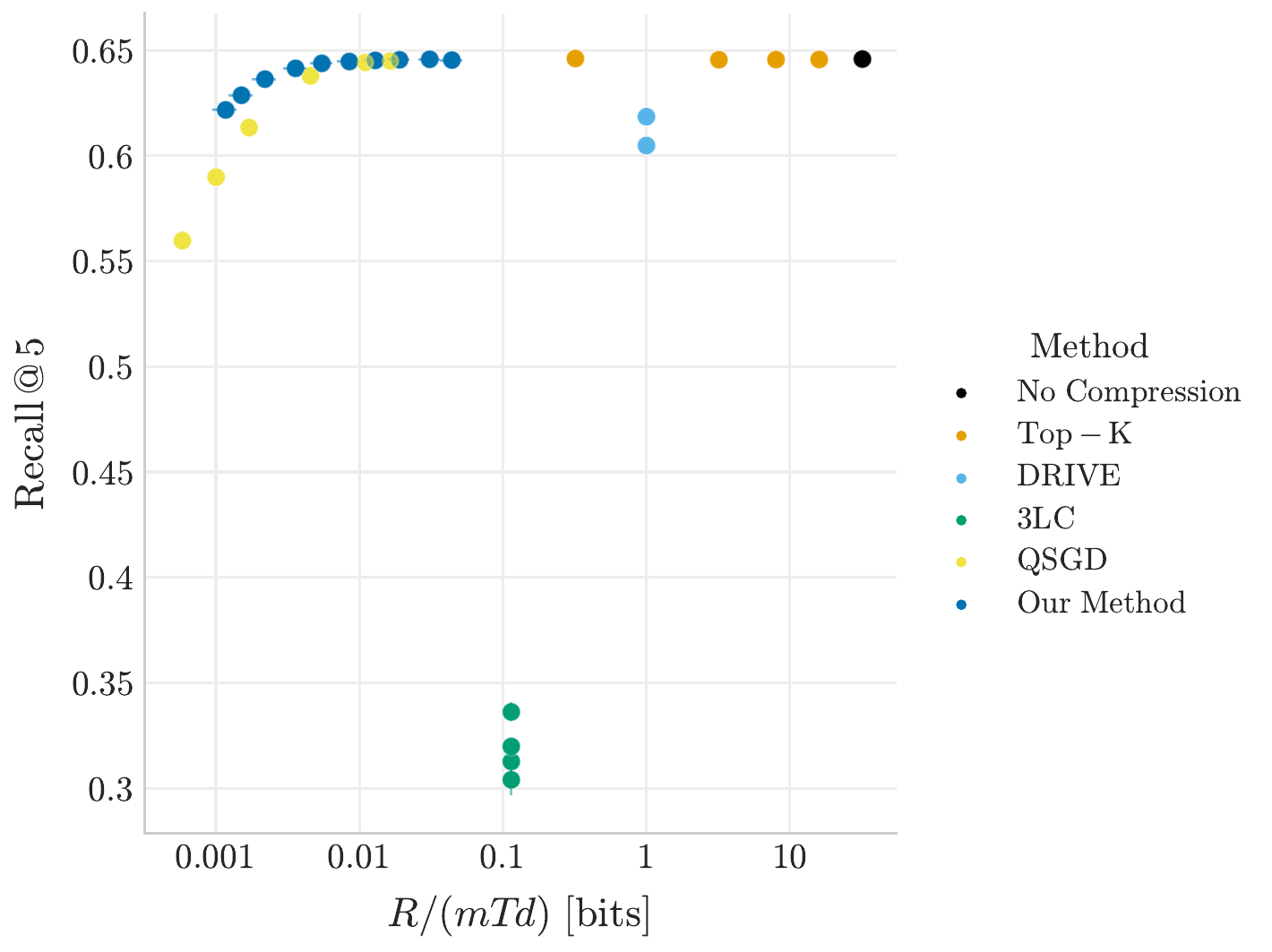}
    \caption{Stack Overflow TP, FedAdam}
\end{subfigure}
\caption{Our method performs competitively against all others in terms of the accuracy--communication cost trade-off. We use a range of $\Delta \in \{0.05 .. 17.5\}$. Error bars indicate variance in average per-coordinate rate and final model accuracy over five random trials.}
\end{figure}

\newpage
\subsection{Rotations}\label{appendix:rotations}

The same experiment as in Figure~\ref{so-fedadam-rotations}, with other tasks and optimizers. The rate--distortion tradeoff is significantly better without rotation for EMNIST and Stack Overflow datasets. It is only slightly better for CIFAR-100, due to the fact that the client updates are not as sparse as for the other tasks.

\begin{figure}[H]
\begin{subfigure}{0.48\linewidth}
    \centering
    \includegraphics[height=.75\linewidth]{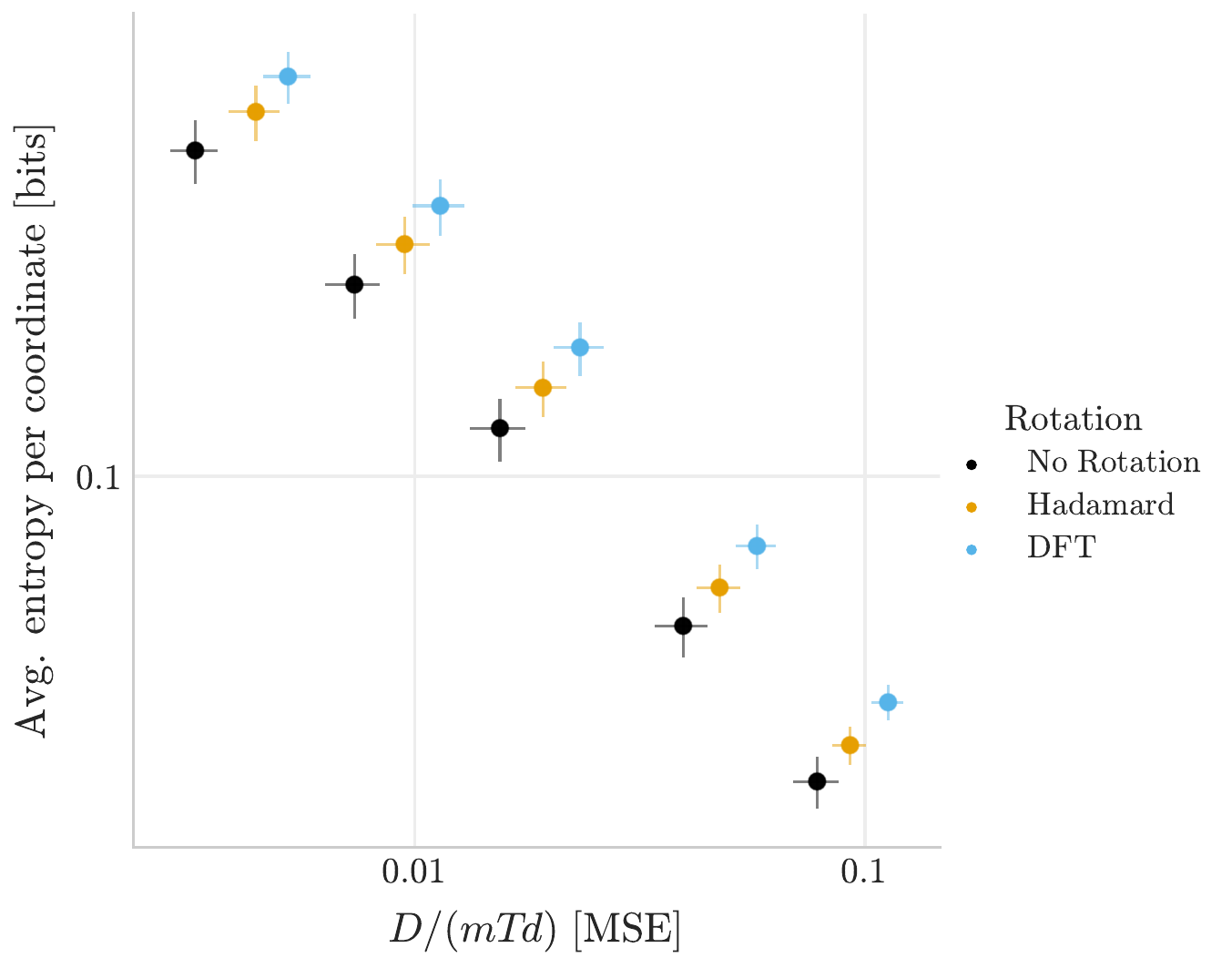}
    \caption{CIFAR-100, FedAdam}
\end{subfigure}
\begin{subfigure}{0.48\linewidth}
    \centering
    \includegraphics[height=.75\linewidth]{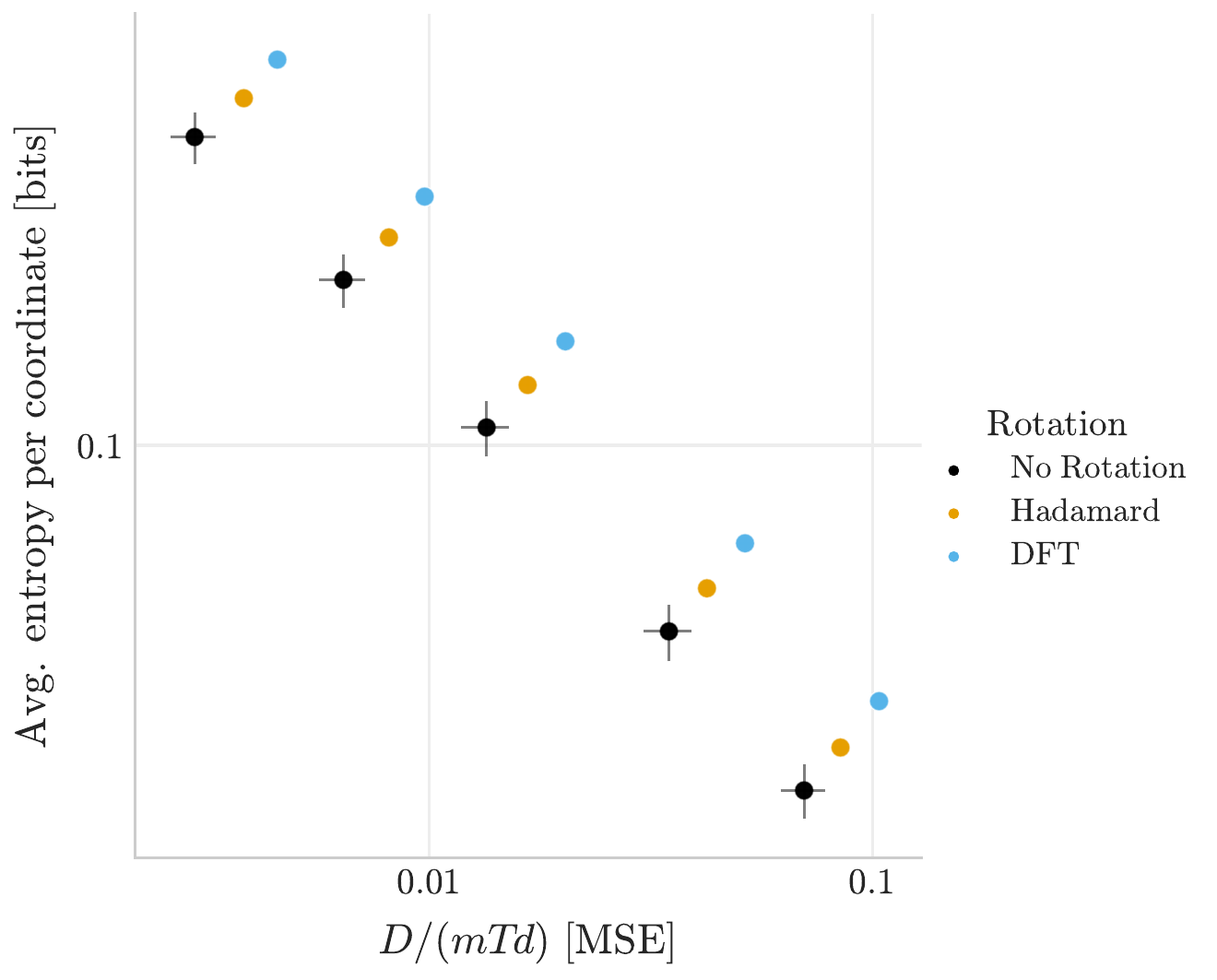}
    \caption{CIFAR-100, FedAvg}
\end{subfigure}
\begin{subfigure}{0.48\linewidth}
    \centering
    \includegraphics[height=.75\linewidth]{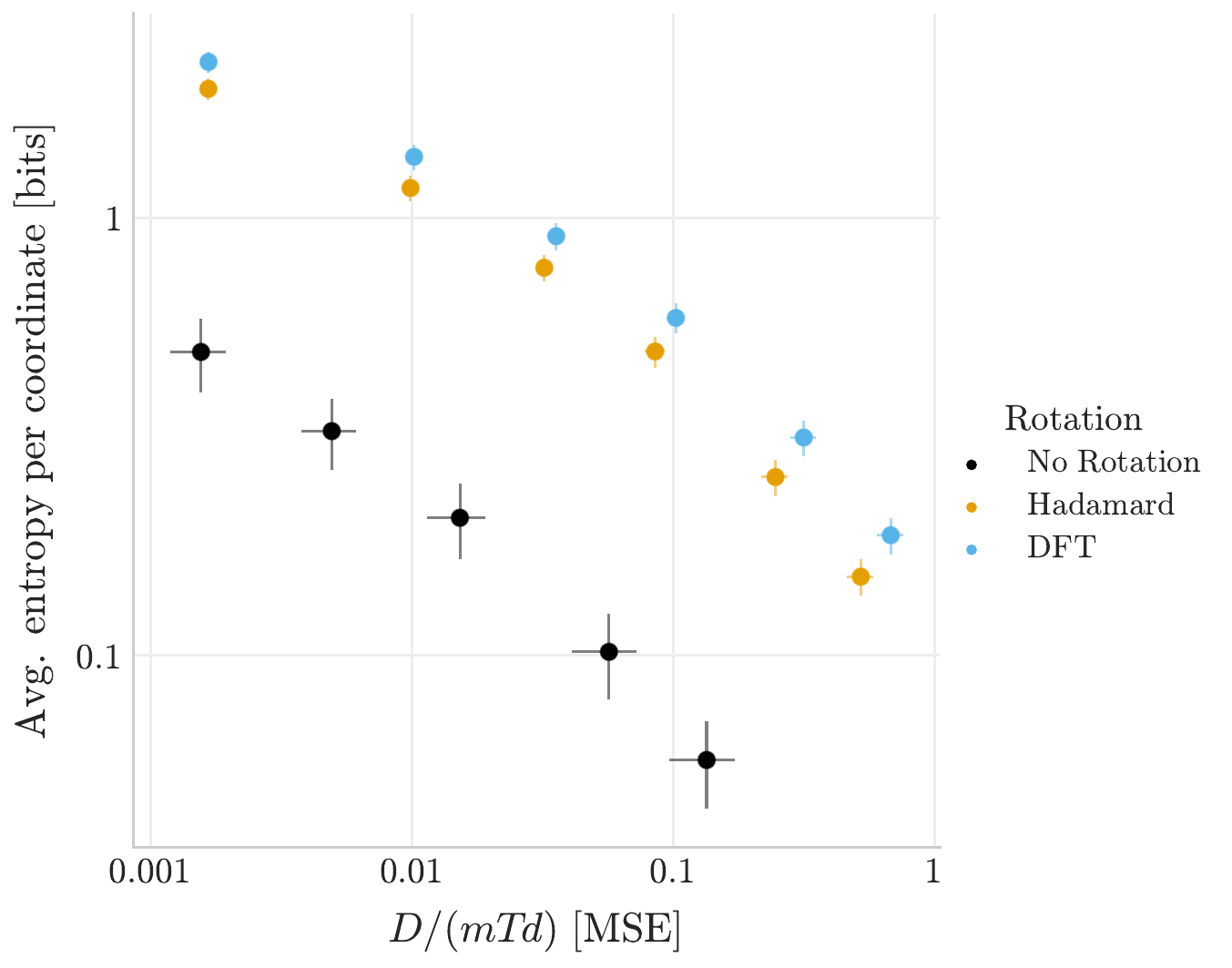}
    \caption{EMNIST, FedAdam}
\end{subfigure}
\begin{subfigure}{0.48\linewidth}
    \centering
    \includegraphics[height=.75\linewidth]{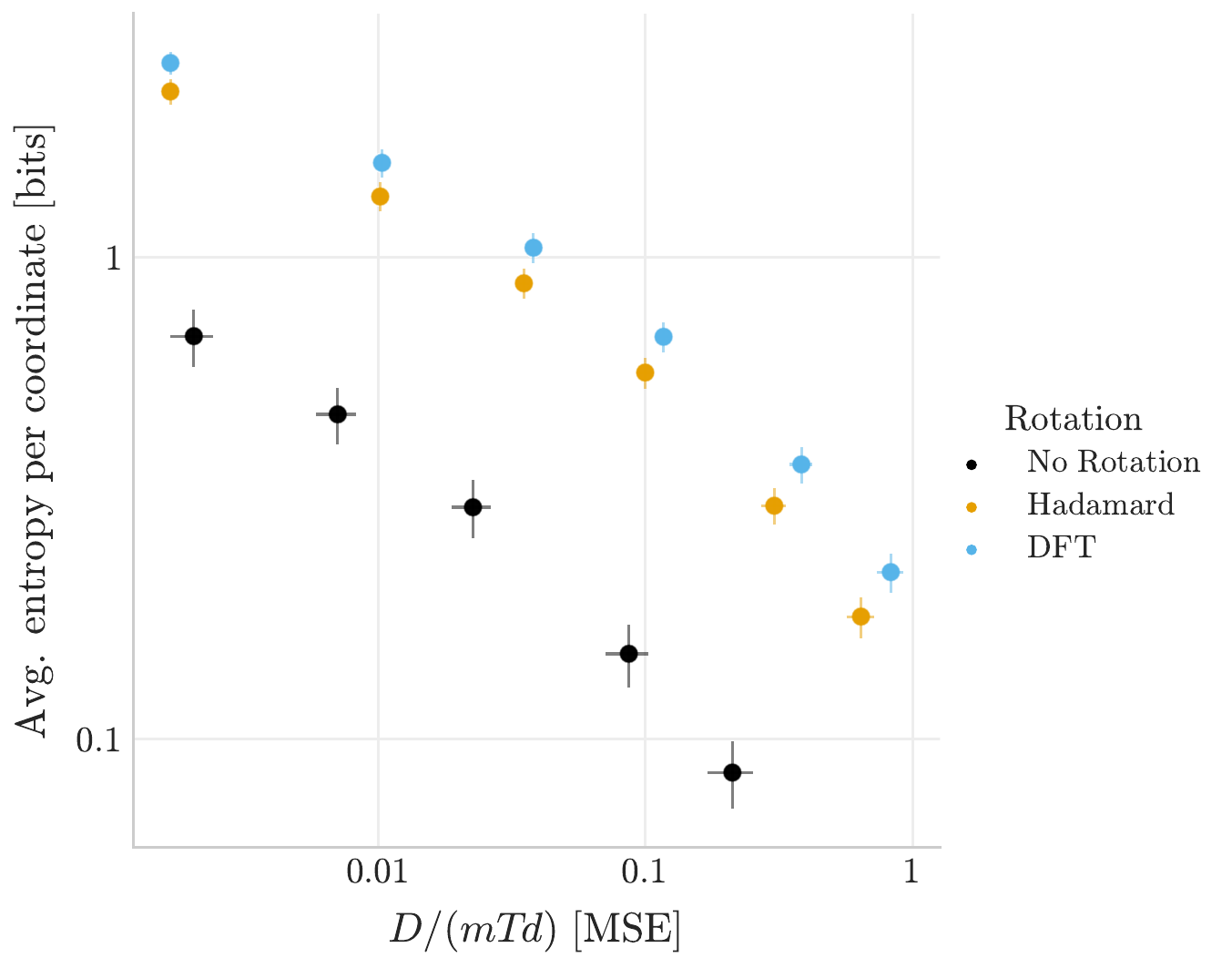}
    \caption{EMNIST, FedAvg}
\end{subfigure}
\begin{subfigure}{0.48\linewidth}
    \centering
    \includegraphics[height=.75\linewidth]{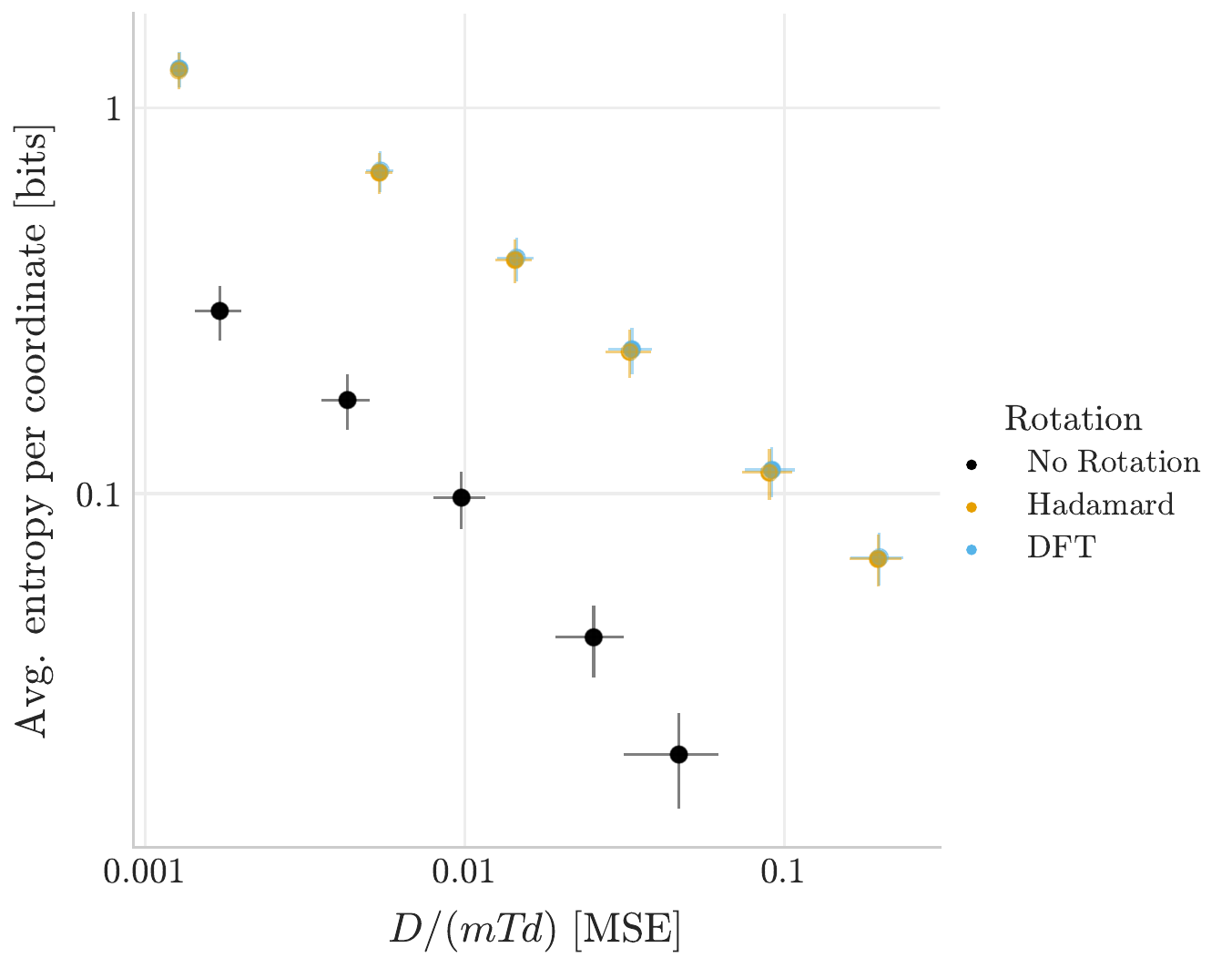}
    \caption{Stack Overflow NWP, FedAdam}
\end{subfigure}
\centering
\begin{subfigure}{0.48\linewidth}
    \centering
    \includegraphics[height=.75\linewidth]{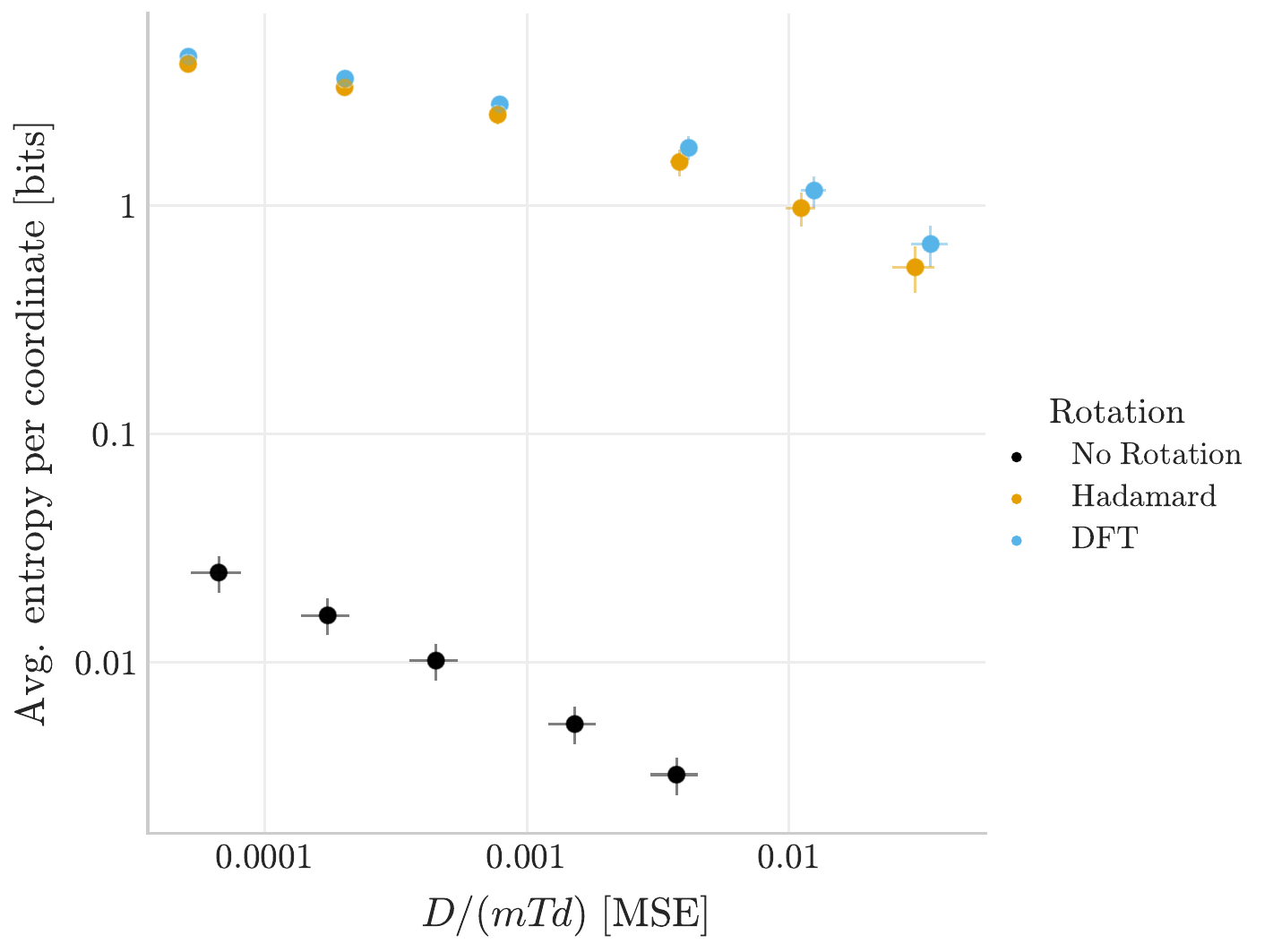}
    \caption{Stack Overflow TP, FedAdam}
\end{subfigure}
\caption{Transforming the client updates via a random rotation increases both entropy and distortion, producing a distribution with a worse entropy–distortion frontier. The effect is most dramatic on highly structured client updates. Error bars indicate variance in average per-coordinate distortion and average per-coordinate entropy over five random trials.}
\end{figure}

\newpage
The same experiment as in Figure~\ref{rotation-histogram}, across tasks and optimizers. Applying a fixed arbitrary rotation yields a lighter-tailed distribution with higher entropy. The effect is especially dramatic on the highly sparse client updates for the EMNIST and Stack Overflow tasks.

\begin{figure}[H]
\begin{subfigure}{0.48\linewidth}
    \centering
    \includegraphics[height=.75\linewidth]{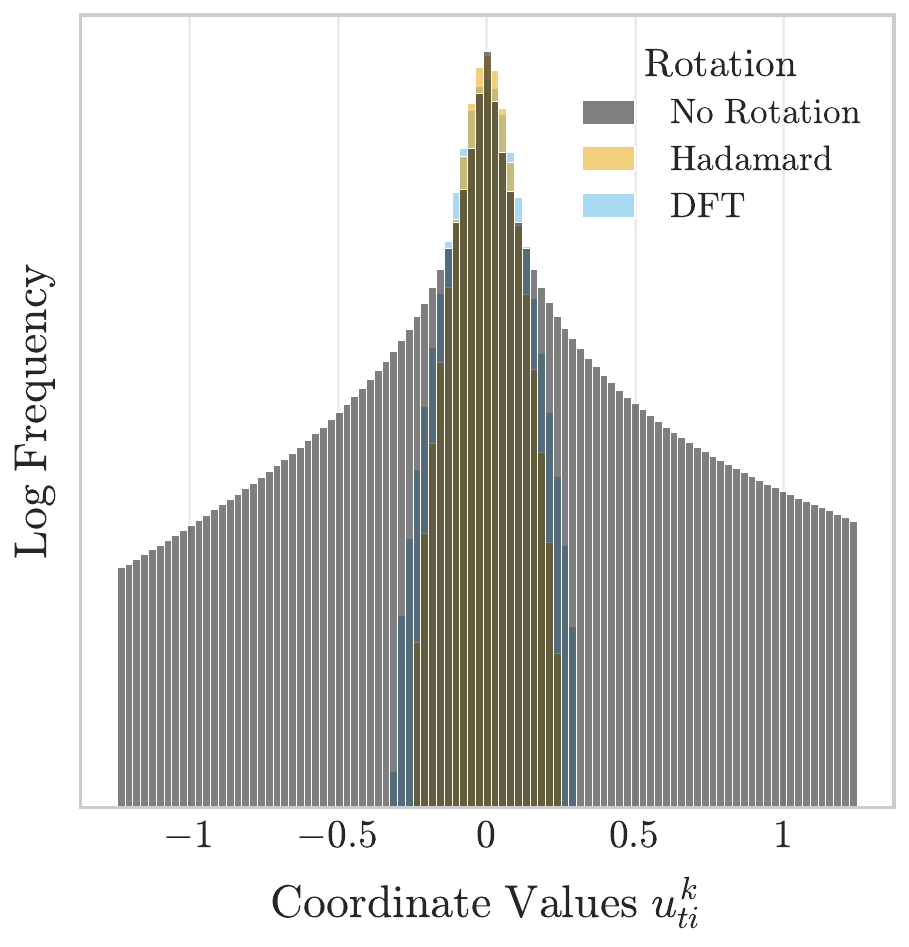}
    \caption{CIFAR-100, FedAdam}
\end{subfigure}
\begin{subfigure}{0.48\linewidth}
    \centering
    \includegraphics[height=.75\linewidth]{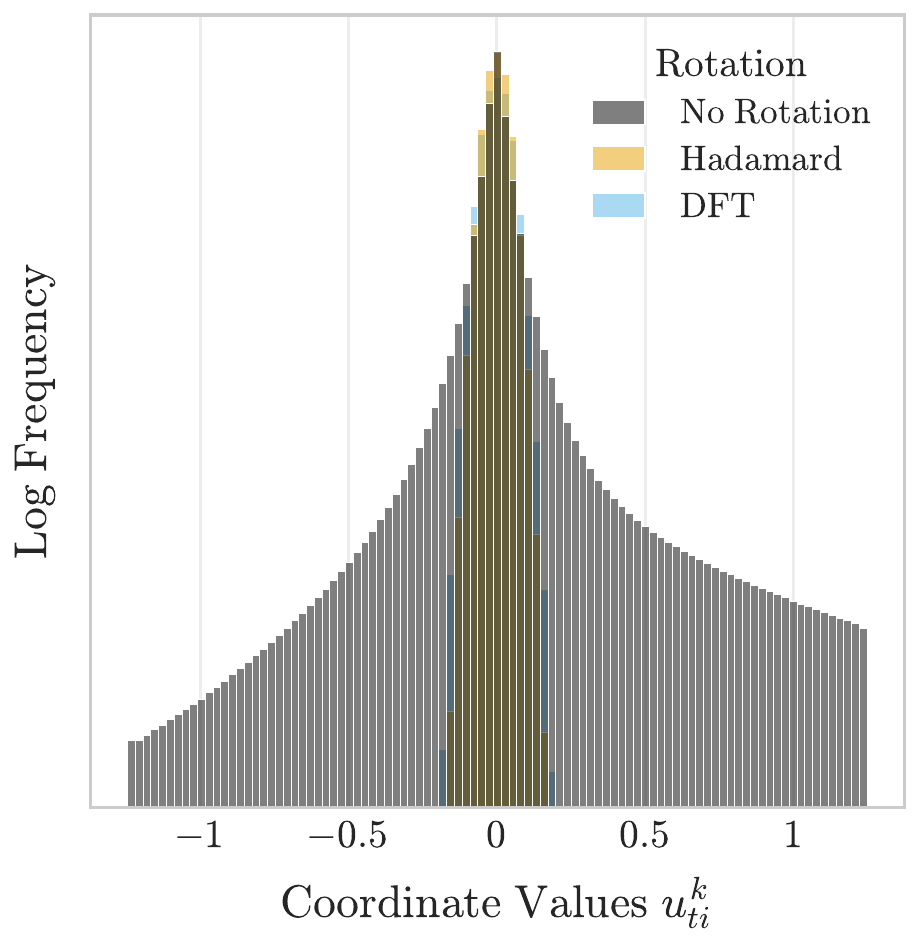}
    \caption{CIFAR-100, FedAvg}
\end{subfigure}
\begin{subfigure}{0.48\linewidth}
    \centering
    \includegraphics[height=.75\linewidth]{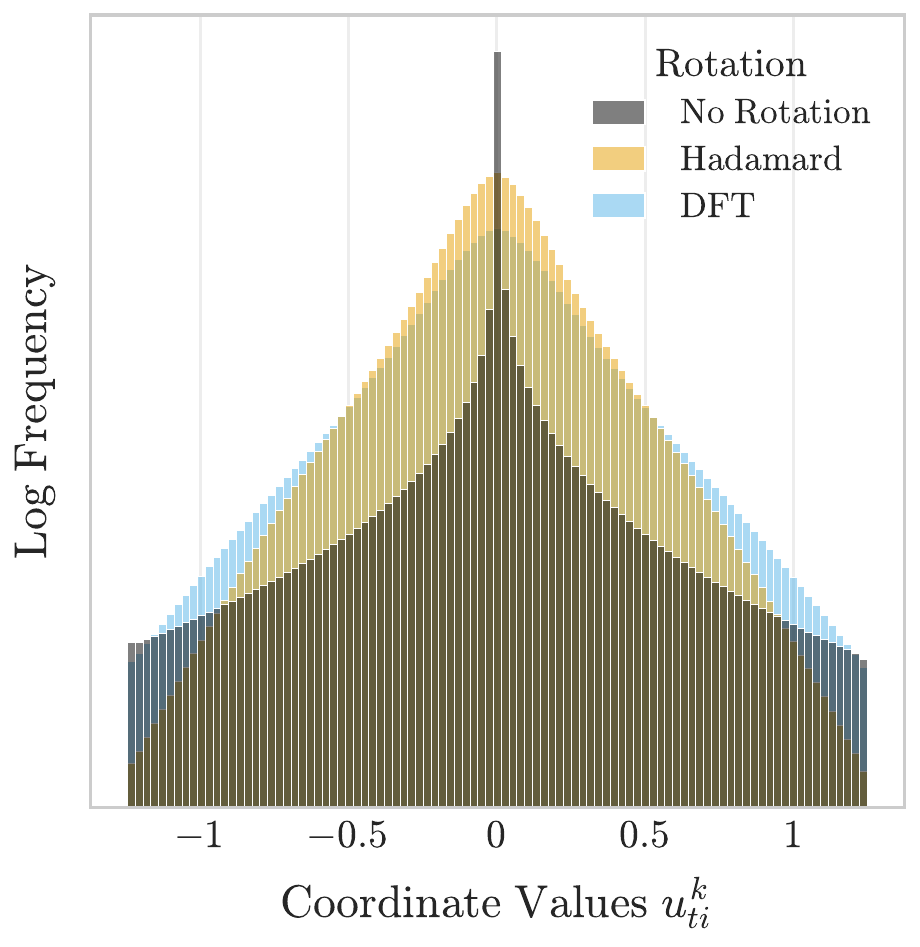}
    \caption{EMNIST, FedAdam}
\end{subfigure}
\begin{subfigure}{0.48\linewidth}
    \centering
    \includegraphics[height=.75\linewidth]{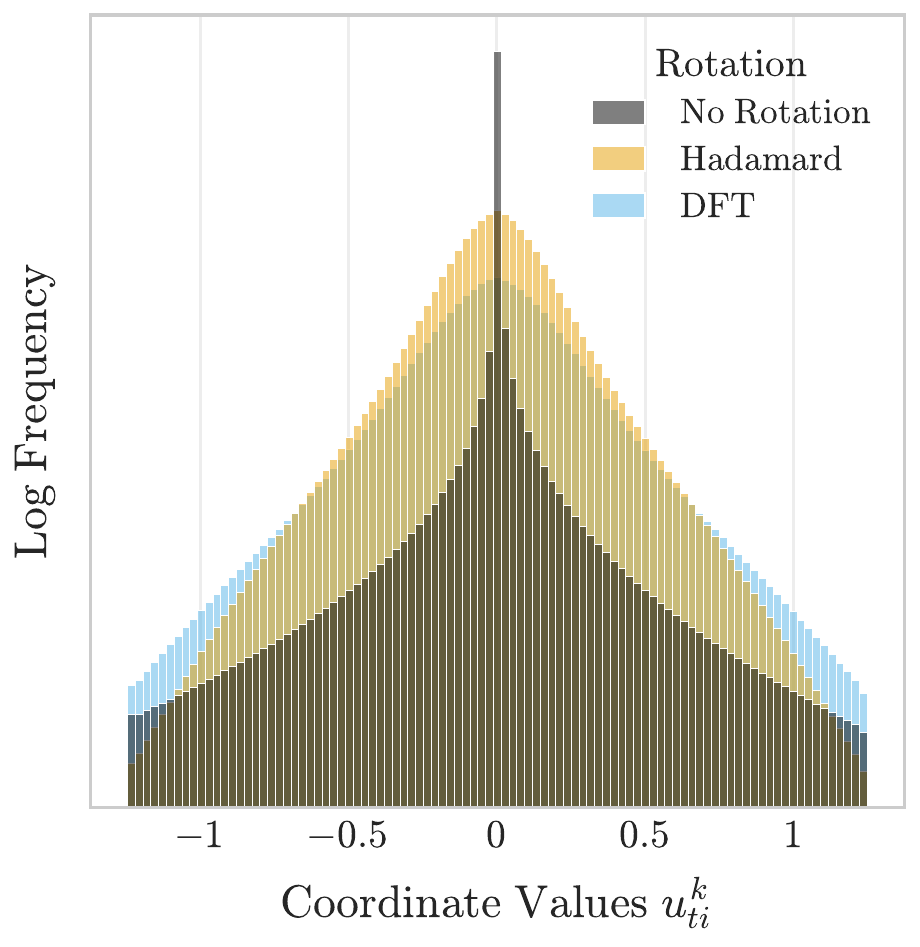}
    \caption{EMNIST, FedAvg}
\end{subfigure}
\begin{subfigure}{0.48\linewidth}
    \centering
    \includegraphics[height=.75\linewidth]{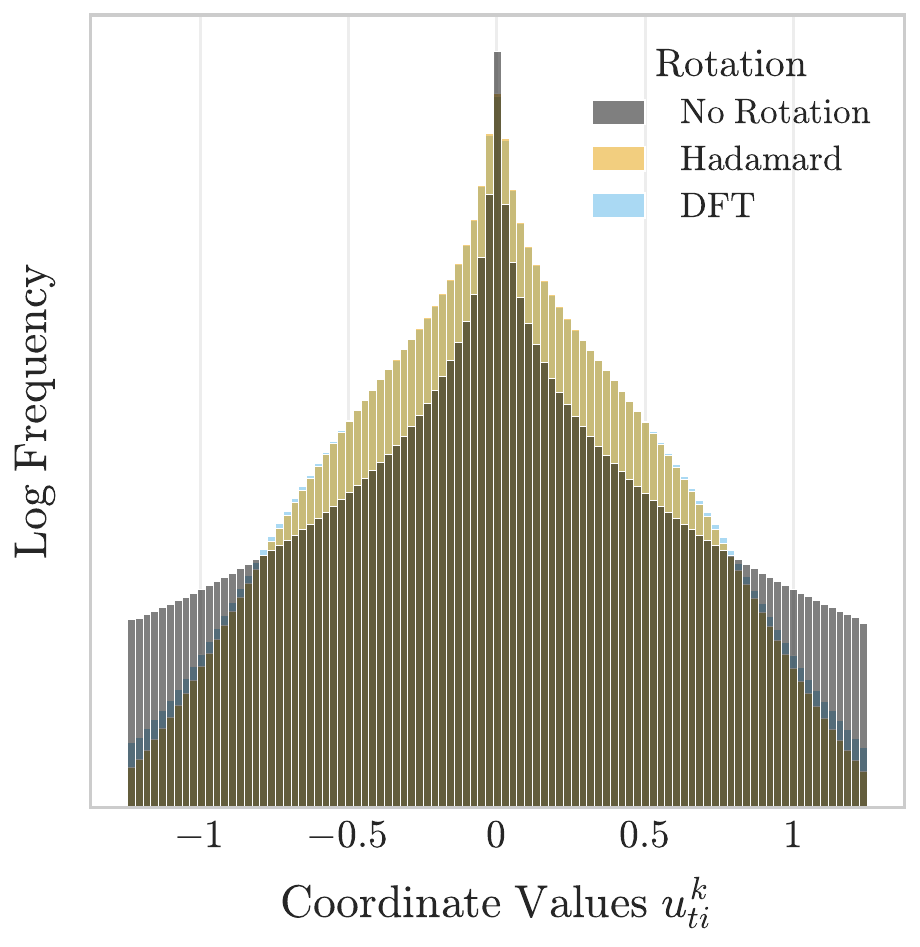}
    \caption{Stack Overflow NWP, FedAdam}
\end{subfigure}
\centering
\begin{subfigure}{0.48\linewidth}
    \centering
    \includegraphics[height=.75\linewidth]{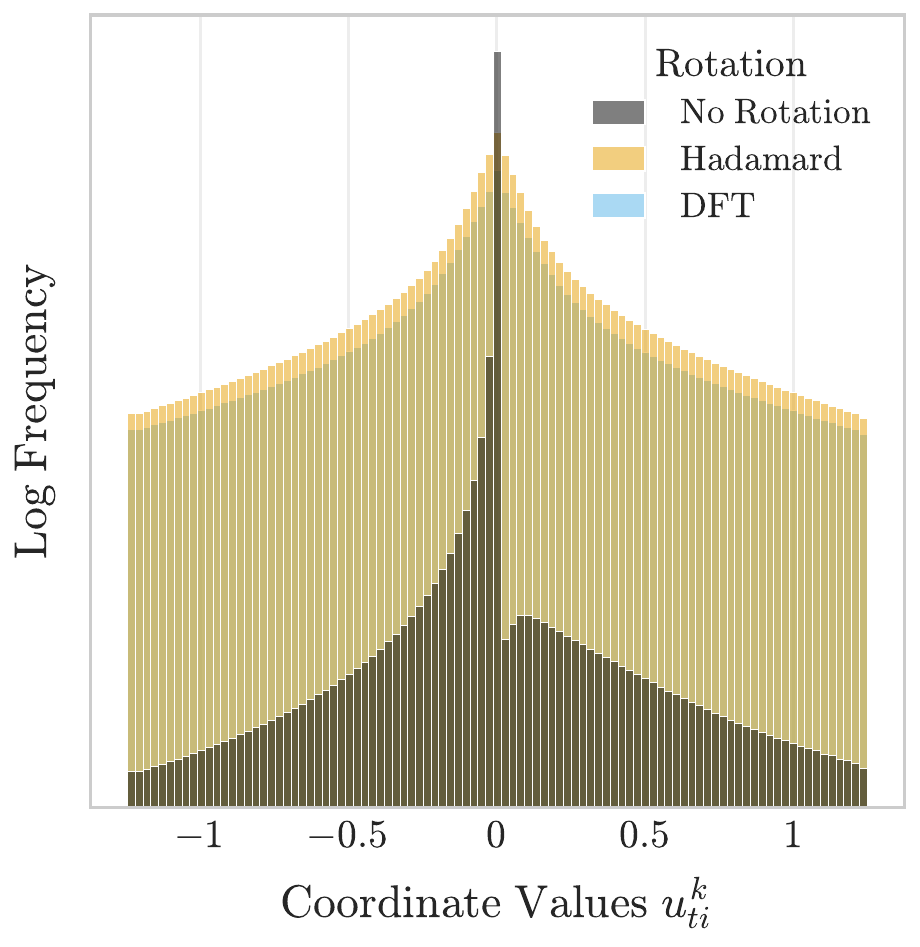}
    \caption{Stack Overflow TP, FedAdam}
\end{subfigure}
\caption{Transforming the client updates via a fixed rotation produces a lighter-tailed distribution with higher entropy. The effect is most dramatic on highly structured client updates.}
\end{figure}

\newpage
\subsection{Normalization}\label{appendix:normalization}

The same experiment as in Figure~\ref{stackoverflow_word-fedadam-normalization}, with other tasks and optimizers. QSGD yields similar performance on the CIFAR-100 dataset, where the clients have the same amount of training data. The gain in performance by our method is more pronounced on the more heterogeneous Stack Overflow tasks. 

\begin{figure}[H]
\begin{subfigure}{0.48\linewidth}
    \centering
    \includegraphics[height=.75\linewidth]{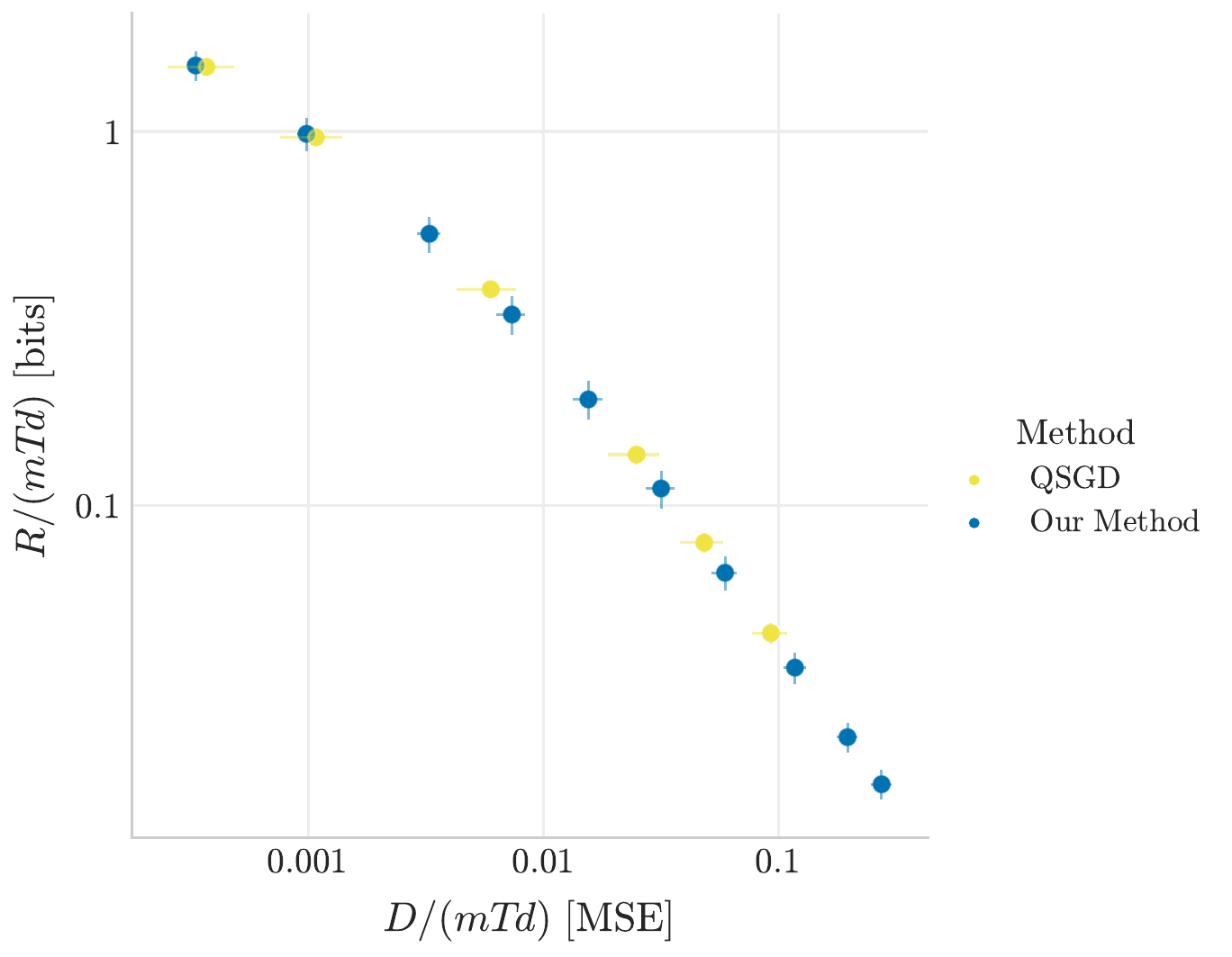}
    \caption{CIFAR-100, FedAdam}
\end{subfigure}
\begin{subfigure}{0.48\linewidth}
    \centering
    \includegraphics[height=.75\linewidth]{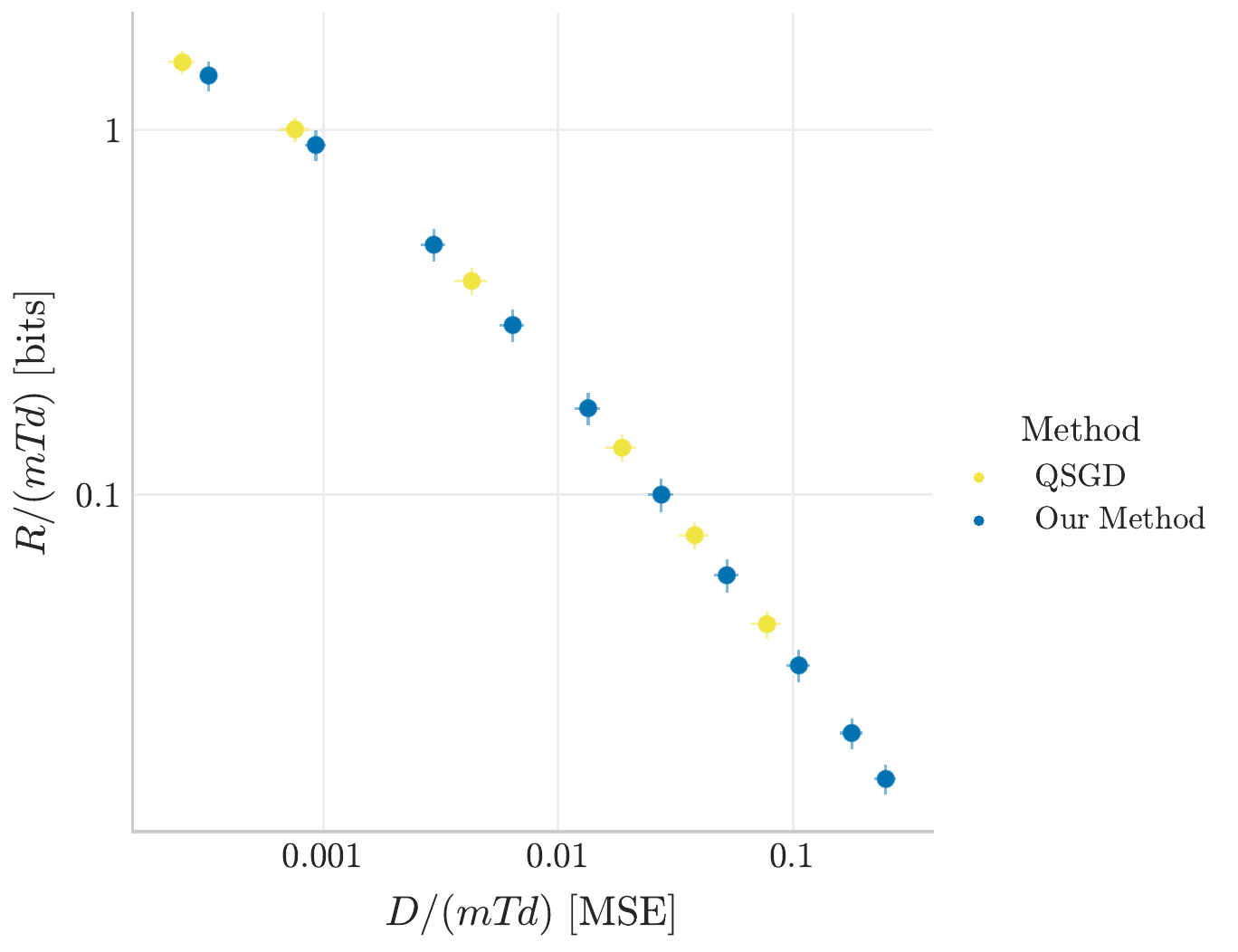}
    \caption{CIFAR-100, FedAvg}
\end{subfigure}
\begin{subfigure}{0.48\linewidth}
    \centering
    \includegraphics[height=.75\linewidth]{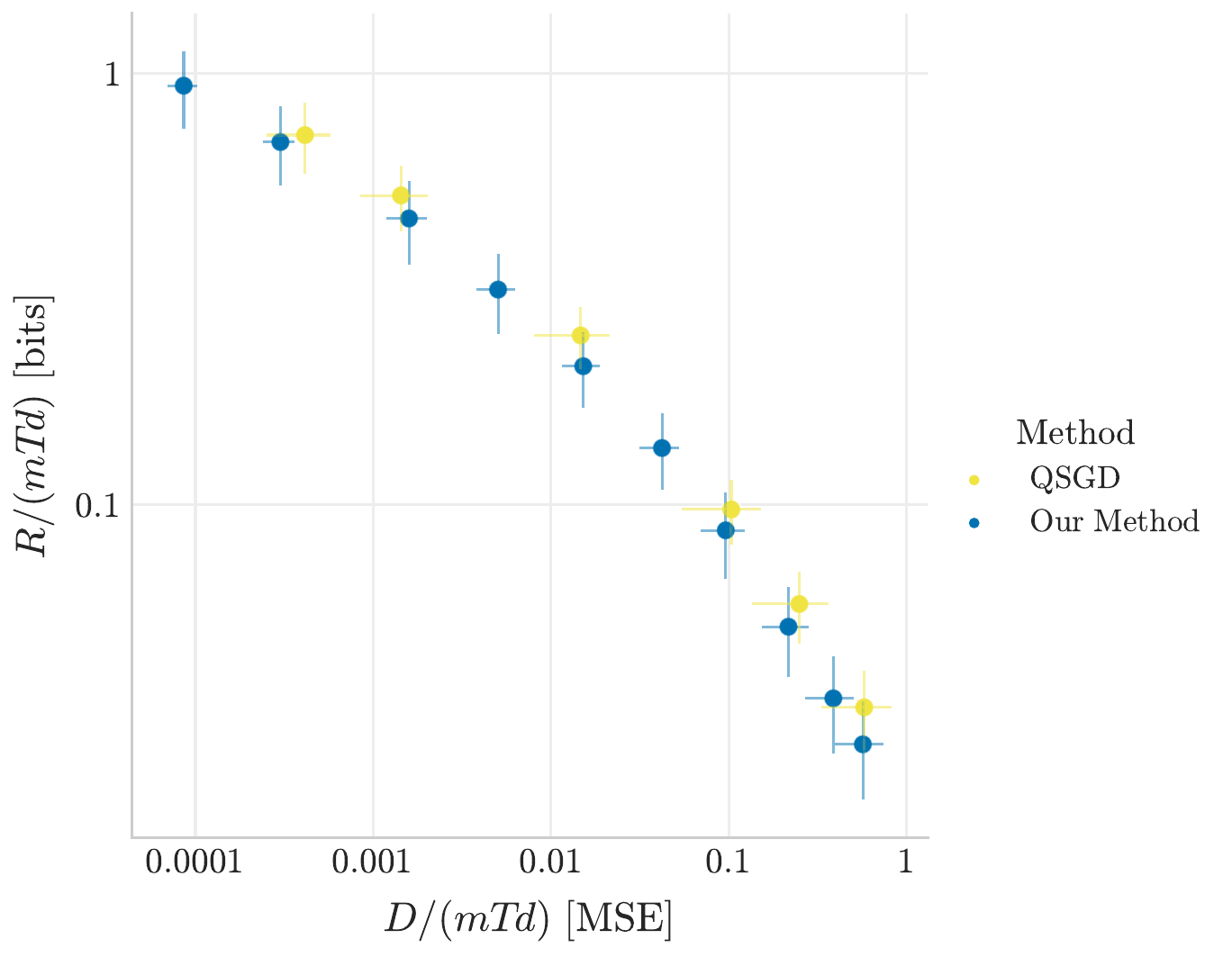}
    \caption{EMNIST, FedAdam}
\end{subfigure}
\begin{subfigure}{0.48\linewidth}
    \centering
    \includegraphics[height=.75\linewidth]{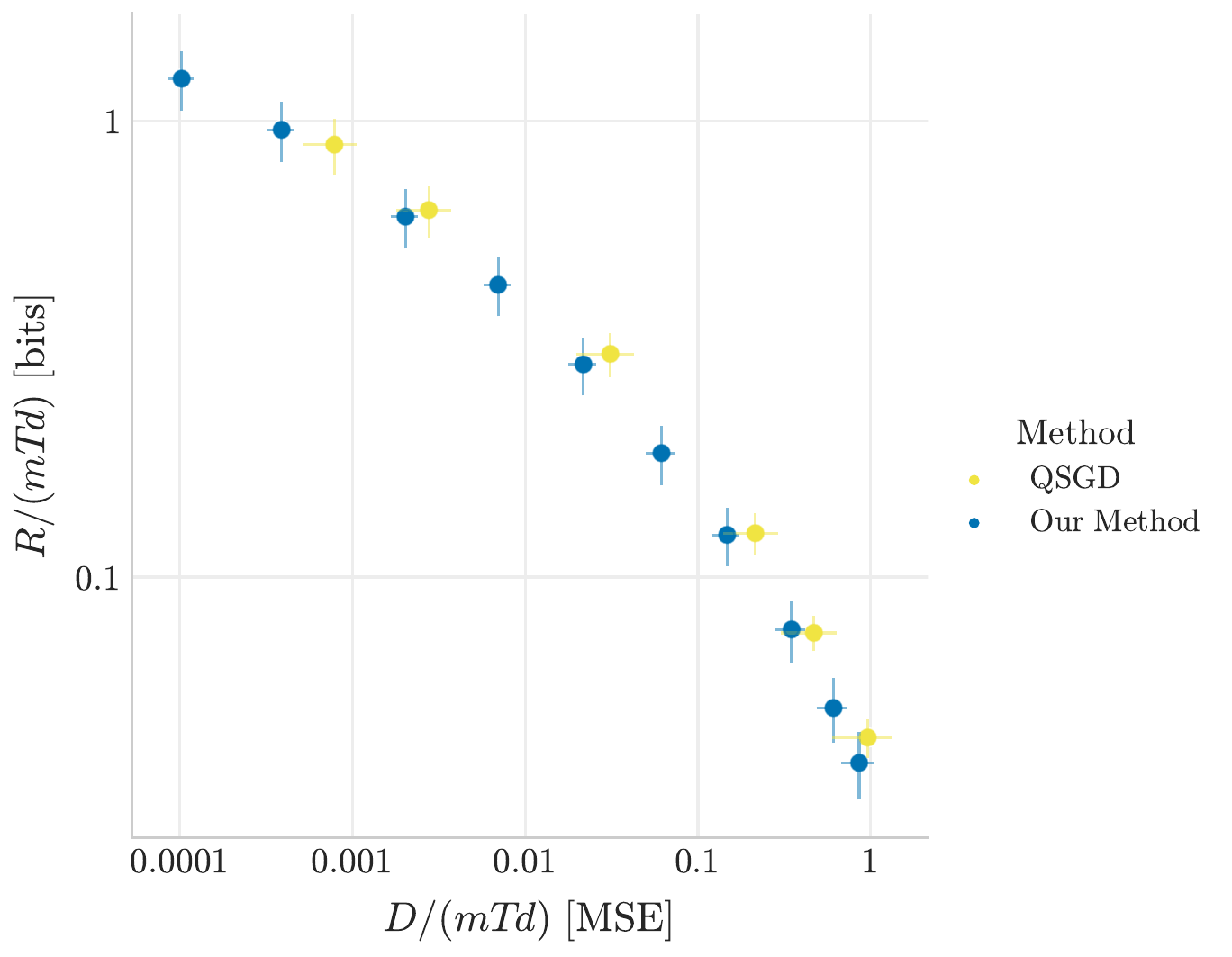}
    \caption{EMNIST, FedAvg}
\end{subfigure}
\begin{subfigure}{0.48\linewidth}
    \centering
    \includegraphics[height=.75\linewidth]{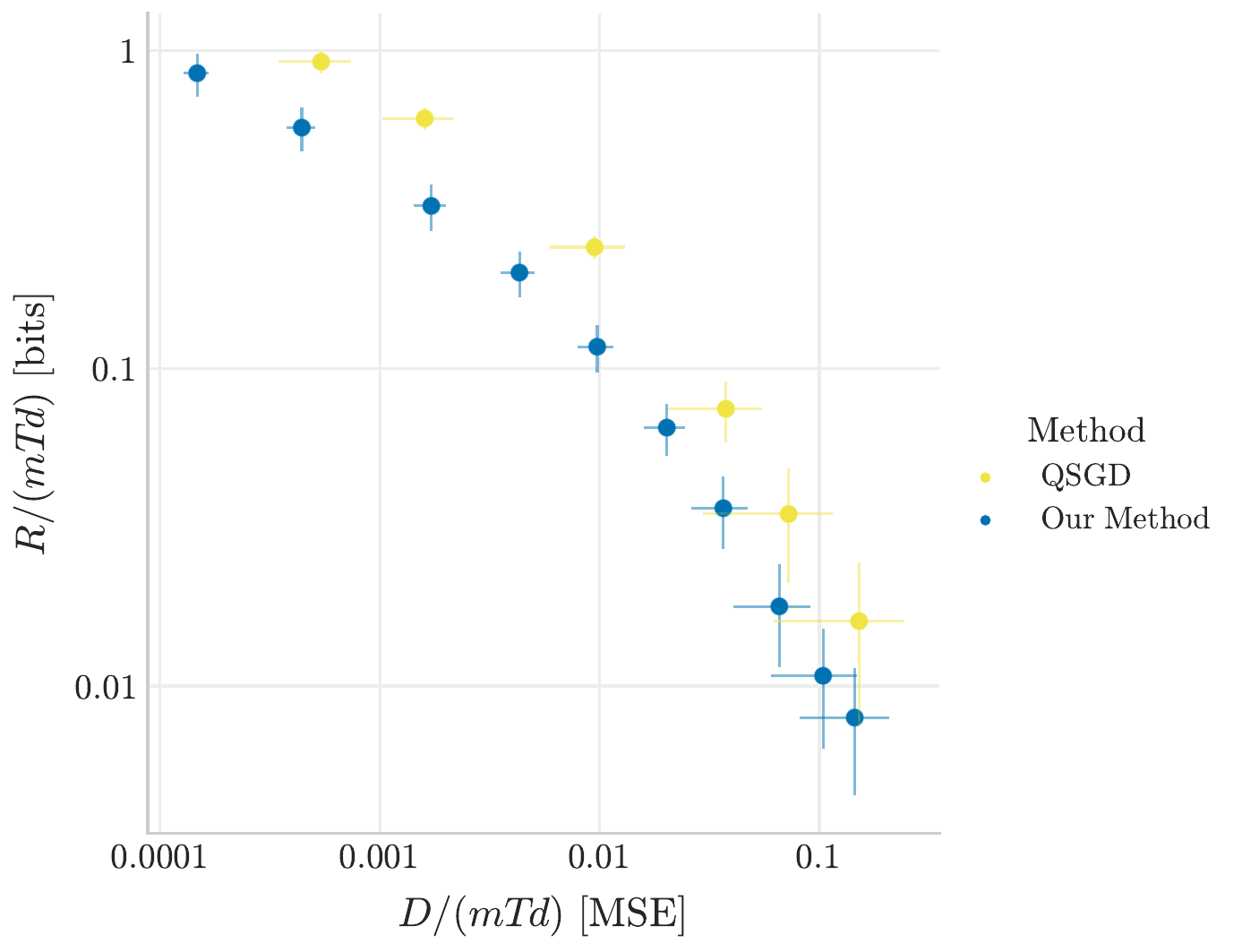}
    \caption{Stack Overflow NWP, FedAdam}
\end{subfigure}
\centering
\begin{subfigure}{0.48\linewidth}
    \centering
    \includegraphics[height=.75\linewidth]{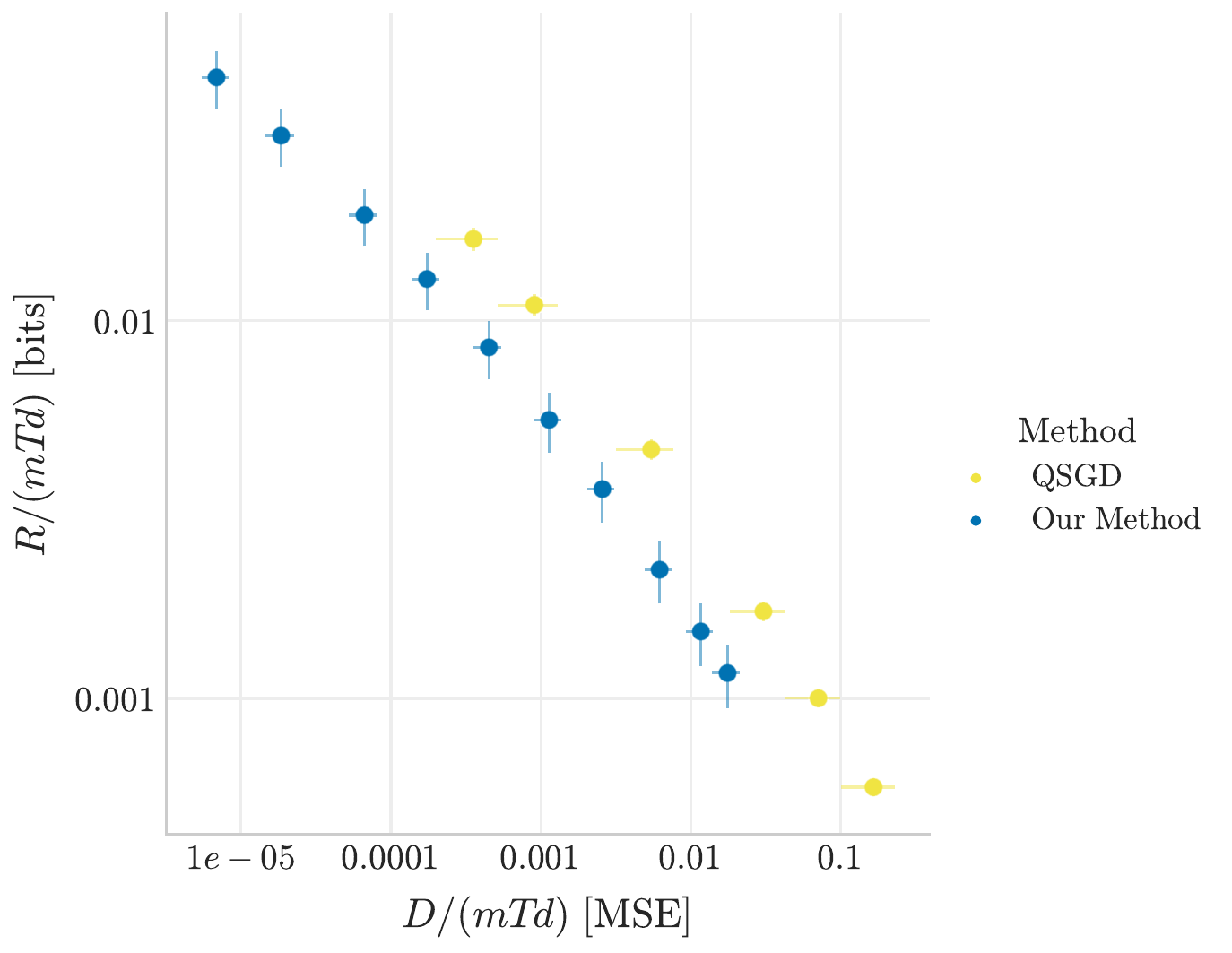}
    \caption{Stack Overflow TP, FedAdam}
\end{subfigure}
\caption{Transforming heterogeneous client updates via normalization results in a slightly worse rate--distortion performance. The effect is negligible on tasks with client updates of the same magnitude. Error bars indicate variance in average per-coordinate distortion and average per-coordinate rate over five random trials.}
\end{figure}

\newpage
\subsection{Decaying $\Delta$}\label{appendix:delta_decay}

The same experiment as in Figure~\ref{stackoverflow_word-fedadam-delta_decay}, with other tasks. We decay $\Delta$ over the course of training according to an exponential decay schedule specified by $\Delta_t = (\Delta_0 - \Delta_{min}) e^{-\rho t} + \Delta_{min}$. We select $\rho$, such that $\Delta_T \approx \Delta_{min}$ for $T=1500$. For the Stack Overflow next-word prediction task we set $\Delta_0 = 10.0$, $\Delta_{min} = 0.1$ and $\rho = 0.006$. For the other tasks we set $\Delta_0 = 20.0$, $\Delta_{min} = 1.0$ and $\rho = 0.004$. We observe that spending fewer bits initially and more bits later can improve the validation accuracy achieved for the cumulative communication cost incurred during training.

\begin{figure}[H]
\begin{subfigure}{0.48\linewidth}
    \centering
    \hspace{-1cm}
    \includegraphics[height=.65\linewidth]{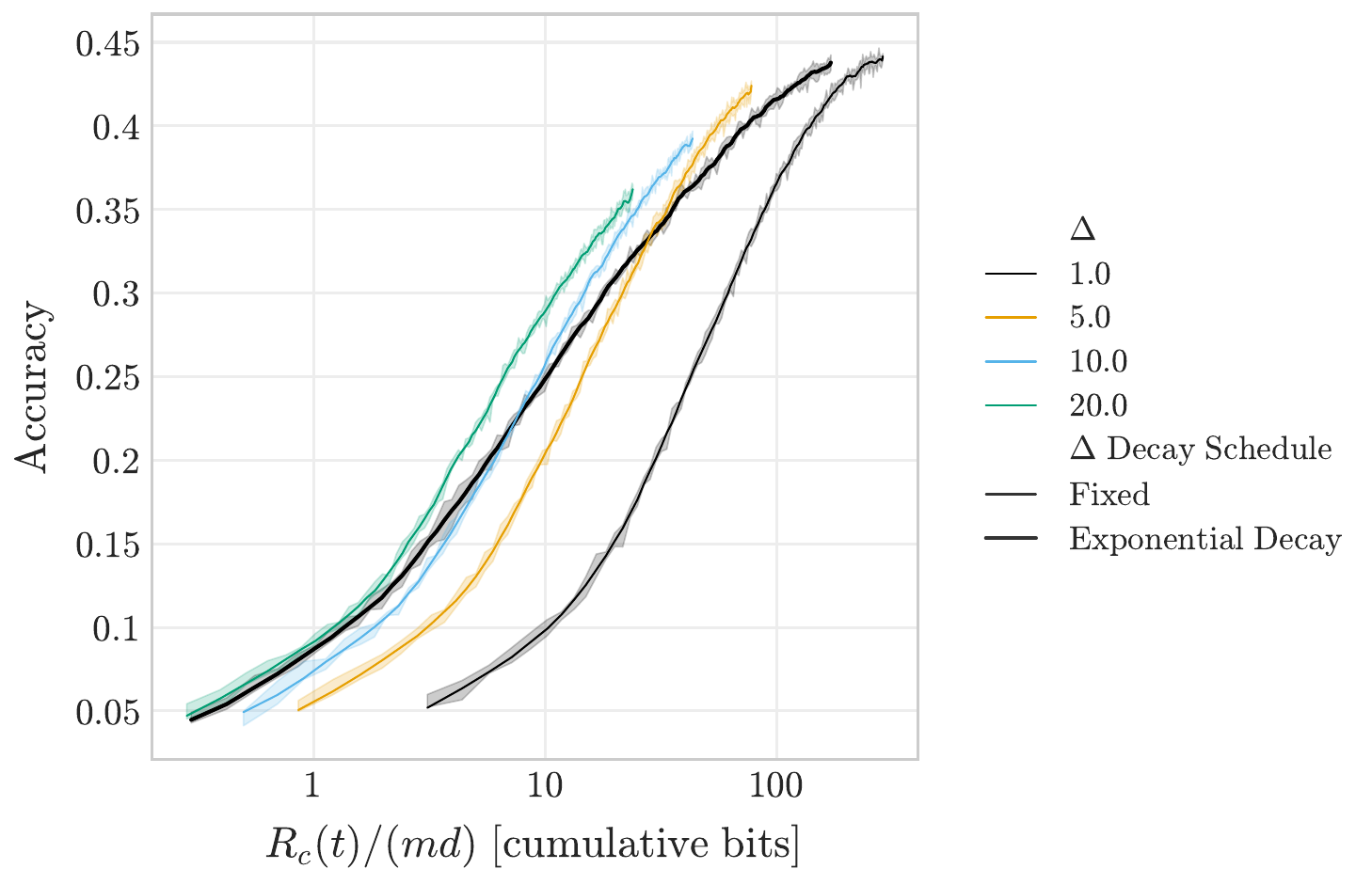}
    \caption{CIFAR-100, FedAdam}
\end{subfigure}
\begin{subfigure}{0.48\linewidth}
    \centering
    \hspace{1cm}
    \includegraphics[height=.65\linewidth]{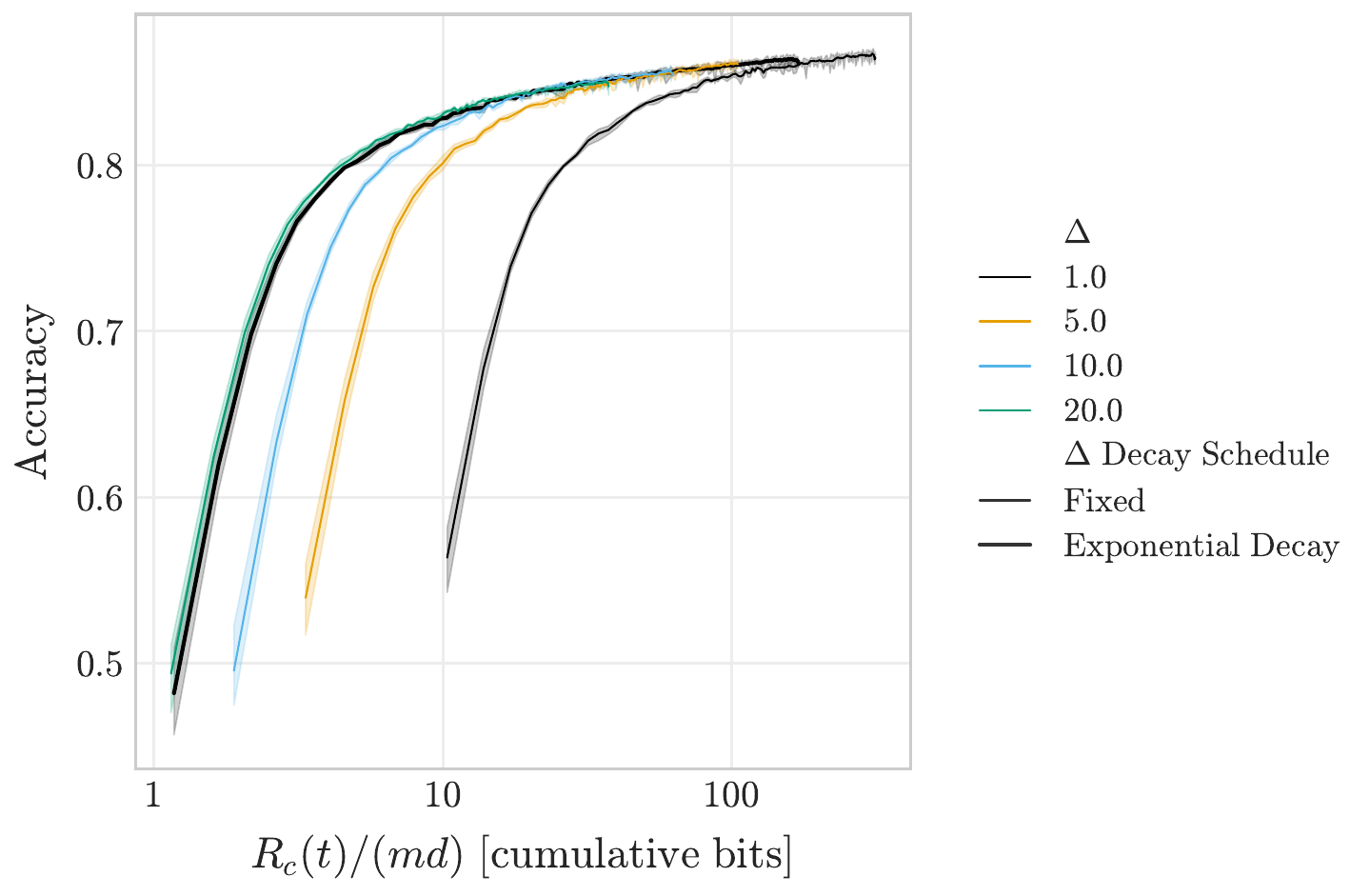}
    \caption{EMNIST, FedAdam}
\end{subfigure}
\begin{subfigure}{0.48\linewidth}
    \centering
    \hspace{-1cm}
    \includegraphics[height=.65\linewidth]{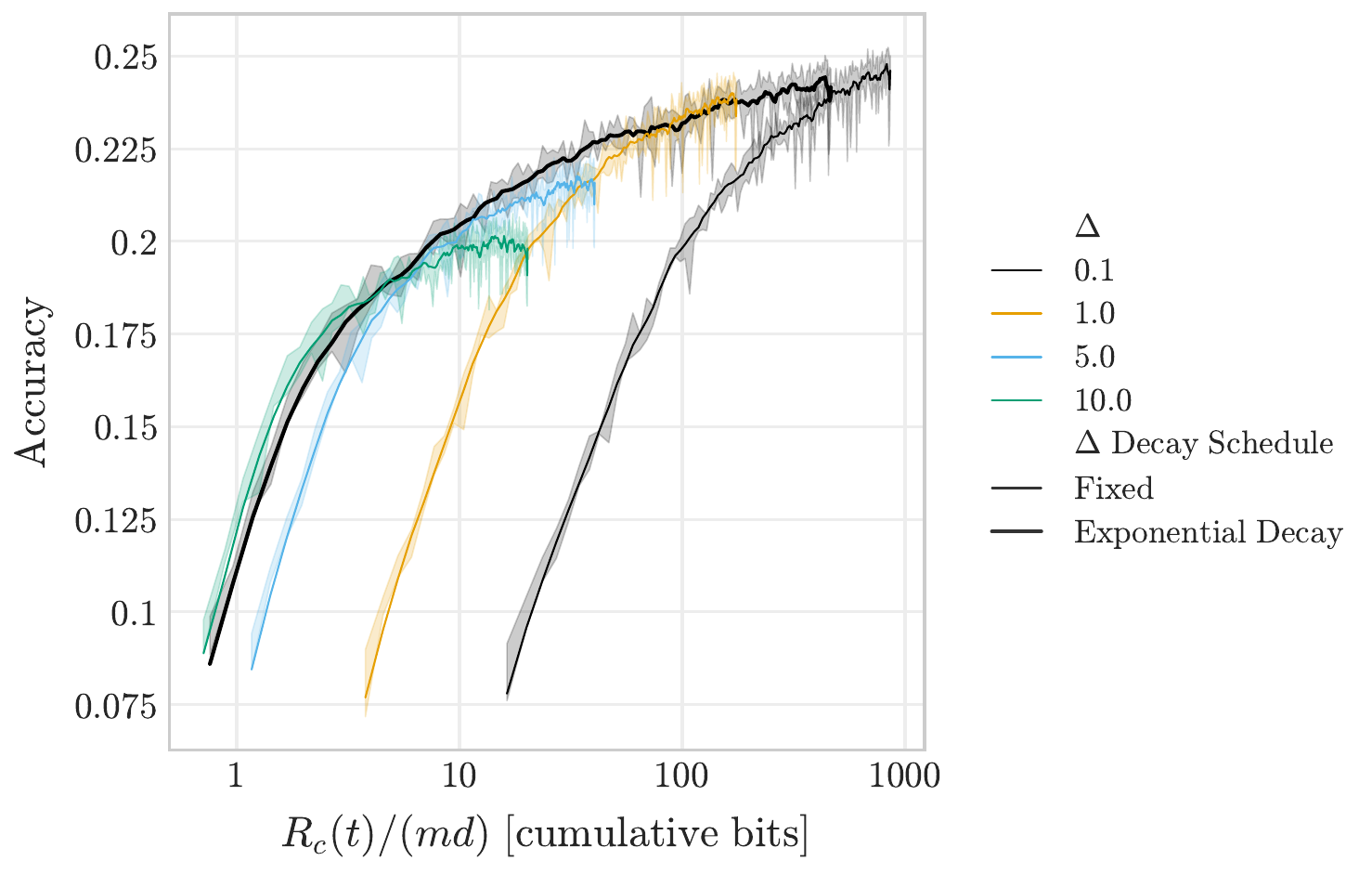}
    \caption{Stack Overflow NWP, FedAdam}
\end{subfigure}
\centering
\begin{subfigure}{0.48\linewidth}
    \centering
    \hspace{1cm}
    \includegraphics[height=.65\linewidth]{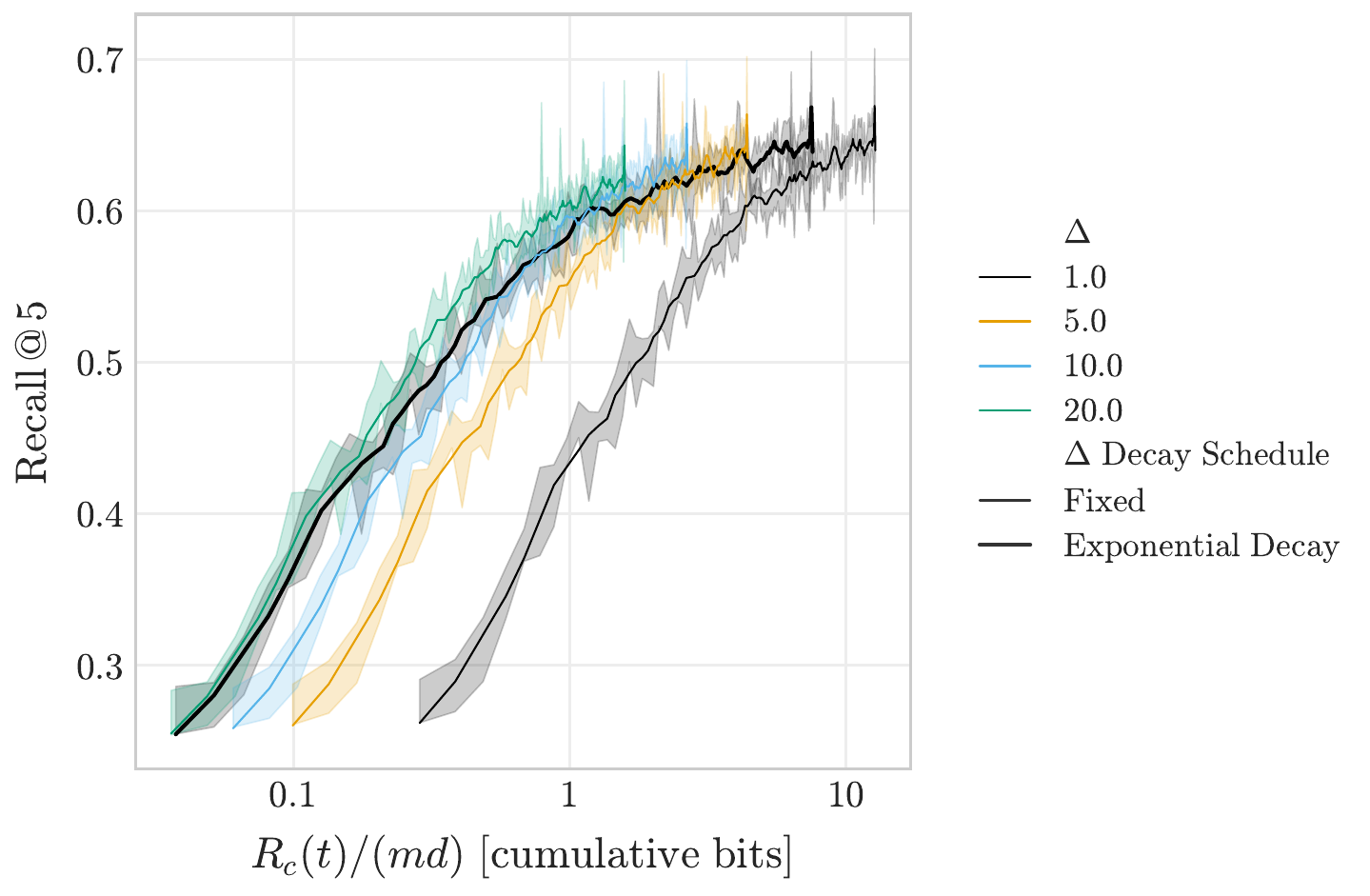}
    \caption{Stack Overflow TP, FedAdam}
\end{subfigure}
\caption{Exponentially decaying $\Delta$ results in improved performance over keeping $\Delta$ fixed when measuring the validation accuracy achieved for the cumulative bits spent on communicating each coordinate. This is particularly apparent on the Stack Overflow and EMNIST tasks. Here, $\Delta$ is exponentially decayed from $10.0$ to $0.1$ on Stack Overflow NWP and from $20.0$ to $1.0$ on CIFAR-100, EMNIST and Stack Overflow TP, and compared to fixed $\Delta$ within the respective decayed interval.}
\end{figure}

\end{document}